\documentclass[10pt,twocolumn,letterpaper]{article}

\makeatletter
\def\thanks#1{\protected@xdef\@thanks{\@thanks
        \protect\footnotetext{#1}}}
\makeatother

\usepackage{iccv}
\usepackage{times}
\usepackage{epsfig}
\usepackage{graphicx}
\usepackage{amsmath}
\usepackage{amssymb}
\usepackage{pifont}
\usepackage{booktabs}
\newcommand{\cmark}{\ding{51}}%
\newcommand{\xmark}{\ding{55}}%
\usepackage{makecell}
\usepackage{multirow}
\usepackage{multicol}
\usepackage[table]{xcolor}

\usepackage[pagebackref=true,breaklinks=true,colorlinks,bookmarks=false]{hyperref}

\newcommand{\tabincell}[2]{\begin{tabular}{@{}#1@{}}#2\end{tabular}}

\newcommand{\PreserveBackslash}[1]{\let\temp=\\#1\let\\=\temp}
\newcolumntype{C}[1]{>{\PreserveBackslash\centering}p{#1}}
\newcolumntype{R}[1]{>{\PreserveBackslash\raggedleft}p{#1}}
\newcolumntype{L}[1]{>{\PreserveBackslash\raggedright}p{#1}}

\iccvfinalcopy 

\def\iccvPaperID{6999} 

\begin{document}

\title{\emph{PlanarTrack}: A Large-scale Challenging Benchmark for Planar Object Tracking}

\author{Xinran Liu$^{*}$ \; Xiaoqiong Liu$^{*}$ \; Ziruo Yi$^{*}$ \; Xin Zhou$^{*}$ \;  Thanh Le \;  Libo Zhang$^{\dag}$ \; \\
Yan Huang \; Qing Yang \; Heng Fan\\
{\small Institute of Software, Chinese Academy of Sciences} \;\; {\small Dept. of Computer Science \& Engineering, University of North Texas}
\thanks{$^{*}$Equal contributions.}
\thanks{$^{\dag}$Corresponding author.}
}

\maketitle

\begin{abstract}
   Planar object tracking is a critical computer vision problem and has drawn increasing interest owing to its key roles in robotics, augmented reality, etc. Despite rapid progress, its further development, especially in the deep learning era, is largely hindered due to the lack of large-scale challenging benchmarks. Addressing this, we introduce \textbf{PlanarTrack}, a large-scale challenging planar tracking benchmark. Specifically, PlanarTrack consists of 1,000 videos with more than 490K images. All these videos are collected in complex unconstrained scenarios from the wild, which makes PlanarTrack, compared with existing benchmarks, more challenging but realistic for real-world applications. To ensure the high-quality annotation, each frame in PlanarTrack is manually labeled using four corners with multiple-round careful inspection and refinement. To our best knowledge, PlanarTrack, to date, is the largest and most challenging dataset dedicated to planar object tracking. In order to analyze the proposed PlanarTrack, we evaluate 10 planar trackers and conduct comprehensive comparisons and in-depth analysis. Our results, not surprisingly, demonstrate that current top-performing planar trackers degenerate significantly on the challenging PlanarTrack and more efforts are needed to improve planar tracking in the future. In addition, we further derive a variant named \textbf{PlanarTrack}$_{\mathbf{BB}}$ for generic object tracking from PlanarTrack.
    Our evaluation of 10 excellent generic trackers on PlanarTrack$_{\mathrm{BB}}$ manifests that, surprisingly, PlanarTrack$_{\mathrm{BB}}$ is even more challenging than several popular generic tracking benchmarks and more attention should be paid to handle such planar objects, though they are rigid. All benchmarks and evaluations  will be released at the \href{https://hengfan2010.github.io/projects/PlanarTrack/}{project webpage}.
\end{abstract}

\section{Introduction}
\label{intro}

\begin{figure}[!t]
    \centering
    \includegraphics[width=0.966\linewidth]{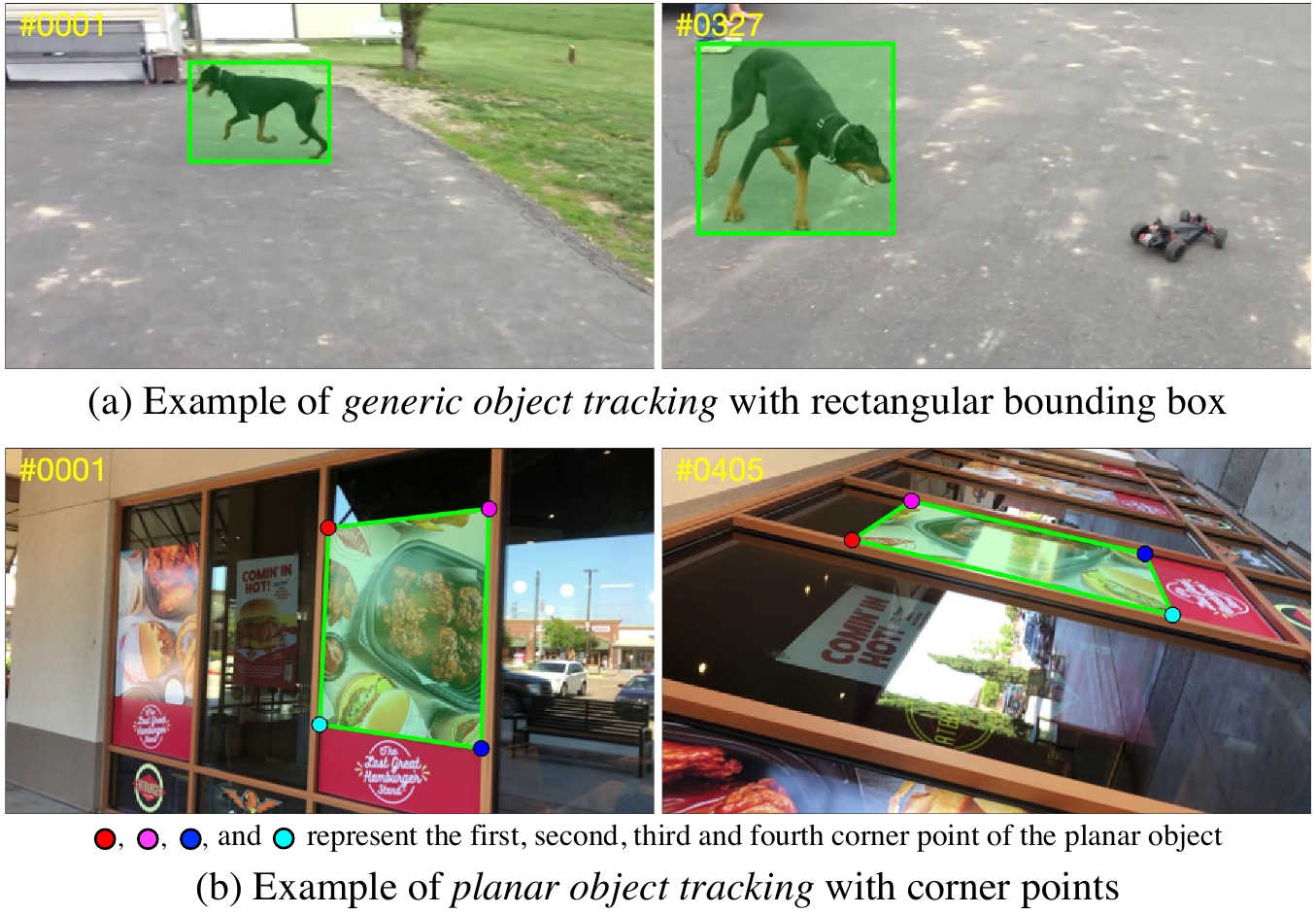}
    \caption{Generic object tracking (a) and planar object tracking (b). The former estimates axis-aligned rectangular bounding boxes for the target object, while the latter (our focus in this work) calculates 2D transformations of the target object to obtain the corresponding corner points for localization. \emph{All figures throughout this paper are best viewed in color and by zooming in}.}
    \label{fig1}
\end{figure}

Planar object tracking is one of the crucial problems in computer vision. Different than generic object tracking in which the goal is to locate the target object with axis-aligned rectangular bounding boxes~\cite{fan2019lasot,wu2013online}, planar object tracking aims to estimate 2D transformations (\eg, homograph) of the target and locate it with corner points (see Fig.~\ref{fig1}). Owing to its importance in robotics and augmented reality (AR), planar object tracking has attracted increasing attentions in recent years. In particular, several benchmarks (\eg,~\cite{liang2018planar,roy2015tracking,liang2021planar}) have been specially developed for evaluating and comparing different planar trackers, which greatly facilitates related research and progress on this topic. Despite this, these benchmarks are severely limited in further pushing the frontier of planar object tracking.

\begin{figure}[!t]
    \centering
    \includegraphics[width=\linewidth]{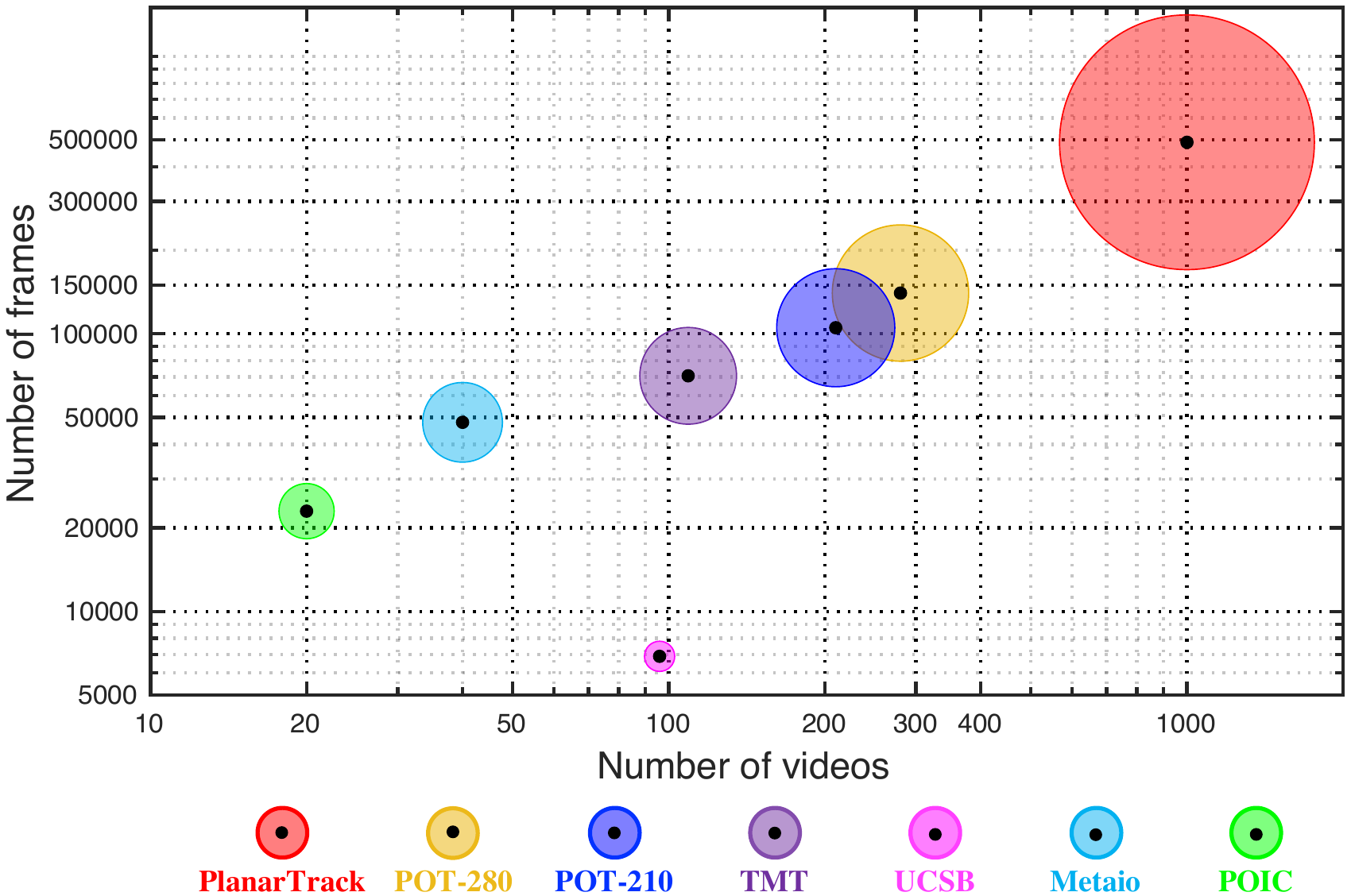}
    \caption{Summary of planar object tracking datasets, containing POT-280~\cite{liang2021planar}, POT-210~\cite{liang2018planar}, TMT~\cite{roy2015tracking}, UCSB~\cite{gauglitz2011evaluation}, Metiao~\cite{lieberknecht2009dataset}, POIC~\cite{chen2017illumination}, and PlanarTrack. The circle diameter is in proportion to the number of frames of a dataset. Our PlanarTrack is the \emph{largest} among all these benchmarks.}
    \label{fig2}
\end{figure}

One of the major issues with existing benchmarks is their relatively small scales. Especially, in the deep learning era, to unleash the potential of deep planar tracking, it is desired to have a large-scale platform. Nevertheless, as displayed in Fig.~\ref{fig2}, currently all planar tracking benchmarks consist of \emph{less than} 300 sequences, which is \emph{insufficient} for large-scale learning of deep planar tracking. As a consequence, researchers are forced to leverage synthetic data generated from images (\eg,~\cite{lin2014microsoft}) for transformation learning in deep planar tracking, which may result in inferior performance due to domain gap between different tasks.

Besides the small-scale issue, another problem is the less challenging scenarios for planar object tracking. Early planar tracking datasets (\eg,~\cite{lieberknecht2009dataset,roy2015tracking,gauglitz2011evaluation,chen2017illumination}) are constructed from the indoor laboratories with simple background, which cannot reflect the diverse and complicated scenarios of real world in performance evaluation. To deal with this, recent datasets (\eg,~\cite{liang2018planar,liang2021planar}) directly collect videos in the wild. However, most of these videos are mainly involved with one challenge factor (or \emph{attribute} in generic tracking), and very few (\eg, 30 in~\cite{liang2018planar} and 40 in~\cite{liang2021planar}) contain multiple challenges (\ie, the unconstrained condition). This may weaken the difficulties of planar tracking in the wild where arbitrary challenges could exist, and thus restricts datasets in assessing generalization of planar tracking in challenging scenes. 

Furthermore, the diversity of current planar object tracking benchmarks is limited. In existing benchmarks, one planar target is usually employed in multiple sequences, which significantly decreases the diversity in target appearance for tracking. Even for current largest benchmark~\cite{liang2021planar} (one target used in 7 videos), the number of planar targets does not exceed 40 (see Tab.~\ref{tab1}). Such lack of diversity makes it difficult to use current datasets for faithful  assessment of planar trackers in practice.

We are aware that there exist several large-scale datasets (\eg,~\cite{muller2018trackingnet,fan2019lasot,huang2019got}) for generic tracking. Nevertheless, due to different setting and goal (see Fig.~\ref{fig1} again), these generic datasets are \emph{not} suitable for planar tracking. To further facilitate research on deep planar tracking, a dedicated large-scale benchmark is desired, which motivates our work.

\subsection{Contributions}

In this paper, we propose a novel large-scale benchmark, dubbed \textbf{PlanarTrack}, dedicated for planar object tracking. Specifically, PlanarTrack consists of 1,000 video sequences. \emph{All} these videos are directly collected in complicated \emph{unconstrained} scenarios from the wild, which makes PlanarTrack, compared to existing datasets (\eg,~\cite{gauglitz2011evaluation,lieberknecht2009dataset,chen2017illumination,roy2015tracking,liang2018planar,liang2021planar}), much more challenging yet realistic for real applications. In order to diversify our PlanarTrack, each planar object appears exclusively in one video, which is different than other datasets. In total, there are over 490K frames in our PlanarTrack, and each one is manually labeled using four corner points\footnote{Four points are the least number of points to determine the homograph of two planar objects, which is the reason to use four points for annotation.} with cautious inspections and refinements to ensure high-qualify annotations. Besides, we offer challenge factor information for each video as in generic tracking~\cite{wu2013online} to enable in-depth analysis. To our best knowledge, PlanarTrack, to date, is the \emph{largest} and \emph{most challenging} planar tracking dataset. By releasing PlanarTrack, we aim to provide a dedicated platform for development and evaluations of planar trackers.

In order to analyze PlanarTrack and provide comparisons for future research, we evaluate 10 representative planar object trackers. Our evaluation exhibits that, not surprisingly, existing top-performing planar trackers severely degrade on more challenging PlanarTrack. For example, the precision (PRE) score (as described later) of WOFT~\cite{vserych2023planar} on POT-210 is 0.805 but drops to 0.433 on PlanarTrack, and the score of HDN~\cite{zhan2022homography} drops from 0.612 on POT-210 to 0.263 on PlanarTrack. This consistently reveals the difficulties for planar tracking brought by realistic complicated scenes, and more efforts are required for improvements. To provide guidance for future research, we further conduct comprehensive analysis to analyze challenges in planar tracking and discuss potential directions to facilitate related research. Besides, our re-training experiments show the usefulness and effectiveness of our benchmark in performance enhancement.

Furthermore, as a by-product of PlanarTrack, we develop a new variant,  \textbf{PlanarTrack}$_{\textrm{BB}}$, which is suitable for generic box tracking. We aim at \emph{large-scale} learning and evaluation of generic object trackers on localizing \emph{rigid} targets, which is rarely investigated before. Our experiments on assessing 10 recent Transformer-based generic trackers reveals heavy performance degeneration on PlanarTrack$_{\textrm{BB}}$ compared with their performance on large-scale generic tracking datasets (\eg, LaSOT~\cite{fan2019lasot} and TrackingNet~\cite{muller2018trackingnet}) and more attention is needed in handling planar objects, though they are rigid. 

\renewcommand\arraystretch{1.0}
\begin{table*}[!tb]
  \centering
  \caption{Detailed comparison of the proposed PlanarTrack with other existing planar object tracking benchmarks.}
    \begin{tabular}{lcccccccccc}
    \specialrule{.1em}{.05em}{.05em} 
    Benchmark & Year  & Targets & Videos & \tabincell{c}{Min\\frames} & \tabincell{c}{Mean\\frames} & \tabincell{c}{Max\\frames} & \tabincell{c}{Total\\frames} & 
    \tabincell{c}{Annotated\\frames} &\tabincell{c}{Unconstrain-\\ed Videos} & \tabincell{c}{In the\\wild} \\
    \hline\hline
    Metaio~\cite{lieberknecht2009dataset} & 2009  & 8     & 40    & 1,200  & 1,200  & 1,200  & 48K &48K  & n/a       & \xmark \\
    UCSB~\cite{gauglitz2011evaluation}  & 2011  & 6     & 96    & 13    & 72    & 500   & 7K &7K   & n/a     & \xmark \\
    TMT~\cite{roy2015tracking}   & 2015  & 12    & 109   & 191   & 648   & 2,518  & 71K &71K  & n/a      & \xmark \\
    POIC~\cite{chen2017illumination}  & 2017  & 20    & 20    & 283   & 1,149  & 2,666  & 23K &23K  & n/a    & \xmark \\
    POT-210~\cite{liang2018planar} & 2018  & 30    & 210   & 501   & 501   & 501   & 105K &53K  & 30       & \cmark \\
    POT-280~\cite{liang2021planar} & 2021  & 40    & 280   & 501   & 501   & 501   & 140K &70K & 40      & \cmark \\
    \hline
    \rowcolor{gray!15} \textbf{PlanarTrack} & \textbf{2023}  & \textbf{1,000}  & \textbf{1,000}  & \textbf{317}   & \textbf{490}   & \textbf{549}   & \textbf{490K} & \textbf{490K}  & \textbf{1,000}    & \cmark \\
    \specialrule{.1em}{.05em}{.05em} 
    \end{tabular}%
  \label{tab1}%
\end{table*}%

In summary, our main contributions are as follows:
\vspace{-0.35em}
\begin{itemize}
\setlength{\itemsep}{2pt}
\setlength{\parsep}{2pt}
\setlength{\parskip}{2pt}

      \item[$\diamond$] \emph{We introduce a novel benchmark termed PlanarTrack for planar tracking. To the best of our knowledge, PlanarTrack is to date the largest as well as the most challenging planar tracking benchmark in the wild.
      }
      
      \item[$\diamond$] \emph{We conduct comprehensive evaluations to analyze PlanarTrack and provide comparison for future research.}
      

      \item[$\diamond$] \emph{We conduct retraining experiments to validate the effectiveness of the proposed PlanarTrack in improving deep planar tracking performance.}
      
      \item[$\diamond$] \emph{Based on PlanarTrack, we develop PlanarTrack$_{\mathrm{BB}}$ for generic tracking on planar-like targets and conduct extensive evaluation and analysis.}
\end{itemize}


\section{Related Work}

\subsection{Planar Tracking Benchmarks}

Datasets have played an important role in facilitating the development of planar object tracking. \textbf{Metaio}~\cite{lieberknecht2009dataset} is one of the earliest datasets for planar tracking. It comprises 40 videos with eight different textures using a camera mounted on the robotic measurement arm. \textbf{UCSB}~\cite{gauglitz2011evaluation} contains 96 videos for investigating interest point detectors and feature descriptors for planar object tracking. \textbf{TMT}~\cite{roy2015tracking} consists of 109 videos and each one is labeled with a challenging factor. The goal is to evaluate different planar tracking algorithms for human and robot manipulation tasks. \textbf{POIC}~\cite{chen2017illumination} provides 20 sequences and mainly focuses on evaluating the performance of planar trackers in complicated illumination environments. In order to assess the planar tracking performance in the wild, \textbf{POT-210}~\cite{liang2018planar} collects 210 videos of 30 planar objects from natural scenarios. Later in~\cite{liang2021planar}, POT-210 is further extended to \textbf{POT-280} by introducing 70 extra videos of 10 planar targets. For each planar object in POT~\cite{liang2018planar,liang2021planar}, seven videos are captured, however, six of them simply comprise one challenge and only one contains multiple challenges in unconstrained conditions.

Despite the above benchmarks, the further development of planar object tracking, especially in the deep learning, is limited due to lacking a large-scale, challenging and diverse platform, which motivates our PlanarTrack, the \emph{largest} and most \emph{challenging} and \emph{diverse} planar tracking benchmark to date. Tab.~\ref{tab1} displays a detailed comparison of PlanarTrack with existing planar tracking benchmarks.

\subsection{Planar Tracking Algorithms}

The goal of planar tracking is to estimate the homograph. Current approaches can be roughly divided into three types: keypoint methods, direct method and deep regression methods. Keypoint-based planar trackers (\eg,~\cite{dick2013realtime,ozuysal2009fast,wang2017gracker}) first detect the keypoints (\eg, SIFT~\cite{lowe2004distinctive} or SURF~\cite{bay2006surf}) of objects and then estimate homograph using these interesting points. Direct methods~\cite{benhimane2004real,richa2011visual,chen2017illumination} aim to directly calculate the homograph by optimizing the alignment of current frame with object of initial frame. In addition to the above two types, another recent trend is to employ the deep neural networks to regress the homograph. These deep regression-based planar trackers~\cite{zhan2022homography,zhang2022hvc,vserych2023planar} avoid complex keypoint feature extraction and can be trained in an end-to-end fashion. Due to outstanding performance, the deep regression-based methods have attracted increasing attentions in planar tracking.

\subsection{Large-scale Tracking Benchmarks}

Large-scale benchmarks have recently greatly facilitated the development of tracking. Representatives include GOT-10k~\cite{huang2019got}, LaSOT~\cite{fan2019lasot,fan2021lasot}, TrackingNet~\cite{muller2018trackingnet}, OxUvA~\cite{valmadre2018long}, and TNL2K~\cite{wang2021towards}. \textbf{GOT-10k} consists of 10K videos with various motion patters for short-term object tracking. \textbf{LaSOT} offers 1,400 videos for long-term tracking, and is later extended by providing 150 extra sequences. \textbf{TrackingNet} comprises more than 30K videos for training of deep trackers. \textbf{OxUvA} contains 366 long videos for long-term performance evaluation. \textbf{TNL2K} consists of 2,000 videos with box and language annotations for vision-language tracking. 

Different from the above benchmarks, the proposed PlanarTrack is specially developed for planar object tracking. For this goal, we provide annotations of corner points in PlanarTrack for targets instead of axis-aligned rectangular bounding boxes in aforementioned datasets.

\begin{figure*}[!t]
		\centering
		\begin{tabular}{c@{\hspace{1.8mm}}c}
\includegraphics[width=2.7cm]{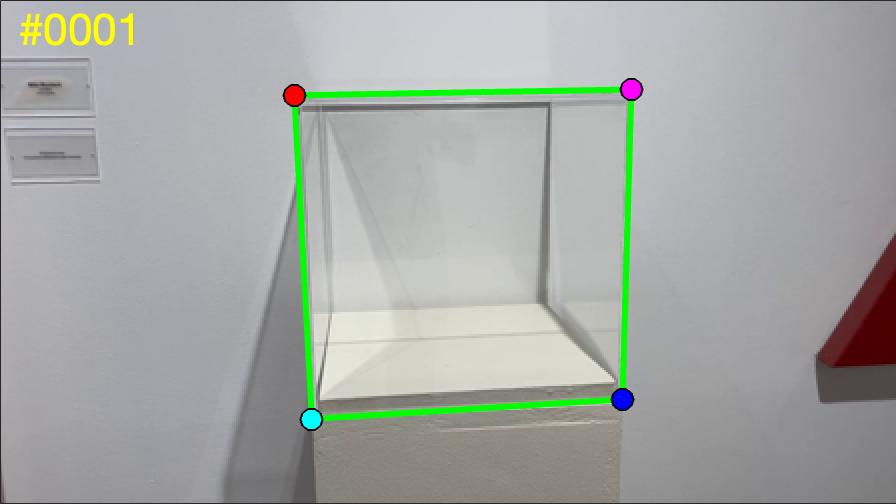} \includegraphics[width=2.7cm]{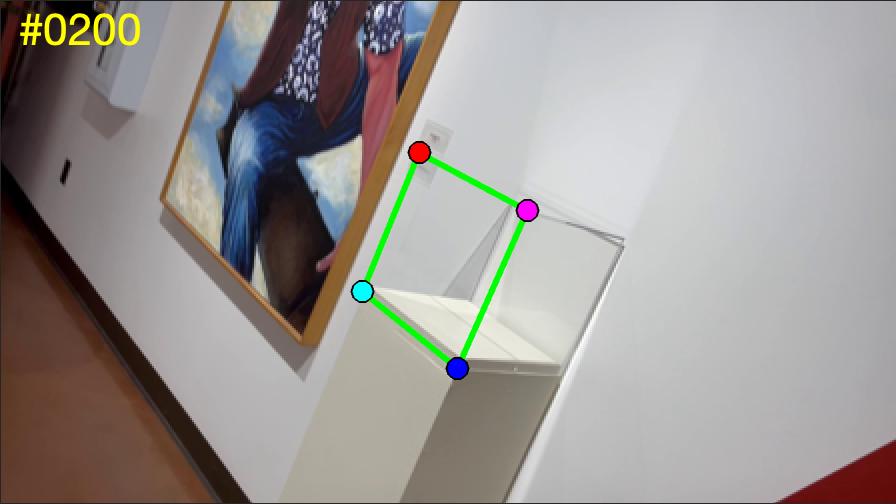} \includegraphics[width=2.7cm]{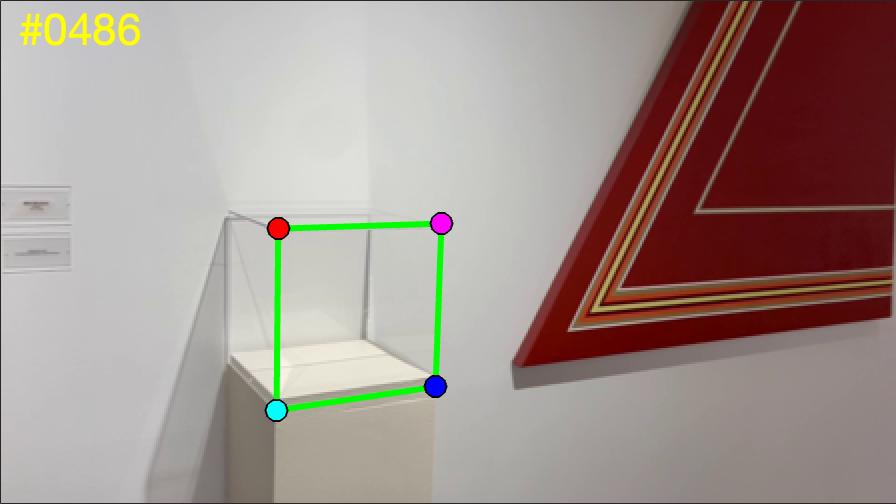} & \includegraphics[width=2.7cm]{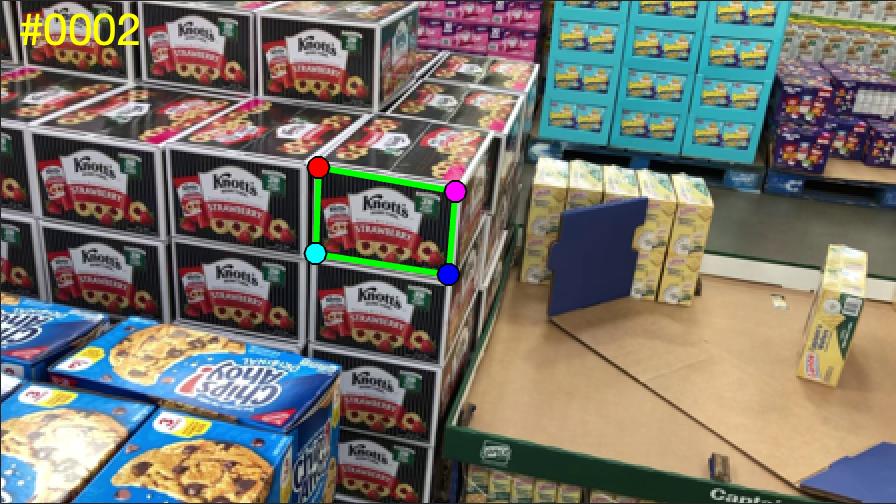}  \includegraphics[width=2.7cm]{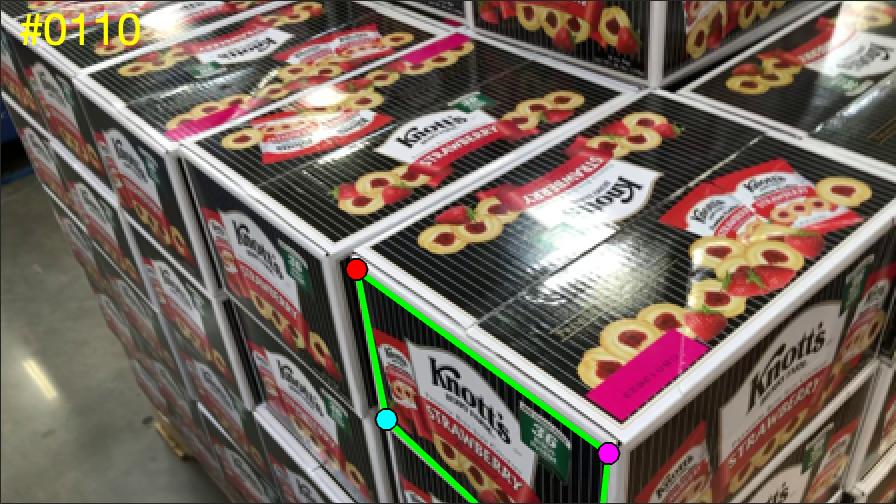} \includegraphics[width=2.7cm]{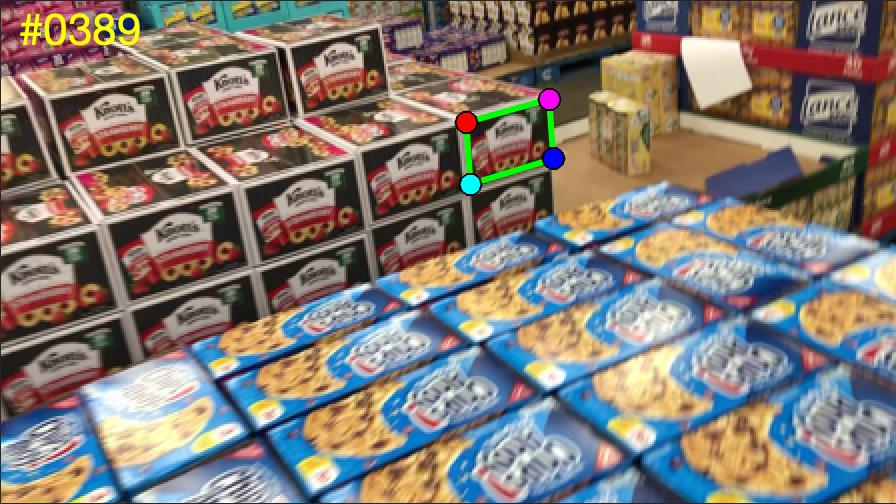} \\
\includegraphics[width=2.7cm]{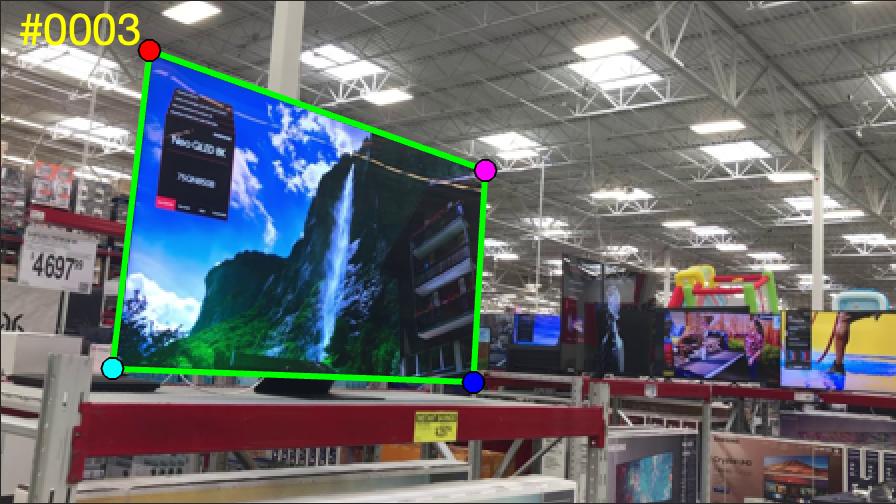} \includegraphics[width=2.7cm]{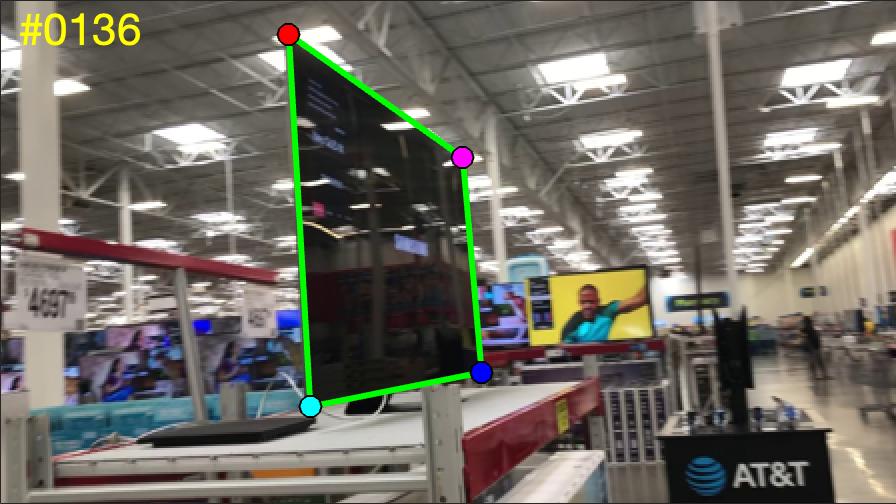} \includegraphics[width=2.7cm]{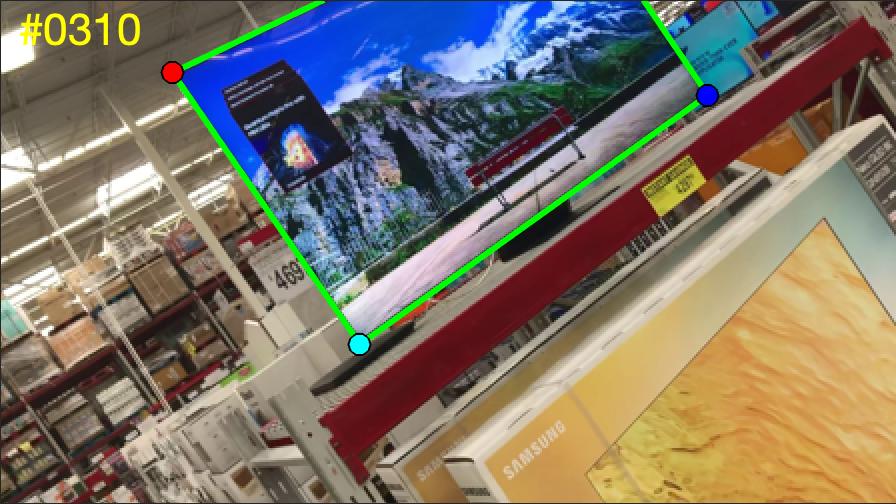}  &
\includegraphics[width=2.7cm]{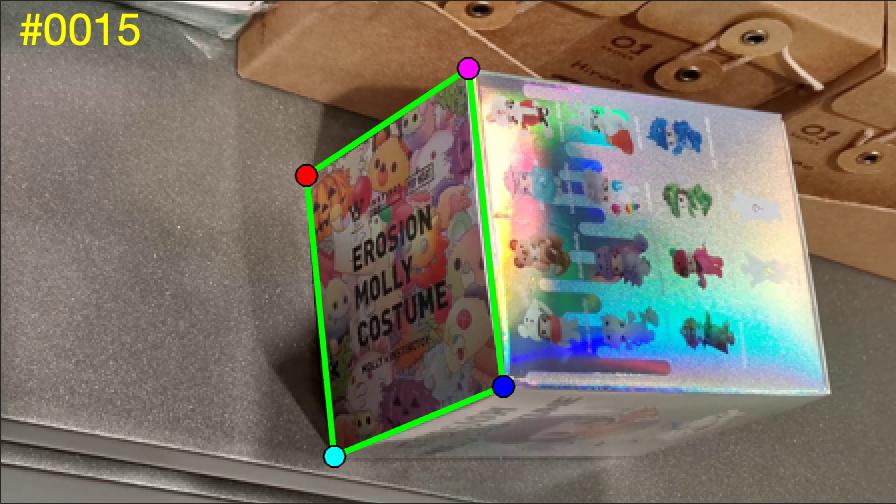}  \includegraphics[width=2.7cm]{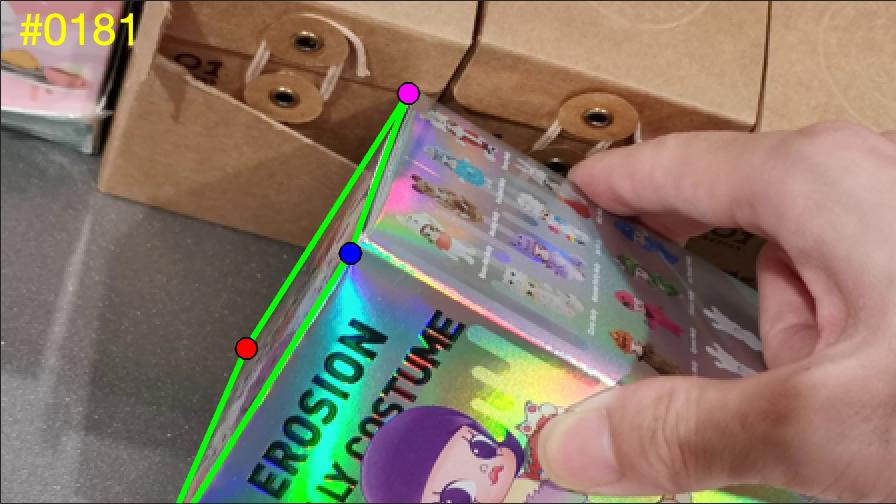} \includegraphics[width=2.7cm]{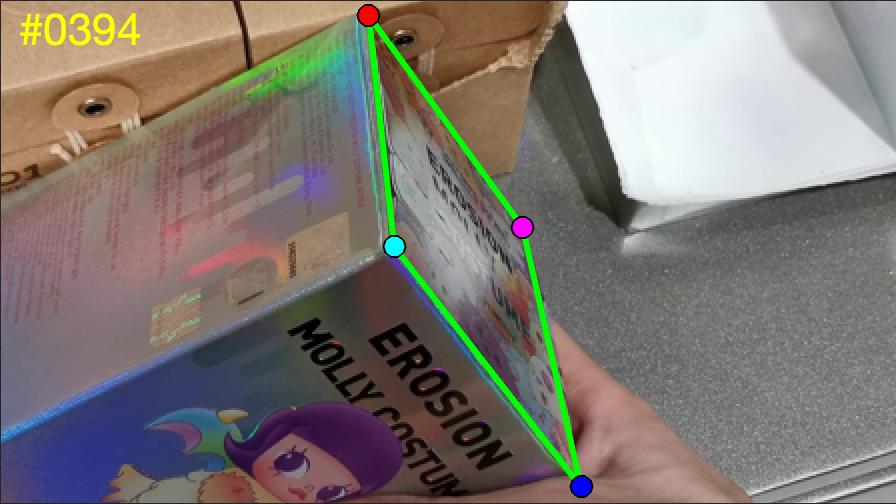} \\
\includegraphics[width=1.585cm]{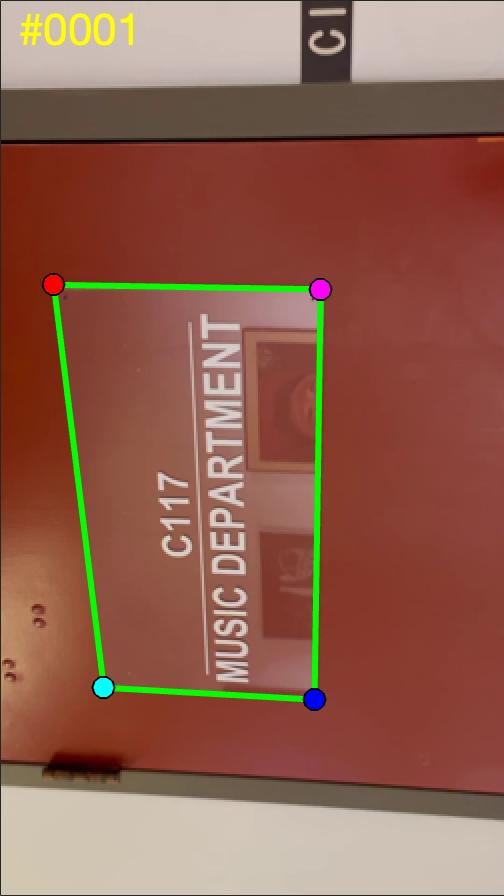}
\includegraphics[width=1.585cm]{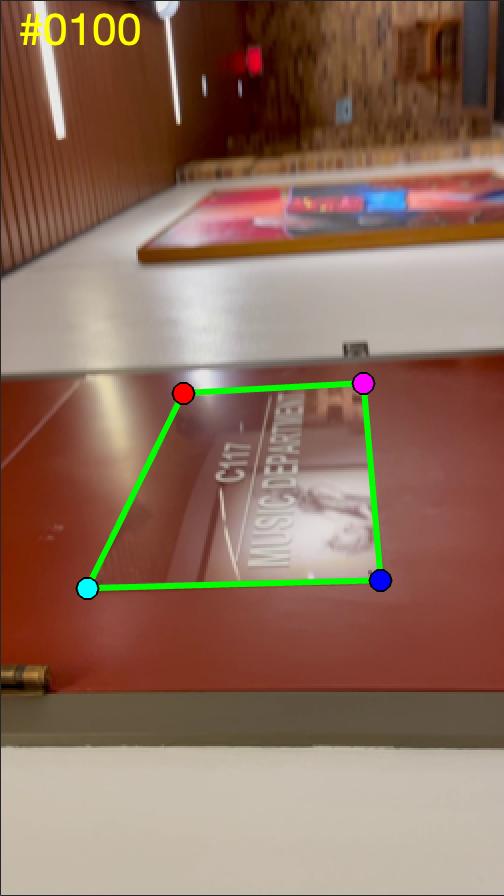}
\includegraphics[width=1.585cm]{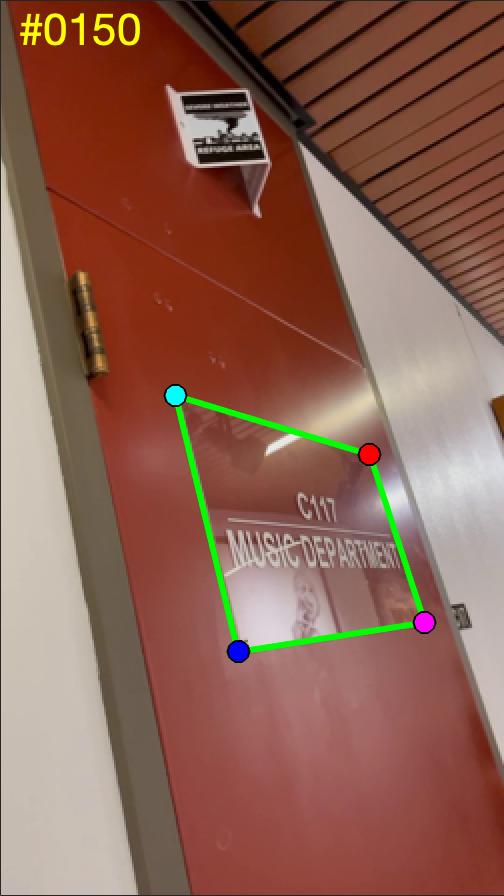}
\includegraphics[width=1.585cm]{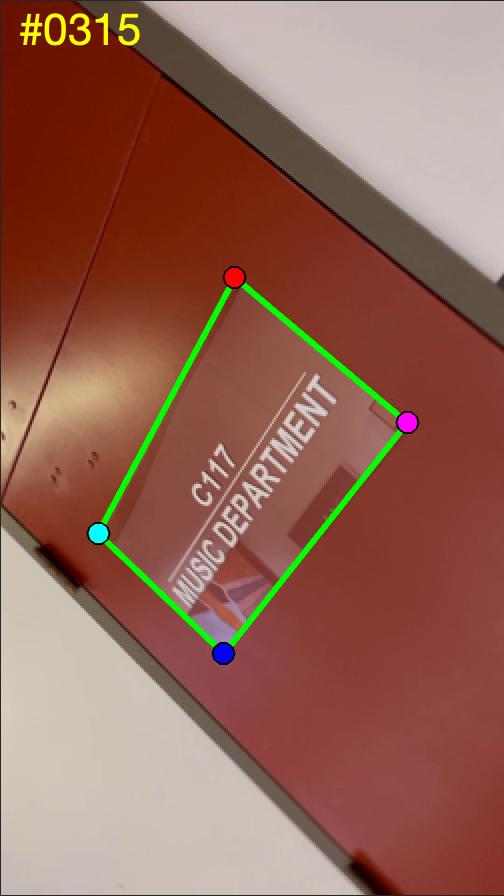}
\includegraphics[width=1.585cm]{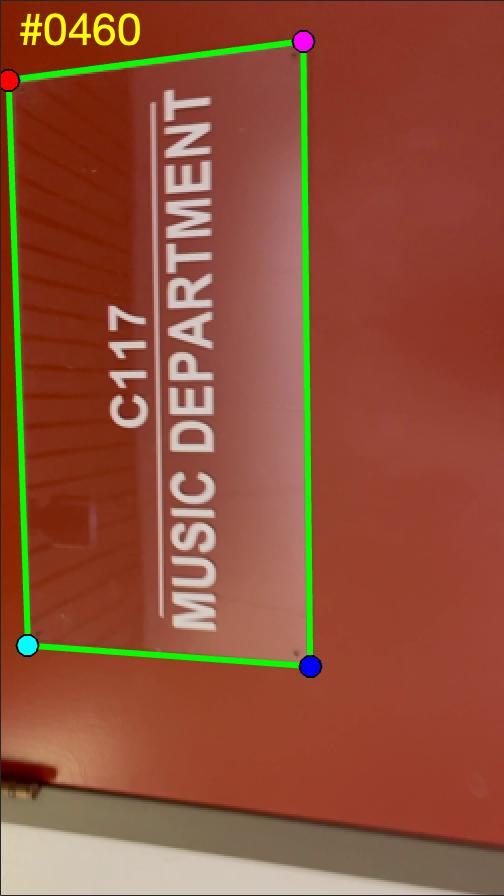}&
\includegraphics[width=1.585cm]{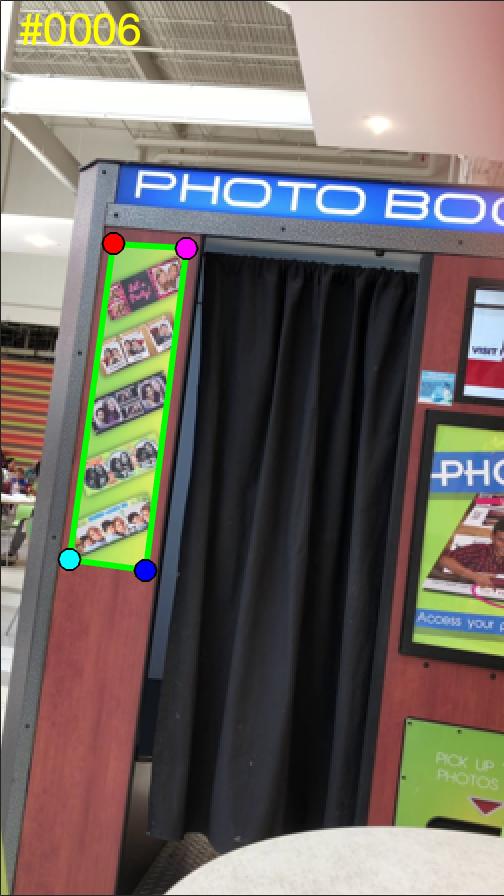}
\includegraphics[width=1.585cm]{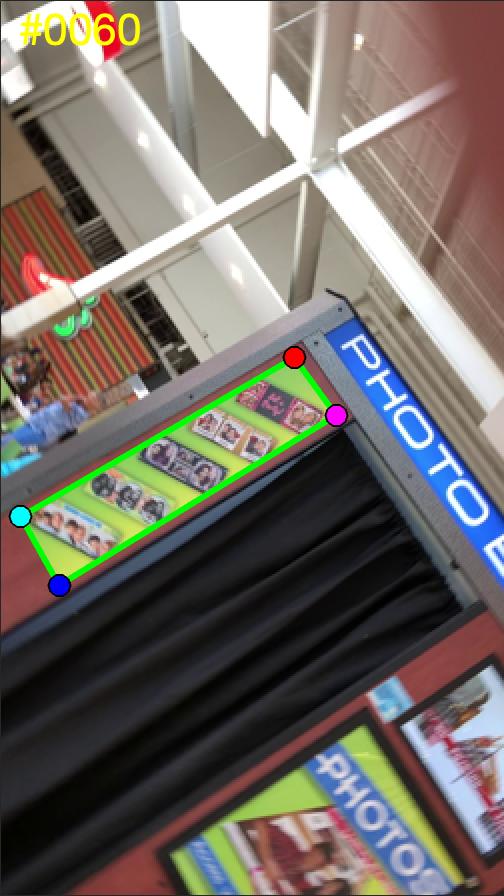}
\includegraphics[width=1.585cm]{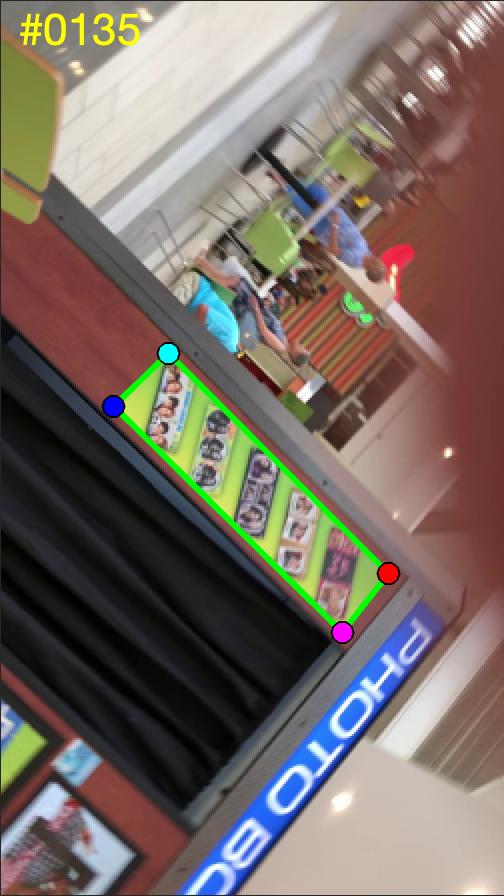}
\includegraphics[width=1.585cm]{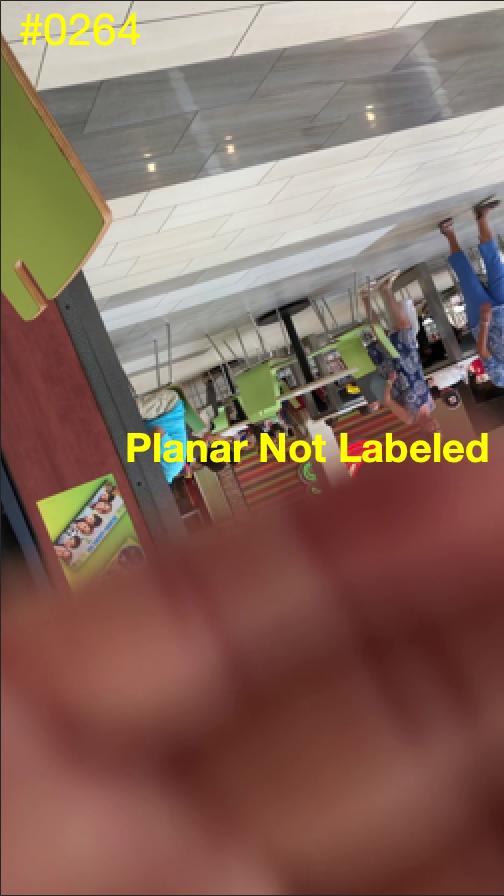}
\includegraphics[width=1.585cm]{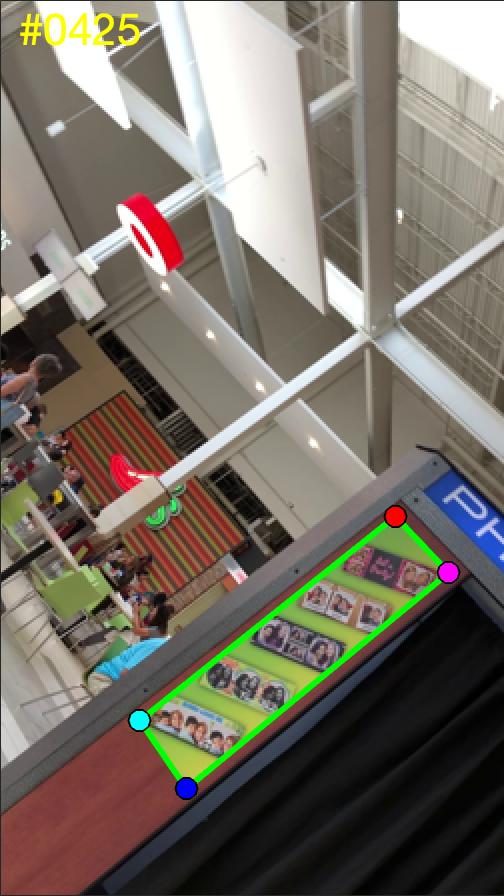}\\
\multicolumn{2}{c}{\includegraphics[width=11cm]{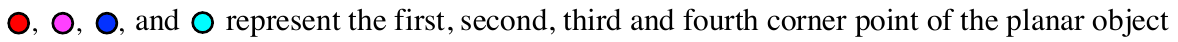}}\\
\end{tabular}
\caption{Examples of annotated sequences in the proposed PlanarTrack. Each video is annotated with four corner points.
}
\label{anno_exa}
\end{figure*}

\section{The Proposed PlanarTrack Benchmark}

\subsection{Design Principle}

PlanarTrack in this work expects to provide a large-scale platform for developing deep planar tracking and to offer a more challenging and faithful testbed for evaluating planar trackers in practice. To meet these requirements, we follow four rules in constructing our PlanarTrack:

\vspace{-0.35em}
\begin{itemize}
\setlength{\itemsep}{2pt}
\setlength{\parsep}{2pt}
\setlength{\parskip}{2pt}

      \item \emph{Dedicated large-scale benchmark.} One important motivation for our work is to facilitate deep planar tracking with a large-scale dedicate benchmark. To this end, we hope to collect 1,000 videos with over 450K frames in the new benchmark.

      \item \emph{Realistic challenge in the wild.} To faithfully reflect the performance of planar trackers in practice, it is crucial to collect videos with realistic challenges. For this purpose, we require all videos in the benchmark captured from natural scenarios in unconstrained conditions.

      \item \emph{Diverse planar objects.} The diversity of targets is beneficial for assessing the generalization of planar trackers. Considering this, the planar targets in the videos should be unique, which differs from current datasets.

      \item \emph{High-quality dense annotation.} The annotation is crucial for both training and evaluation. For this, we manually label every frame in PlanarTrack with careful refinement to ensure its high-quality annotations.
      
\end{itemize}

\subsection{Video Collection}

We construct PlanarTrack starting by collecting videos. Different from generic object tracking benchmarks (\eg, ~\cite{fan2019lasot,huang2019got,muller2018trackingnet}) sourcing videos from YouTube, we record sequences from natural scenarios using smart phones as we observe the videos from YouTube seldom focus on the motion of planar objects. To diversify the video sources, we invite volunteers who are familiar with this task to record the sequences using different phones with different resolutions. With the above principles in mind, we include a wide selection of the planar targets (\eg, \emph{box}, \emph{poster}, \emph{picture}, \emph{board}, \emph{logo}, \emph{door}, \emph{mirror}, \emph{book}, \emph{traffic sign}, \emph{tile}, \emph{wall}, \emph{tile}, \emph{screen}, and \emph{table}) for video recording, and each sequence is captured in unconstrained conditions from various natural scenes (\eg, \emph{shopping mall}, \emph{street}, \emph{library}, \emph{restaurant}, \emph{supermarket}, \emph{playground}, \emph{park}, \emph{museum}, \emph{apartment}, \emph{hall}, and \emph{classroom}).

Initially, we collected over 2,500 videos. After a careful inspection conducted by a few experts (PhD students working on related topics), we choose 1,000 available videos for developing PlanarTrack. It is worth noticing that, for these 1,000 videos, we further verify their contents and remove inappropriate parts to make sure they are suitable for planar tracking. Eventually, we compile a large-scale challenging benchmark dedicated for planar tracking by including 1,000 unconstrained sequences with more than 490K frames from 1,000 unique planar objects. Tab.~\ref{tab1} provides a detailed summary of PlanarTrack and its comparison with existing planar tracking benchmarks.

\subsection{Annotation}

To offer high-quality annotation in PlanarTrack, we manually label each frame. Specifically, for each image, we annotate four corner points for the planar target if all its four corner points or four edges are clearly visible to. Otherwise, if the four corner points and four edges are both not available due to occlusion or out-of-view, or, the planar target is severely blurred, we will assign an absent flag to this frame. 

With the above strategy, we assemble a team with several experts and volunteers for annotation. Each sequence is first annotated by a volunteer. Then, the annotation result will be sent to two experts for verification. If the annotation is not unanimously agreed by the experts, it will be returned back the original annotator for careful refinement. To ensure the high annotation quality, the verification-refinement process may last for multiple rounds until the final annotation result passes the inspection. We demonstrate some annotation examples of PlanarTrack in Fig.~\ref{anno_exa}.

\begin{figure}[!t]
    \centering
    \includegraphics[width=\linewidth]{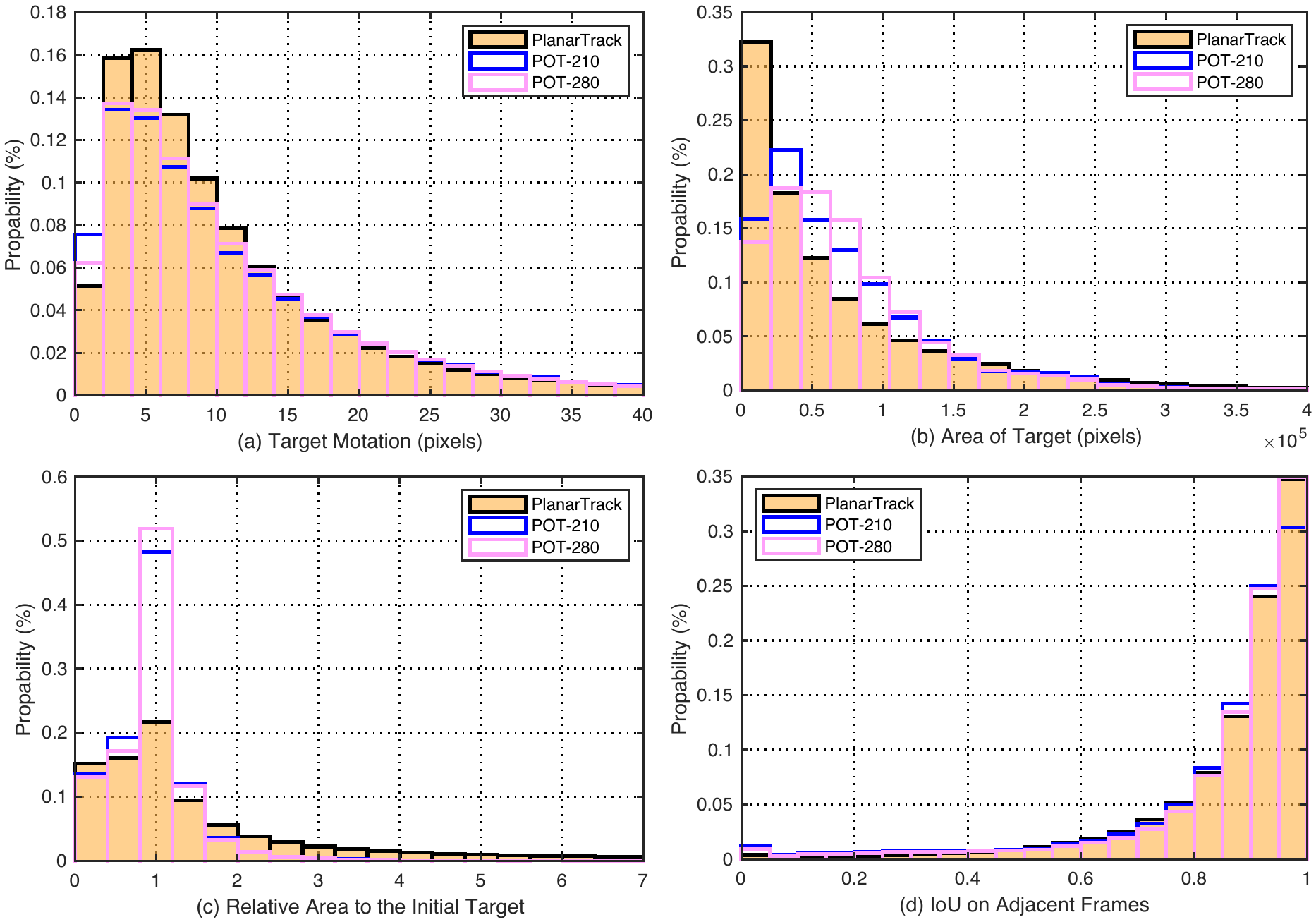}
    \caption{Statistics of planar target motion, size, relative area compared to initial object and IoU of targets in adjacent frames in PlanarTrack and comparison with the recent POT-210/280~\cite{liang2018planar,liang2021planar}. We can see the targets in our dataset have smaller sizes and faster and more challenging motions.}
    \label{fig4}
\end{figure}

\vspace{0.3em}
\noindent
\textbf{Statistics of annotations.} In order to better understand the planar targets in PlanarTrack, we show representative statistics of the annotations in Fig.~\ref{fig4}. In particular, we display the distributions of target motion, target size, relative area to the initial object and Intersection over Union (IoU) between targets in adjacent frames. From Fig.~\ref{fig4}, we see that the planar targets vary rapidly in size and temporal motions. Besides, Fig.~\ref{fig4} also compares our PlanarTrack and the recent POT-210/280~\cite{liang2018planar,liang2021planar}. Notice that, since POT-210/280 are labeled every two frames, we perform linear interpolation on their annotation for the comparison purpose. From Fig.~\ref{fig4}, we can see that the targets in PlanarTrack are relatively smaller and moving faster, which consequently leads to new challenges for planar tracking in the wild.

\subsection{Challenging Factors}

Following other tracking datasets~\cite{wu2013online,liang2018planar,fan2021transparent}, we provide challenging factors (also called \emph{attributes} in other datasets) for each sequence in PlanarTrack to enable further in-depth analysis of different algorithms. In specific, we define eight challenging factors that widely exist for planar tracking and annotate each sequence with these factors, including (1) occlusion (OCC), (2) motion blur (MB), (3) rotation (ROT), (4) scale variation (SV), which is assigned when the ratio of planar annotation is outside the range [0.5, 2], (5) perspective distortion (PD), which is assigned when the perspective between the object and camera is changed, (6) out-of-view (OV), (7) low resolution (LR), which is assigned when the region of the target planar is less than 1,000 pixels, and (8) background clutter (BC), which is assigned when the background region looks visually similar to the target. It is worthy to note that, we exclude a few common challenging factors used in generic object tracking such as deformation and illumination change because they are not suitable for planar targets. Each video in PlanarTrack may simultaneously contain multiple challenging factors (\ie, recorded in \emph{unconstrained condition}), which is, compared to POT-210/280, more challenging and practical for real applications. 

\begin{figure}[!t]
    \centering
    \includegraphics[width=\linewidth]{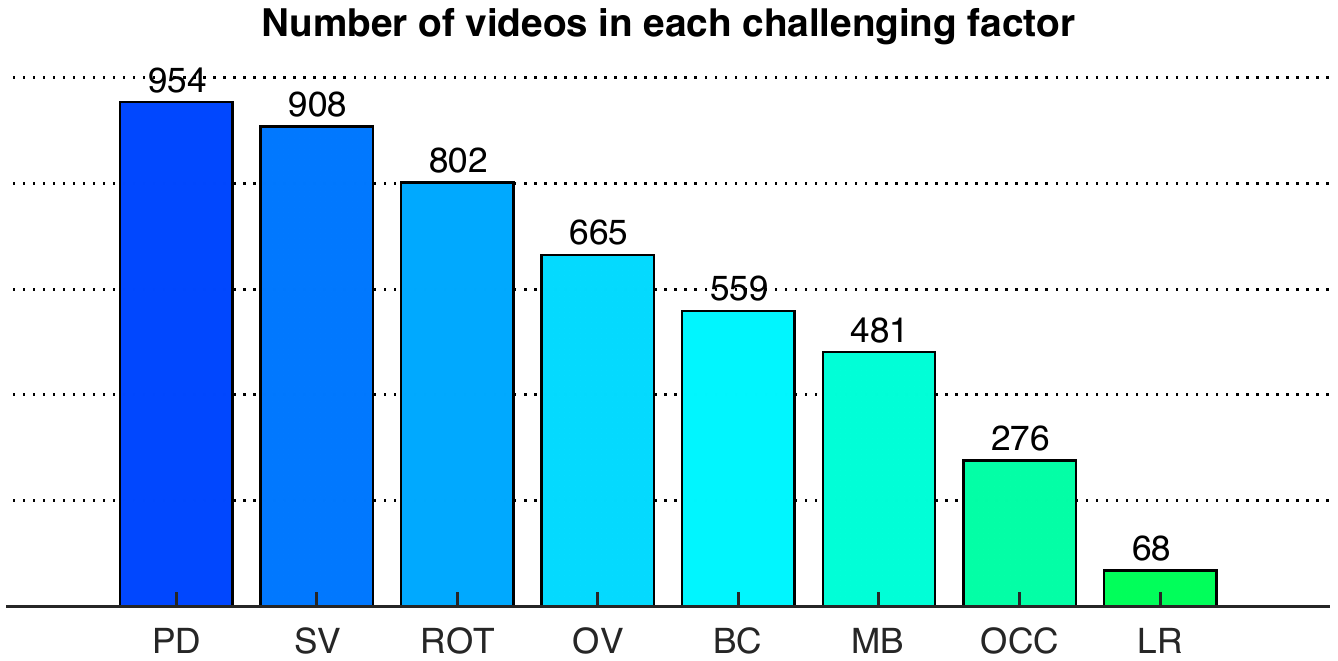}
    \caption{Distribution of sequences on each challenging factor.}
    \label{fig5}
\end{figure}

The distribution of the aforementioned challenging factors on PlanarTrack is presented in Fig.~\ref{fig5}. We observe that the most common challenging factor in PlanarTrack is perspective distortion, which may cause serious misalignment problem for planar tracking. In addition, scale variation and rotation frequently happen in the sequences. 

\subsection{Dataset Split and Evaluation Metric}

\begin{table}[!tb]\small
  \centering
  \caption{Comparison of \emph{training} and \emph{test} sets.}
  \begin{tabular}{@{}L{2.15cm}@{}ccccc@{}}
  \specialrule{.1em}{.05em}{.05em} 
    & Videos & \tabincell{c}{Min \\ frames} & \tabincell{c}{Mean \\frames} & \tabincell{c}{Max \\frames} & \tabincell{c}{Total\\ frames} \\
    \hline\hline
    PlanarTrack$_\mathrm{Tst}$ & 300    & 346    & 493    & 534    & 148K \\
    PlanarTrack$_\mathrm{Tra}$ & 700    & 317    & 489    & 549    & 342K \\
    \specialrule{.1em}{.05em}{.05em} 
    \end{tabular}%
  \label{tab2}%
\end{table}%

\noindent
{\textbf{Training/Test Split.}} PlanarTrack consists of 1,000 videos. We use 700 sequences for training (PlanarTrack$_\mathrm{Tra}$) and the rest 300 for evaluation (PlanarTrack$_\mathrm{Tst}$). We try our best to keep the distributions of training and test sets close to each other. Tab.~\ref{tab2} shows the comparison of these two sets, and please refer to \emph{supplementary material} for challenge-wise comparisons. The detailed split will be released at our project website.

\vspace{0.3em}
\noindent
{\textbf{Evaluation Metric.}} For the evaluation, we follow~\cite{liang2018planar} and adopt the \emph{precision} (PRE) and \emph{success} (SUC) metrics. It is worthy to notice, the PRE and SUC differ from those used for generic tracking~\cite{wu2013online}. Specifically, for planar tracking, the PRE is defined as the percentage of frames where alignment error between the corner points of tracking result and groundtruth is within a given threshold (\eg, typically 5 pixels). The SUC is calculated by the percentage of successful frames in which the discrepancy between estimated and real homography is smaller than or equal to a certain threshold. We set the threshold to 30 in our evaluation as the threshold of 10 in~\cite{liang2018planar} is too tight. For more details of PRE and SUC for planar tracking evaluation, please kindly refer to~\cite{liang2018planar}.

\section{Experiments on PlanarTrack}

\subsection{Evaluated Planar Trackers}

Since there are not many planar object trackers compared to generic tracking (in fact, it motivates us to introduce PlanarTrack for fostering research on planar object tracking), we select 10 representative algorithms with available source codes consisting of two very recent ones. Specifically, these trackers are Gracker~\cite{wang2017gracker}, GIFT~\cite{liu2019gift}, 
ESM~\cite{benhimane2004real},
LISRD~\cite{pautrat2020online}, SOL~\cite{hare2012efficient},
SIFT~\cite{lowe2004distinctive}, 
IC~\cite{baker2004lucas},
SCV~\cite{richa2011visual}, HDN~\cite{zhan2022homography}, and WOFT~\cite{vserych2023planar}.  Particularly, the HDN~\cite{zhan2022homography} and WOFT~\cite{vserych2023planar} are two recently specially developed planar trackers using deep learning. Notice that, we do not evaluate generic trackers on our PlanarTrack due to incompatible inputs and tracking results. Instead, we will create a new PlanarTrack$_{\mathrm{BB}}$ suitable for generic tracking evaluation, as described later.

\begin{figure}[!t]
    \centering
    \includegraphics[width=\linewidth]{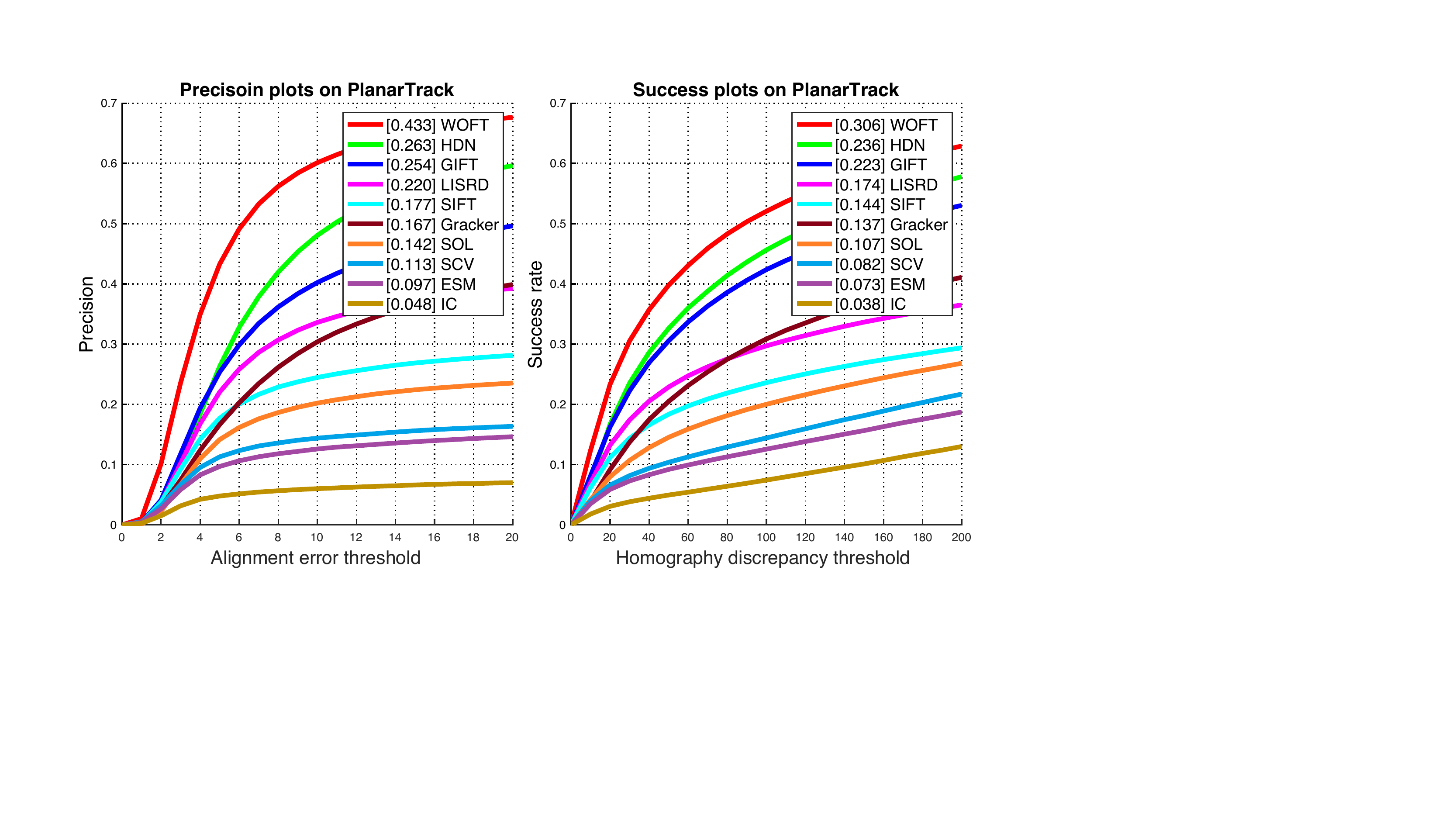}
    \caption{Overall performance on PlanarTrack$_\textbf{Tst}$ in terms of precision (left) and success (right). }
    \label{fig6}
\end{figure}

\subsection{Evaluation Results}

\noindent
\textbf{Overall Performance.} We evaluate 10 typical planar object trackers on the test set of PlanarTrack. Please note that, the methods of HDN and WOFT are utilized without modifications in our evaluation as they are specifically developed for the planar tracking task. For all other approaches, they are customized to achieve the planar tracking. Their implementations except for LISRD and GIFT are borrowed from~\cite{liang2018planar}, and we adapt LISRD and GIFT to planar tracking because of some setting problems provided by~\cite{liang2018planar}. The evaluation results of these approaches are reported in Fig.~\ref{fig6} using precision (PRE) and success (SUC). From Fig.~\ref{fig6}, we can observe that WOFT demonstrates the best PRE score of 0.433 and SUC score of 0.306, and HDN shows the second best PRE score of 0.263 and SUC score of 0.236. Both WOFT and HDN are recent planar trackers which formulate planar tracking as a deep homography estimation problem. Compared with HDN, WOFT introduces the optical flow into homography estimation and effectively boosts the robustness of tracking, which exhibits the importance of video temporal information for tracking. The method of GIFT applies transformation-invariant deep visual descriptors for planar tracking and achieves the third best of PRE score of 0.254 and SUC score of 0.233. It is worth mentioning that, all the top four trackers leverage deep neural networks for planar target localization, which demonstrates the great potential of deep planar tracking in the future. This is also the motivation of our work to offer a dedicated large-scale platform for developing deep planar trackers. 

\begin{figure}[!t]
    \centering
    \includegraphics[width=\linewidth]{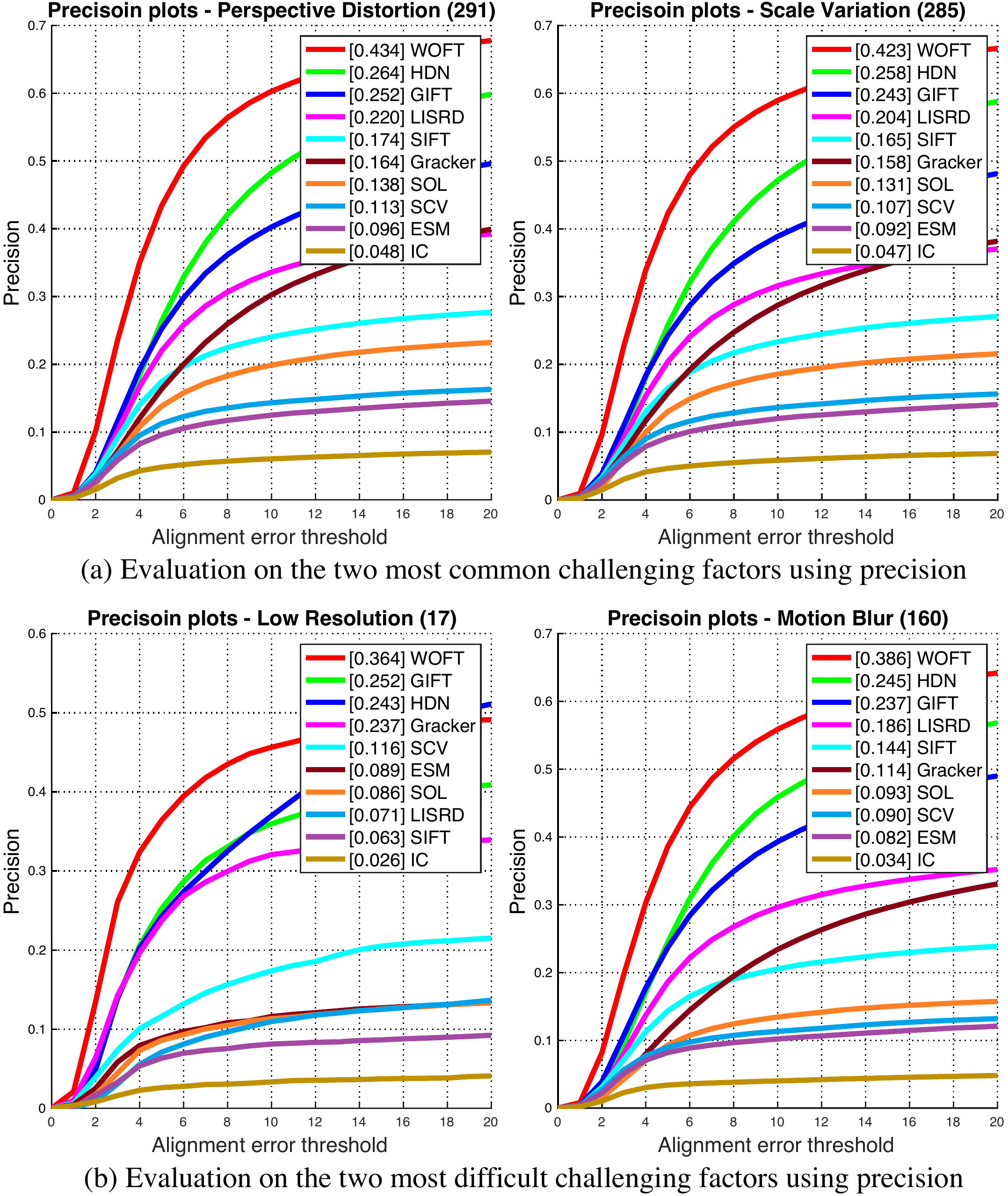}
    \caption{Performance evaluation of trackers on the two most common challenging factors including \emph{perspective distortion} and \emph{scale variation} and on the two most difficult challenging factors including \emph{low resolution} and \emph{motion blur} using precision (Please refer to \emph{supplementary material} for full results and comparisons). }
    \label{fig7}
\end{figure}

\begin{figure*}[!t]
		\centering
		\begin{tabular}{c@{\hspace{1.8mm}}c}
\includegraphics[width=2.7cm]{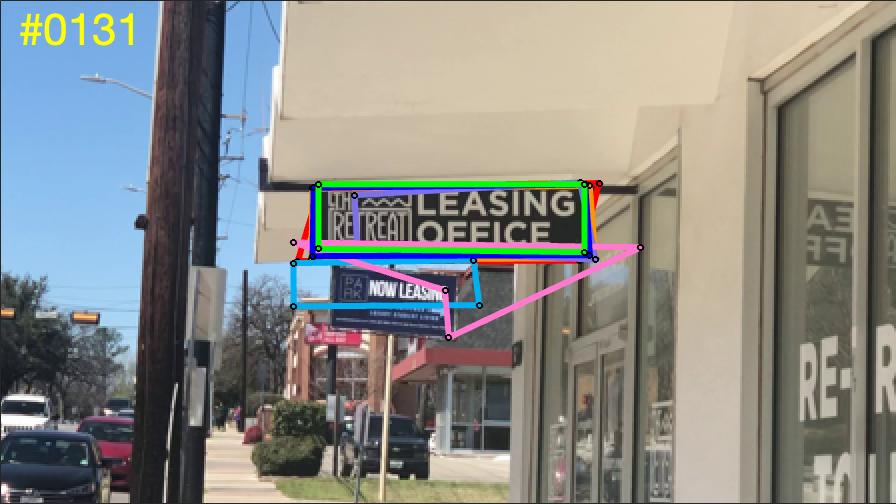} \includegraphics[width=2.7cm]{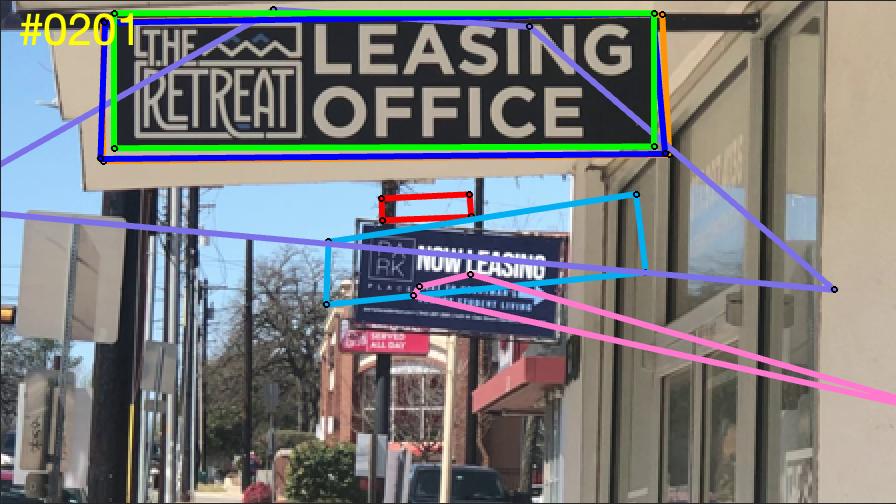} \includegraphics[width=2.7cm]{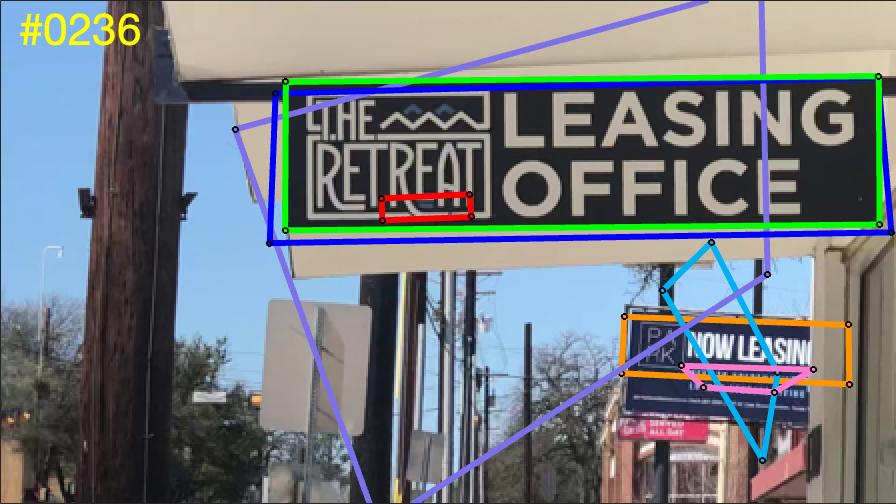} & \includegraphics[width=2.7cm]{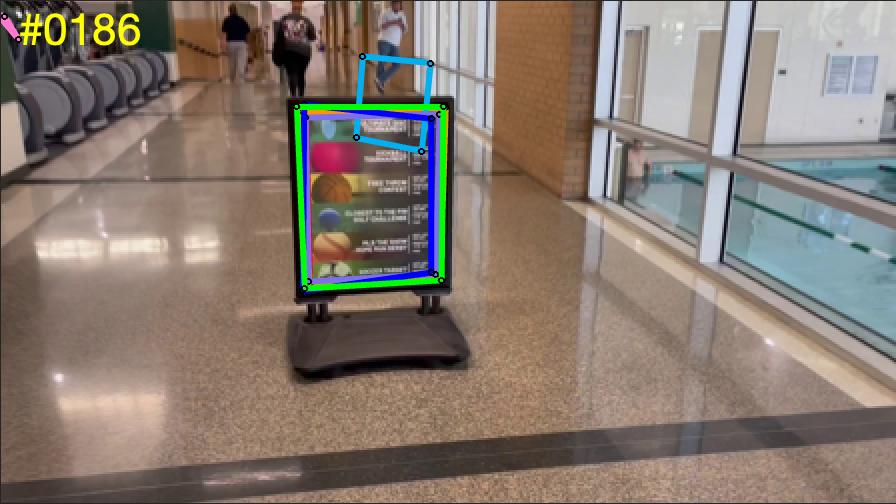}  \includegraphics[width=2.7cm]{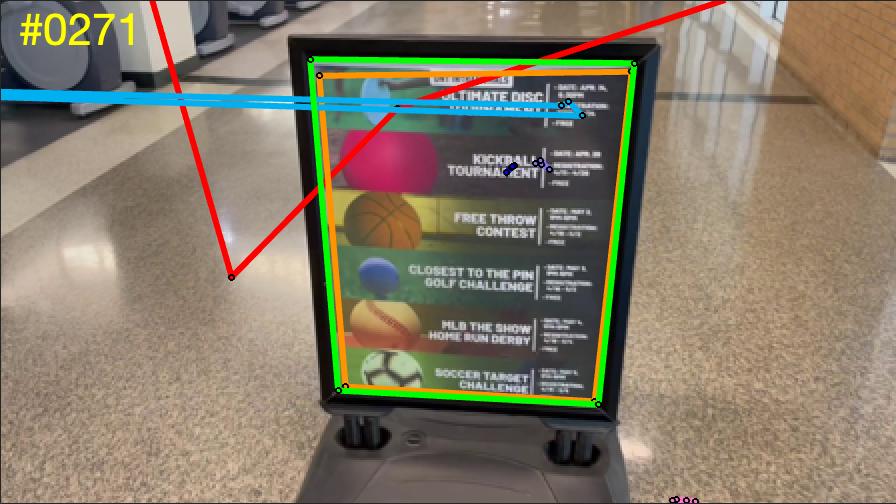} \includegraphics[width=2.7cm]{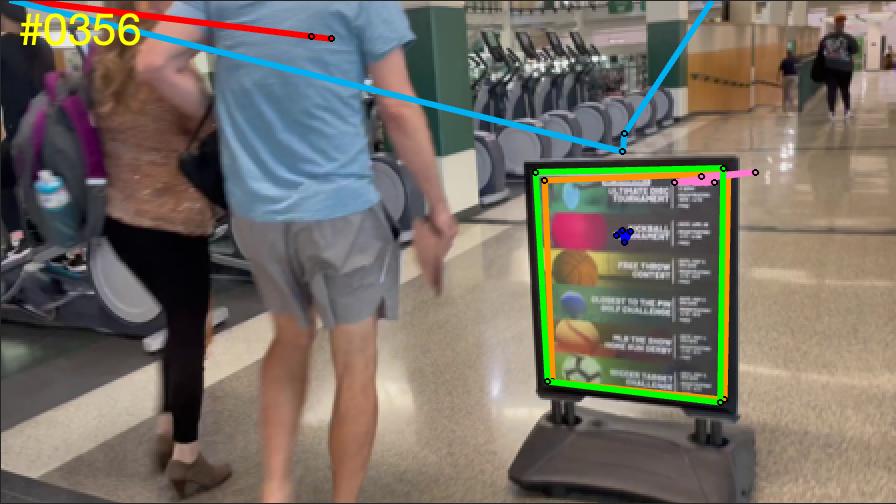} \\
{\small (a) Sequence with background clutter and scale variation}& {\small (b) Sequence with scale variation and perspective distortion} \\
\includegraphics[width=2.7cm]{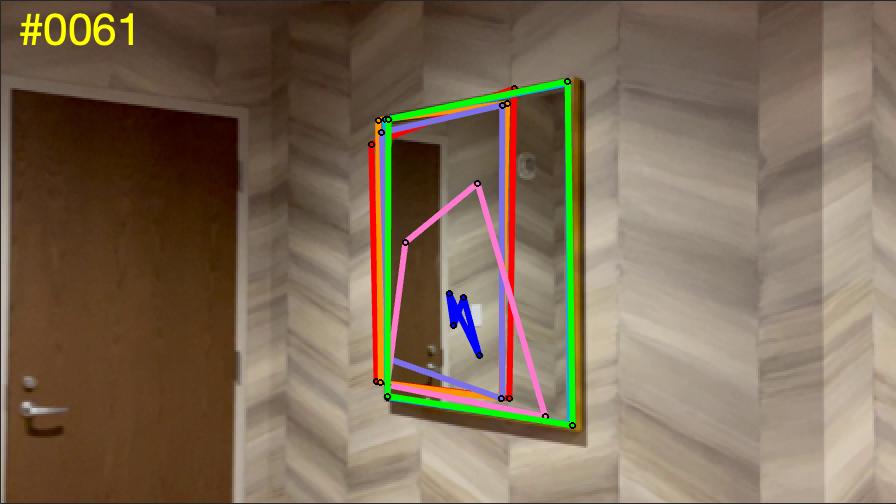} \includegraphics[width=2.7cm]{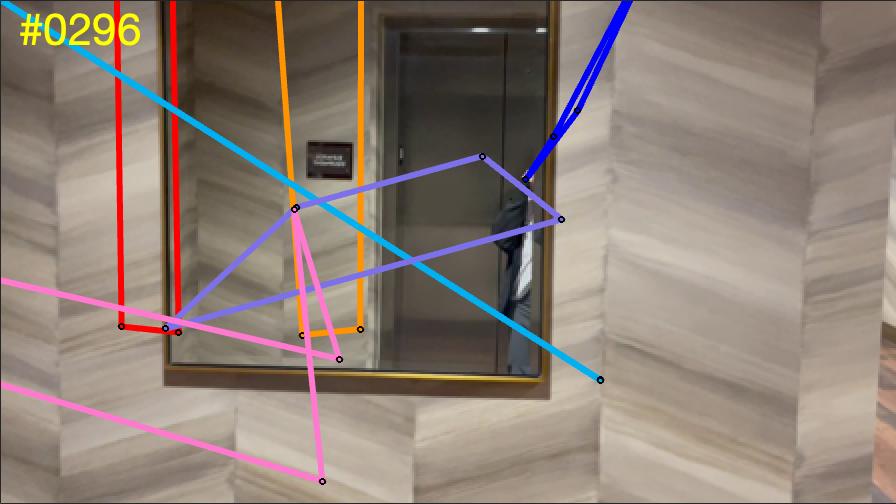} \includegraphics[width=2.7cm]{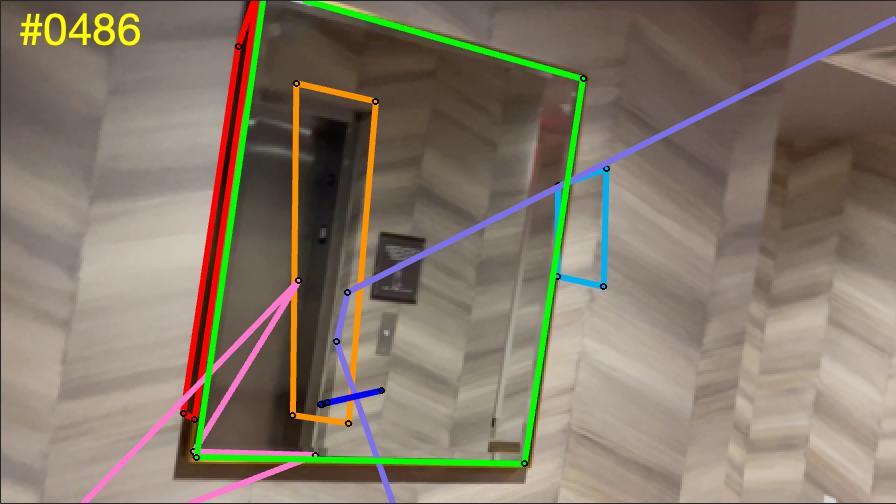} & \includegraphics[width=2.7cm]{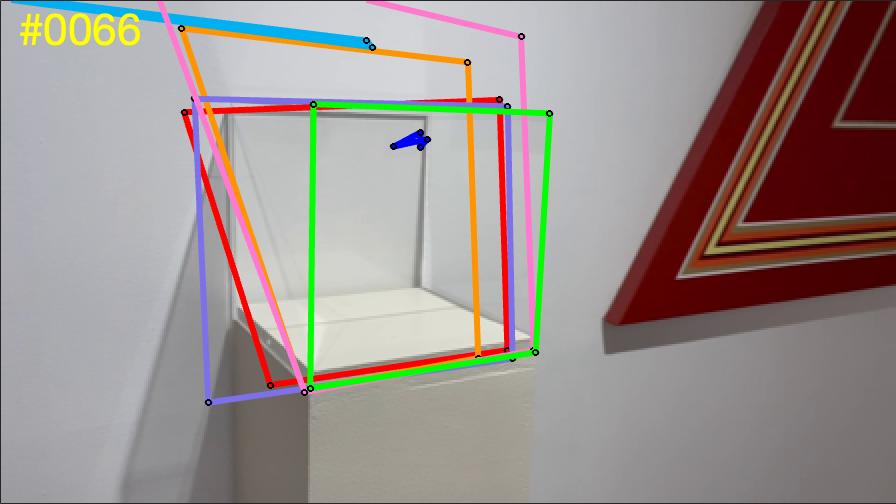}  \includegraphics[width=2.7cm]{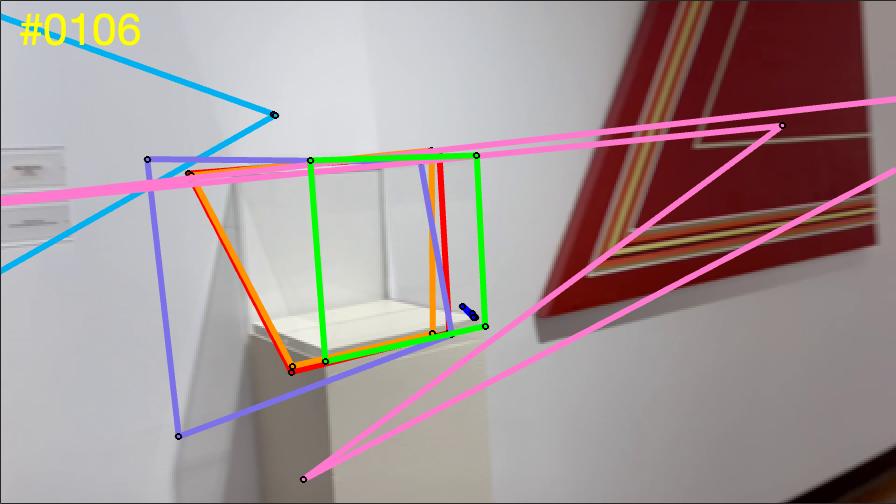} \includegraphics[width=2.7cm]{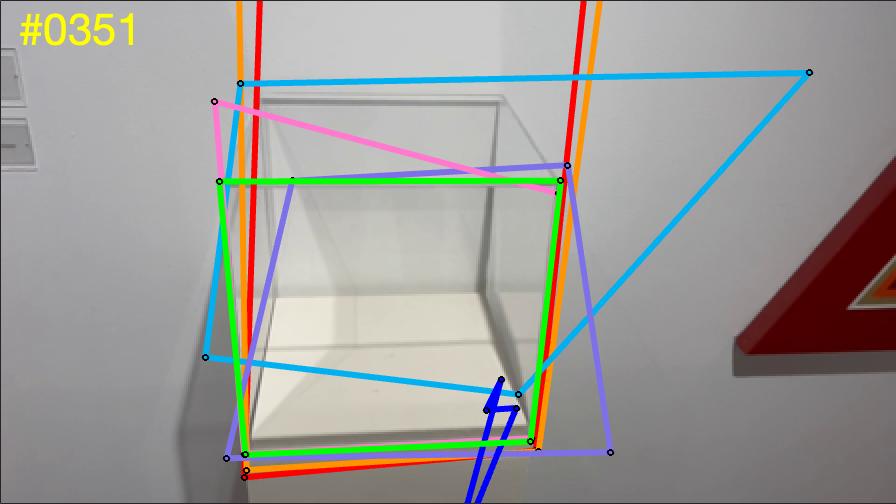} \\
{\small (c) Sequence with background clutter and perspective distortion}& {\small (d) Sequence with background clutter and rotation} \\
\includegraphics[width=2.7cm]{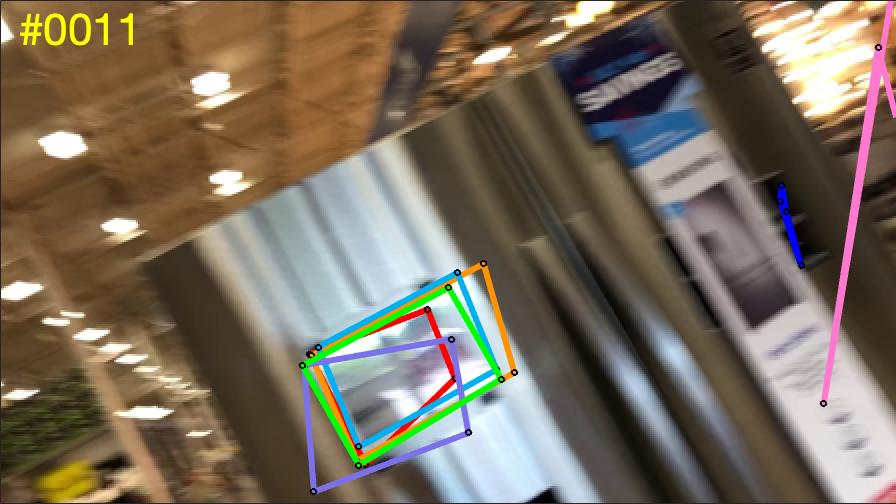} \includegraphics[width=2.7cm]{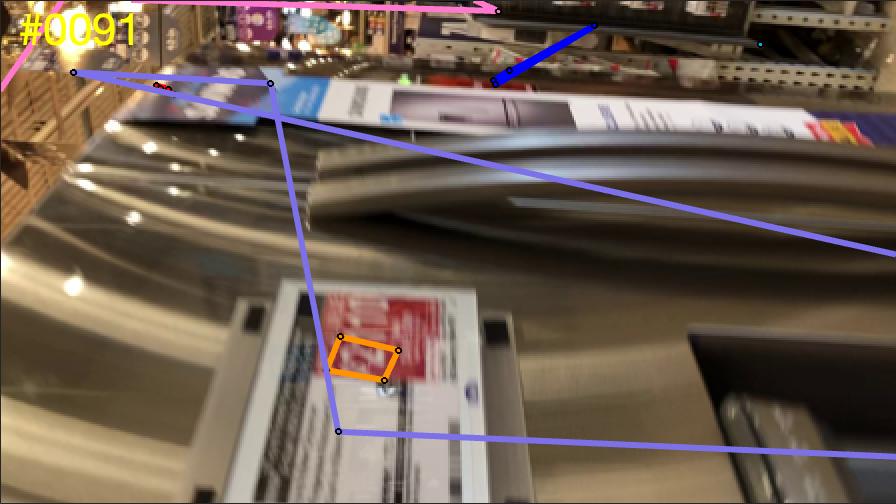} \includegraphics[width=2.7cm]{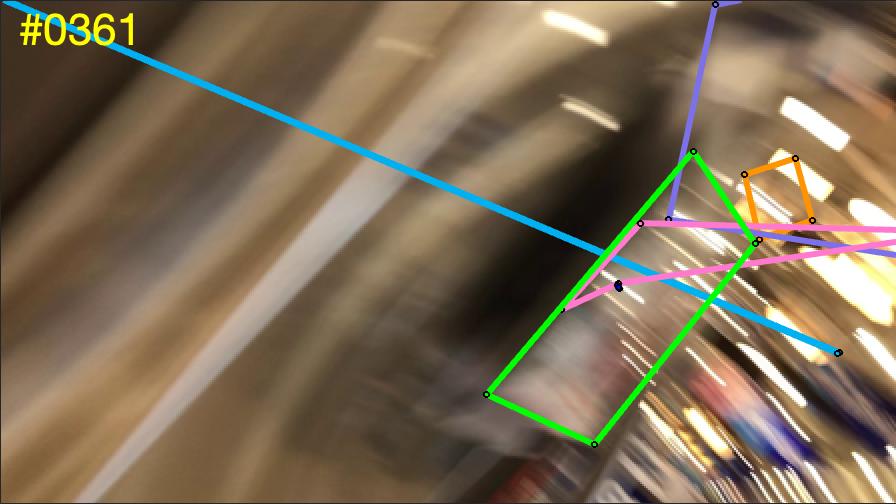} & \includegraphics[width=2.7cm]{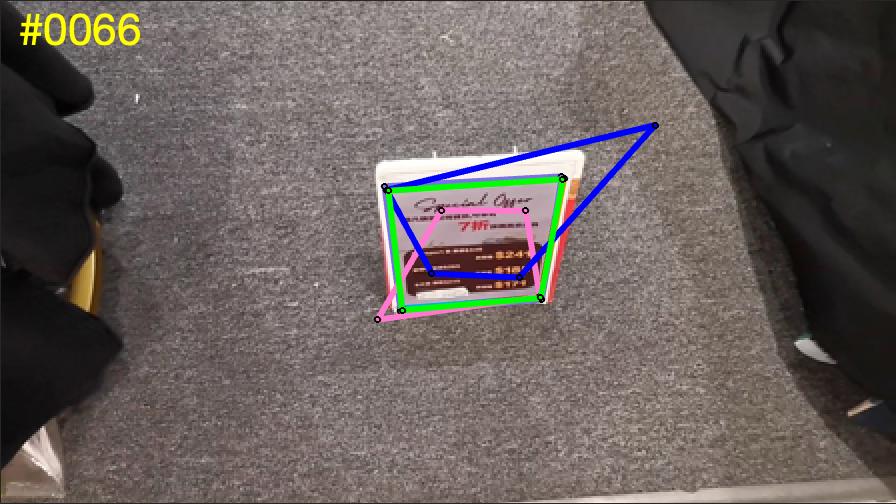}  \includegraphics[width=2.7cm]{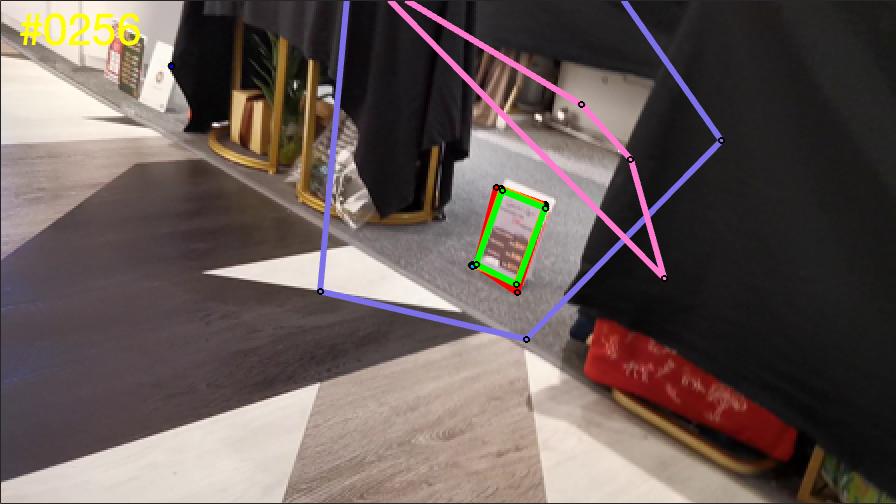} \includegraphics[width=2.7cm]{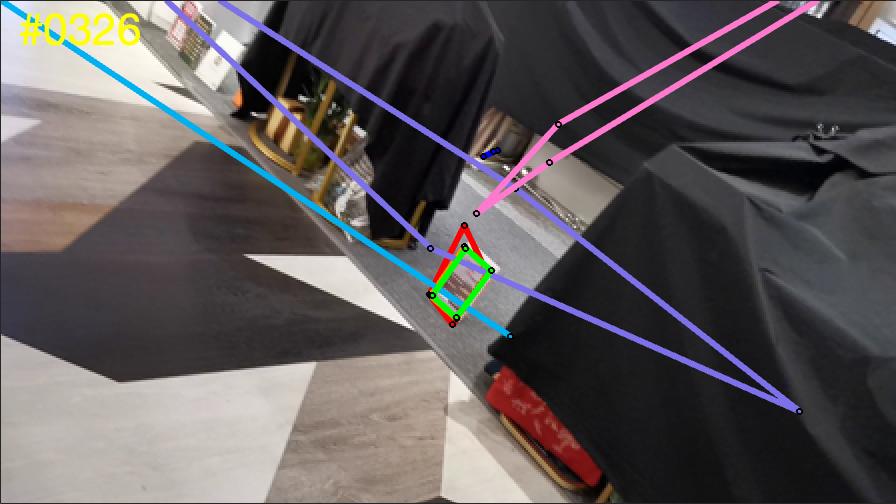} \\
{\small (e) Sequence with motion blur and out-of-view }& {\small (f) Sequence with low resolution and perspective distortion} \\
\multicolumn{2}{c}{\includegraphics[width=11.5cm]{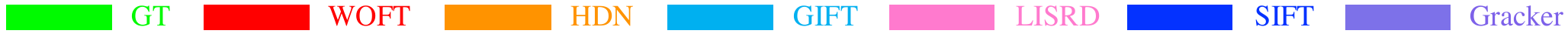}}\\
\end{tabular}
\caption{Qualitative results of five trackers with the highest precision scores on different sequences. We observe that these planar trackers drift to the background region or even lose the target object due to different challenging factors in the videos such as background clutter, scale variation, perspective distortion, motion blur, rotation, out-of-view and low resolution.
}
\label{tracking_res}
\end{figure*}

\renewcommand\arraystretch{0.91}
\begin{table*}[!t]
  \setlength{\tabcolsep}{6.5pt}
  \centering
  \caption{Comparison of PlanarTrack$_\textbf{Tst}$ to POT-210~\cite{liang2018planar} and its subset POT-210$_\text{UC}$ in unconstrained condition on PRE and SUC. Note that, the threshold for SUC is set to the same 30 for all experiments for fair comparison.}
    \begin{tabular}{lccccccccccc}
    \specialrule{.1em}{.05em}{.05em}  
          &       & \tabincell{c}{WOFT\\\cite{vserych2023planar}}  & \tabincell{c}{HDN\\\cite{zhan2022homography}}   & \tabincell{c}{GIFT\\\cite{liu2019gift}}
          & \tabincell{c}{LISRD\\\cite{pautrat2020online}}
          & \tabincell{c}{SIFT\\\cite{lowe2004distinctive}}
          & \tabincell{c}{Gracker\\\cite{wang2017gracker}} & \tabincell{c}{SOL\\\cite{hare2012efficient}}    & \tabincell{c}{SCV\\\cite{richa2011visual}}   & \tabincell{c}{ESM\\\cite{benhimane2004real}}     & \tabincell{c}{IC\\\cite{baker2004lucas}} \\
    \hline\hline
    \multirow{2}[0]{*}{\textbf{POT-210}~\cite{liang2018planar}} & PRE   & 0.805 & 0.612 & 0.553 & 0.617 & 0.692 & 0.392 & 0.417 & 0.228 & 0.204 & 0.121 \\
          & SUC   & 0.572 & 0.484 & 0.404 & 0.463 & 0.445 & 0.331 & 0.312 & 0.200 & 0.183 & 0.114 \\
    \hline
    \multirow{2}[0]{*}{\textbf{POT-210$_\text{UC}$}~\cite{liang2018planar}} & PRE   & 0.768 & 0.567 & 0.528 & 0.581 & 0.578 & 0.185 & 0.289 & 0.105 & 0.100 & 0.053 \\
          & SUC   & 0.536 & 0.442 & 0.379 & 0.419 & 0.378 & 0.195 & 0.224 & 0.092 & 0.086 & 0.050 \\
    \hline
     \rowcolor{gray!15} & PRE   & 0.433 & 0.263 & 0.254 & 0.167 & 0.142 & 0.121 & 0.113 & 0.097 & 0.064 & 0.048 \\
     \rowcolor{gray!15} \multirow{-2}{*}{\textbf{PlanarTrack$_\textbf{Tst}$}} &  SUC   & 0.306 & 0.236 & 0.223 & 0.137 & 0.107 & 0.098 & 0.082 & 0.073 & 0.147 & 0.038 \\
    \specialrule{.1em}{.05em}{.05em} 
    \end{tabular}%
  \label{tab3}%
\end{table*}%

\vspace{0.3em}
\noindent
\textbf{Challenging Factor-based Evaluation.} For in-depth analysis of different planar trackers, we further conduct evaluation on the eight challenging factors. Due to limited space, we display the results on the two most common challenging factors including \emph{perspective distortion} (PD) and \emph{scale variation} (SV) and on the two most difficult challenging factors including \emph{low resolution} (LR) and \emph{motion blur} (MB) in Fig.~\ref{fig7}, and refer reader to \emph{supplementary material} for more results. From Fig.~\ref{fig7}, we can observe that WOFT shows the best performance on both the commonest and most difficult challenges. In specific, it achieves the PRE scores of 0.434, 0.423, 0.364 and 0.386 on PD, SV, LR and MB, which outperform HDN, the second best on PD, SV and MB with PRE scores of 0.264, 0.258 and 0.252, and GIFT, the second best on LR with 0.252 PRE score. This again demonstrates the importance of temporal information for planar tracking. In addition, the tracking performance severely degrades on LR and MB. We argue that these two challenges may result in ineffective feature extraction of points or targets, causing tracking drifts or failures. Future research can be devoted to improvements in these two situations.

\vspace{0.3em}
\noindent
\textbf{Qualitative Results.} To better understand the planar tracking algorithms, we demonstrate the qualitative results of the top six trackers with the highest precision scores, consisting of WOFT, HDN, GIFT, LISRD, SIFT, and Gracker, in different challenging factors such as \emph{background clutter}, \emph{scale variation}, \emph{perspective distortion}, \emph{motion blur}, \emph{rotation}, \emph{out-of-view} and \emph{low resolution} in Fig.~\ref{tracking_res}. As in Fig.~\ref{tracking_res}, we can see that although some trackers can deal with certain challenging factor. However, when multiple challenging factors occur simultaneously, the trackers may drift to the background region or even lose the planar target. 

\subsection{Comparison with POT-210.}

POT-210~\cite{liang2018planar} is currently one of the most popular benchmarks for planar object tracking. However, most sequences in POT-210 contain mainly one challenging factors and very few (\ie, 30) are involved with different challenges, which may not faithfully reflect the difficulties and complexities in real scenarios for evaluation. In addition, the lack of diversity in planar targets also limits its usage. To mitigate these, all sequences in PlanarTrack are freely recorded in unconstrained conditions and the planar targets are unique in each video for diversity. Consequently, our PlanarTrack is more challenging and realistic in practical applications.

To verify the above, we compare existing planar trackers on POT-210 and PlanarTrack$_\textbf{Tst}$. Tab.~\ref{tab3} shows the comparison results. From Tab.~\ref{tab3}, we can see that the best performing tracker on POT-210 is WOFT that achieves 0.805/0.572 PRE/SUC scores. Nevertheless, when utilized for tracking planar targets on PlanarTrack$_\textbf{Tst}$, its performance is severely degenerated. In specific, the PRE/SUC scores are decreased from 0.805/0.572 to 0.433/0.306, showing absolute perform drop of 37.2\%/26.6\% in PRE/SUC. Besides, SIFT with the second best PRE score of 0.692 on POT-210 heavily degrades to 0.142 on PlanarTrack$_\textbf{Tst}$, and HDN with the second best SUC score of 0.484 to 0.236. Furthermore, other trackers are degenerated as well on PlanarTrack$_\textbf{Tst}$.

In addition to POT-210, we further compare POT-210$_\text{UC}$, a subset of POT-210 with all videos in unconstrained conditions, with PlanarTrack$_\textbf{Tst}$ as they are both unconstrained. The comparisons are shown in Tab.~\ref{tab3}. As in Tab.~\ref{tab3}, we can see that POT-210$_\text{UC}$ is more challenging than POT-210, yet less difficult than PlanarTrack. The best tracker WOFT on POT-210$_\text{UC}$ demonstrates PRE/SUC scores of 0.786/0.536, while it degrades to 0.433/0.306 on PlanarTrack$_\textbf{Tst}$ with performance drop of 35.3\% and 23.0\%.

Through the above comparisons and analysis, we clearly see that PlanarTrack is more challenging and complex, and there is still a big room for improvements.

\subsection{Retraining on PlanarTrack}

\begin{table}[!t]\small
  \centering
  \caption{Retraining of HDN~\cite{zhan2022homography} using PlanarTrack$_\textbf{Tra}$.}
    \begin{tabular}{lccc}
    \specialrule{.1em}{.05em}{.05em}  
          &       & \tabincell{c}{Original\\HDN~\cite{zhan2022homography}} & \tabincell{c}{Retrained\\HDN} \\
    \hline\hline
    \multirow{2}[0]{*}{POT-210~\cite{liang2018planar}} & PRE   & 0.612 & 0.637 (\textcolor[RGB]{61,145,64}{+2.5\%}) \\
          & SUC   & 0.484 & 0.497 (\textcolor[RGB]{61,145,64}{+1.3\%}) \\
    \hline
    \multirow{2}[0]{*}{PlanarTrack$_\textbf{Tst}$} & PRE   & 0.263 & 0.294 (\textcolor[RGB]{61,145,64}{+3.1\%}) \\
          & SUC   & 0.236 & 0.260 (\textcolor[RGB]{61,145,64}{+2.4\%}) \\
    \specialrule{.1em}{.05em}{.05em} 
    \end{tabular}%
  \label{tab4}%
\end{table}%

One of the major goals for our PlanarTrack is to provide a dedicated platform for developing deep planar trackers. To validate its effectiveness, we conduct retraining experiments using PlanarTrack$_{\text{Tra}}$ instead of the synthetic data on the recent HDN. Please notice that, we do not perform retraining on WOFT because it does not provide the training implementation. In the retraining, the parameters and settings are kept the same as in the original approach. Tab.~\ref{tab4} demonstrates the results of the retraining experiment. From Tab.~\ref{tab4}, we can observe clearly that, when leveraging task-specific data for training, the performance of planar tracker is significantly increased. In specific, the PRE/SUC scores are increased from 0.612/0.484 to 0.637/0.495 on POT-210 and from 0.263/0.236 to 0.294/0.260 on our PlanarTrack$_\textbf{Tst}$, which demonstrates the effectiveness and necessity of large-scale platform for improving planar object tracking.

\section{PlanarTrack$_\mathrm{BB}$ and Experiments}

Planar objects are common to see in our daily life. However, localization of planar targets with \emph{generic visual trackers} has rarely been studied at large scale, even in the existing large-scale generic tracking benchmarks (\eg,~\cite{fan2019lasot,huang2019got,muller2018trackingnet}). For generic trackers, they should be able to locate the targets regardless of their categories. To discover the capacities of these generic object trackers in handling planar-like targets, we introduce PlanarTrack$_{\text{BB}}$, a by-product of PlanarTrack. Specifically, PlanarTrack$_{\text{BB}}$ shares the same images and dataset split from PlanarTrack but converts four annotated corner points to an axis-aligned bounding box in each frame, and it is specially used for large-scale evaluation of generic trackers in dealing with planar-like targets. We refer readers to \emph{supplementary material} for detailed construction of PlanarTrack$_{\text{BB}}$ and examples.

\renewcommand\arraystretch{1.0}
\begin{table}[!t]\small
  \setlength{\tabcolsep}{4.5pt}
  \centering
  \caption{Evaluation of generic trackers on PlanarTrack$_\text{BB}$ and comparison with other popular generic benchmarks using SUC$_\text{BB}$.}
    \begin{tabular}{lcc>{\columncolor{gray!15}}c}
    \specialrule{.1em}{.05em}{.05em} 
          & \tabincell{c}{\textbf{TrackingNet}\\\cite{muller2018trackingnet}} & \tabincell{c}{\textbf{LaSOT}\\\cite{fan2019lasot}} &  \tabincell{c}{\textbf{PlanarTrack$_\text{BB}$}\\(ours)} \\
    \hline\hline
    SwinTrack~\cite{lin2022swintrack} & 0.840 & 0.713 & 0.663 \\
    MixFormer~\cite{cui2022mixformer} & 0.839 & 0.701 & 0.657 \\
    OStrack~\cite{ye2022joint} & 0.839 & 0.711 & 0.648 \\
    TransInMo~\cite{guo2022learning} & 0.817 & 0.657 & 0.636 \\
    AiATrack~\cite{gao2022aiatrack} & 0.827 & 0.690 & 0.624 \\
    STARK~\cite{yan2021learning} & 0.820 & 0.671 & 0.618 \\
    TransT~\cite{chen2021transformer} & 0.814 & 0.649 & 0.608 \\
    SimTrack~\cite{chen2022backbone} & 0.834 & 0.705 & 0.606 \\
    ToMP~\cite{mayer2022transforming}  & 0.815 & 0.685 & 0.605 \\
    TrDiMP~\cite{wang2021transformer} & 0.784 & 0.639 & 0.584 \\
    \specialrule{.1em}{.05em}{.05em} 
    \end{tabular}%
  \label{tab5}%
\end{table}%

We select ten state-of-the-art generic trackers for evaluation. Notice that, these trackers are all Transformer-based, consisting of SwinTrack~\cite{lin2022swintrack}, OStrack~\cite{ye2022joint}, SimTrack~\cite{chen2022backbone}, MixFormer~\cite{cui2022mixformer}, AiATrack~\cite{gao2022aiatrack}, ToMP~\cite{mayer2022transforming}, STARK~\cite{yan2021learning}, TransInMo~\cite{guo2022learning}, TransT~\cite{chen2021transformer} and TrDiMP~\cite{wang2021transformer}, and the best version of each visual tracker is employed for evaluation with SUC$_\text{BB}$ which is success score for bounding box-based tracking~\cite{wu2013online}. Tab.~\ref{tab5} reports the evaluation results and comparisons with other large-scale generic tracking benchmarks including LaSOT~\cite{fan2019lasot} and TrackingNet~\cite{muller2018trackingnet}. Notice, GOT-10k~\cite{huang2019got} is not included for comparison because it adopts a different evaluation metric. From  Tab.~\ref{tab5}, we can observe that although existing generic trackers achieve outstanding performance, they are heavily degraded when dealing with planar-like target objects. For example, the top-performing generic trackers SwinTrack ans OStrack obtain 0.713/0.840 and 0.701/0.839 SUC scores on LaSOT/TrackingNet, while degrade 0.663 and 0.648, respectively, on PlanarTrack$_\text{BB}$, which indicates that more attention should be paid to handle such planar trackers, though they are rigid. Due to limited space, please see \emph{supplementary material} for more results.

\section{Conclusion and Limitation}

In this work, we introduce a new benchmark named PlanarTrack. PlanarTrack consists of 1,000 videos collected in unconstrained conditions from natural scenes, and has more than 490K image frames. To the best of our knowledge, PlanarTrack is, to date, the first large-scale challenging dataset dedicated for planar tracking. To understand existing methods on PlanarTrack and provide comparison for future research, we perform experiments by evaluating ten representative planar trackers and conduct in-depth analysis. By releasing PlanarTrack, we expect to facilitate research and applications of planar tracking. Furthermore, we develop a by-produce dataset, dubbed PlanarTrack$_\text{BB}$, based on PlanarTrack for studying generic trackers on localizing planar-like target objects.

Despite contributions, there are limitations of this work. First, given the propose large-scale PlanarTrack, a baseline that outperforms other planar trackers is not provided. Second, since videos in PlanarTrack are relatively short, they may not be suitable for long-term tracking. Considering  our aim is to make the first attempt for large-scale planar tracking, we keep these as open questions for future research.

\vspace{0.3em}
\noindent
{\bf Acknowledgement.} We sincerely thank anonymous volunteers
for their help in constructing PlanarTrack.

\section*{Supplementary Material}

In this supplementary material, we present additional details of PlanarTrack and experimental results. Specifically, \textbf{S1} shows more comparison of the training and testing sets on different challenging factors. In \textbf{S2}, we display more detailed results of each tracker on challenging factors using in terms of precision and success on the proposed PlanarTrack. \textbf{S3} presents the detailed construction of PlanarTrack$_\text{BB}$ from PlanarTrack for generic object tracking and demonstrates several examples. \textbf{S4} shows more results of generic trackers on PlanarTrack$_\text{BB}$.

\subsection*{S1. Comparison of Training and Testing Sets}

\begin{figure}[ht]
    \centering
    \includegraphics[width=0.85\linewidth]{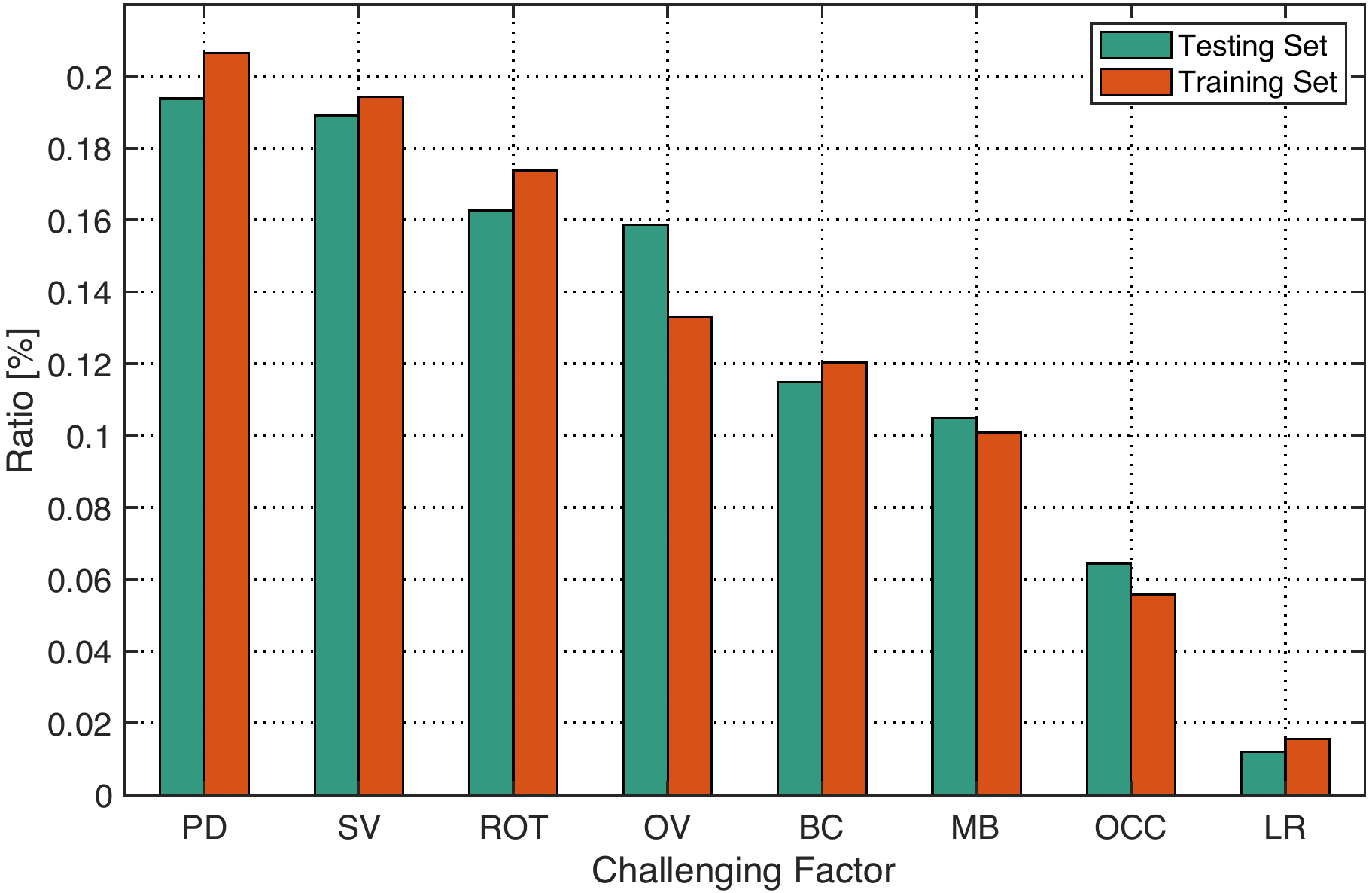}
    \caption{Distribution of sequences on each challenging factor.}
    \label{figs1}
\end{figure}

In order to further compare the training and testing sets of PlanarTrack, we demonstrate the ratios of sequences in these two sets on eight different challenging factors in Fig~\ref{figs1}. From Fig~\ref{figs1}, we can see that the training and testing sets are close to each other in  the distributions of videos in different challenges, which shows the consistency of training/testing split in PlanarTrack.

\subsection*{S2. Detailed Challenging Factor-based Results}

\begin{figure}[!t]
		\centering
		\begin{tabular}{c}
\includegraphics[width=3.99cm]{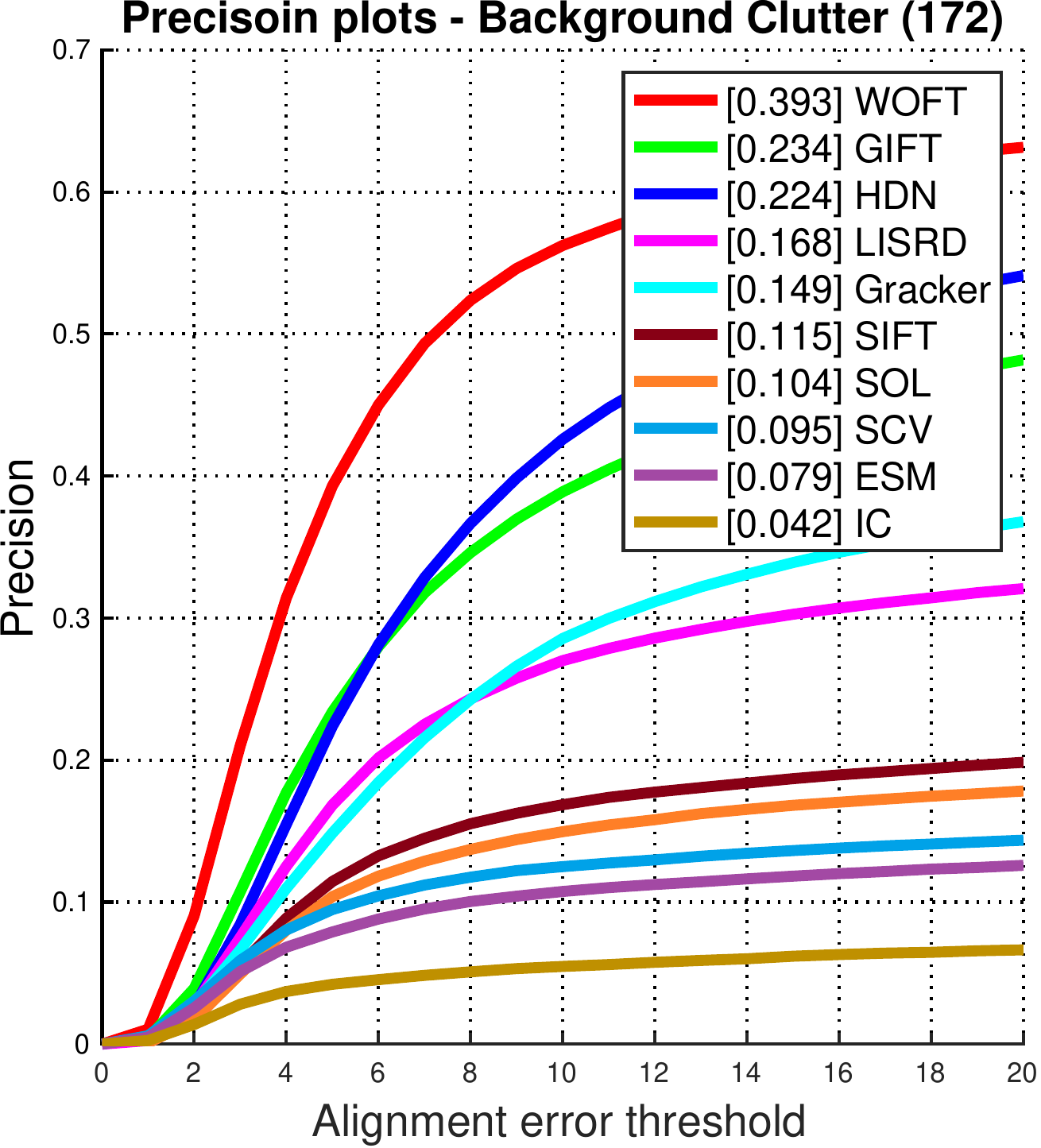} \includegraphics[width=3.99cm]{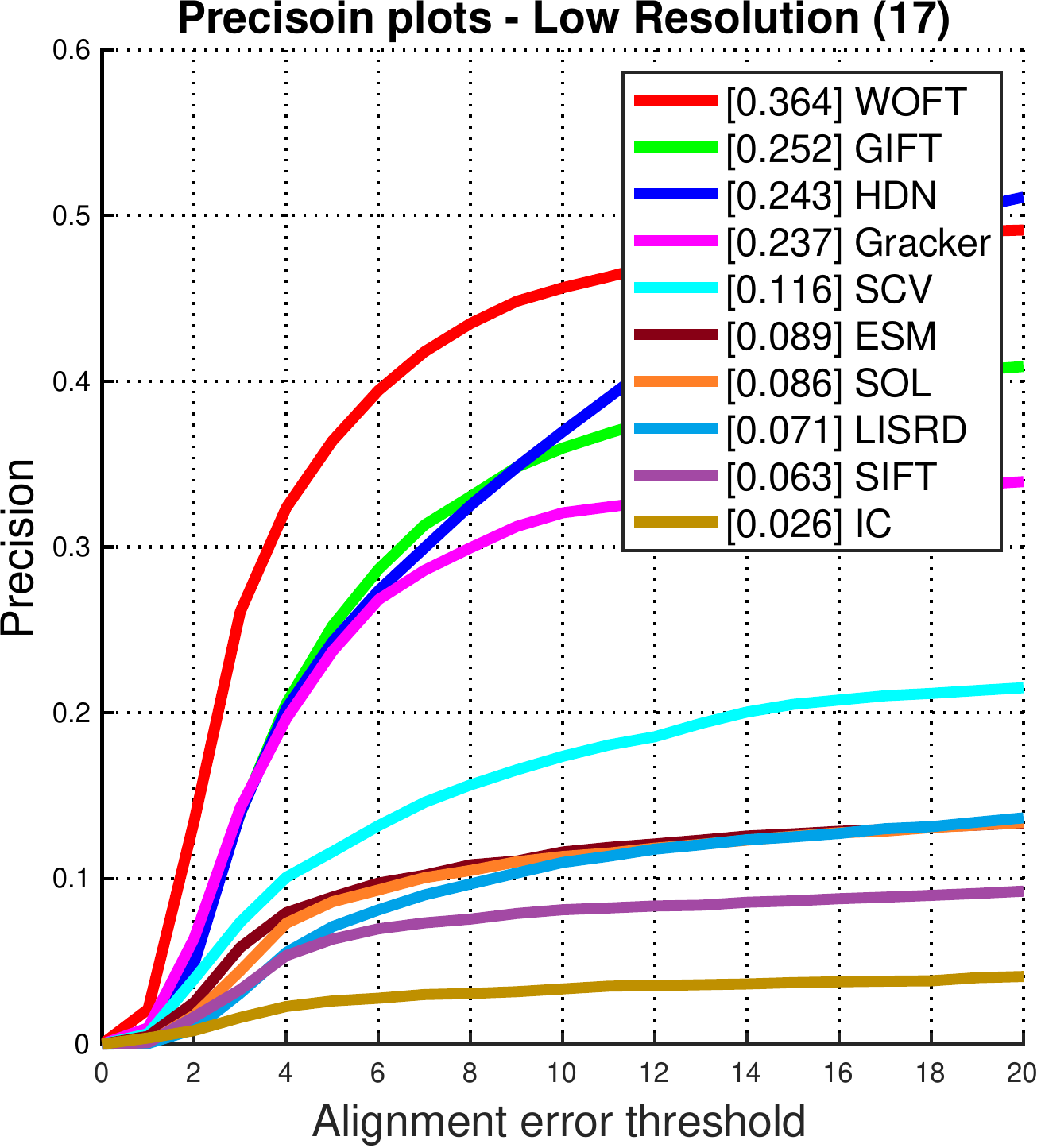} \\
\includegraphics[width=3.99cm]{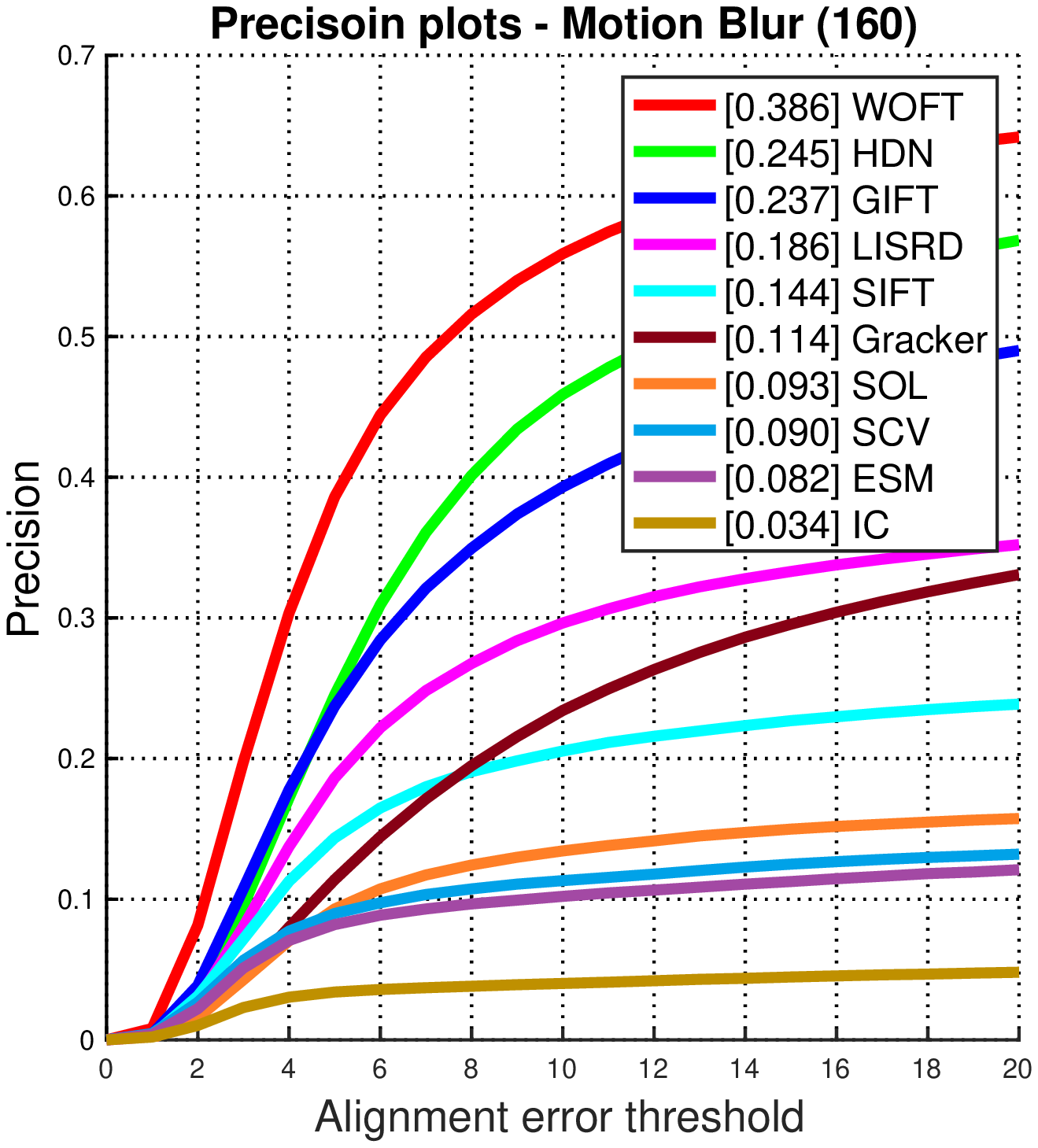} \includegraphics[width=3.99cm]{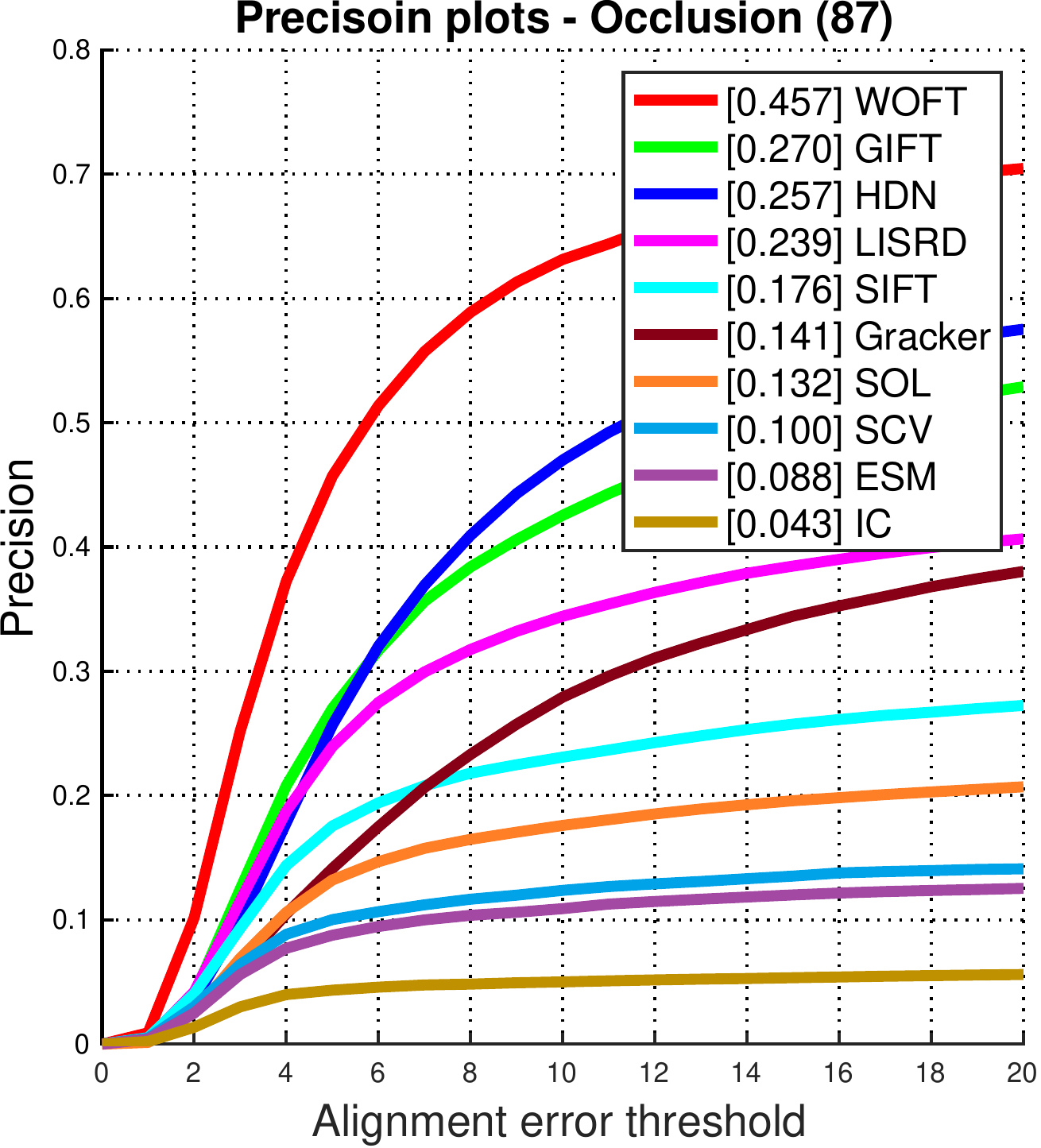} \\
\includegraphics[width=3.99cm]{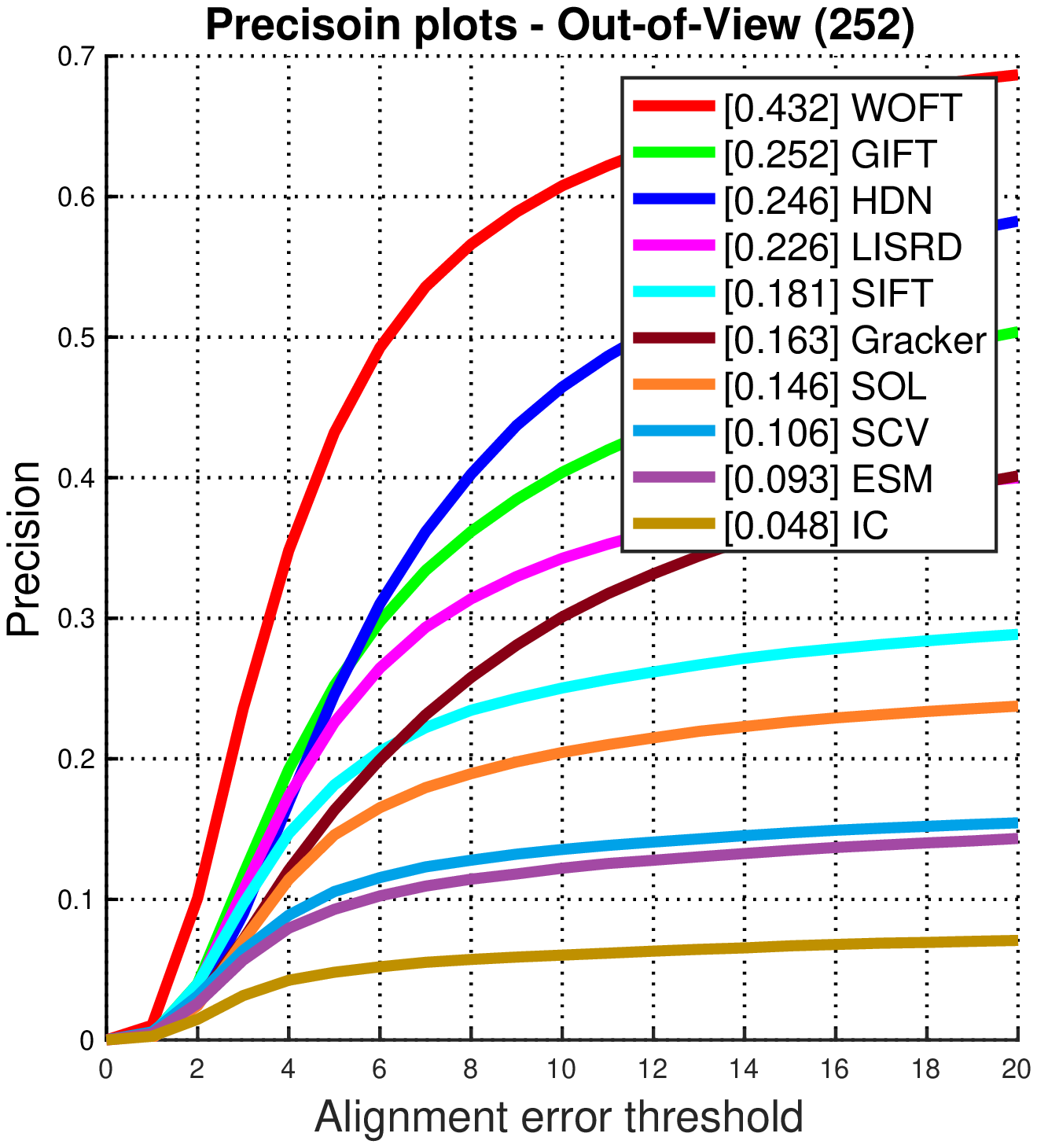} \includegraphics[width=3.99cm]{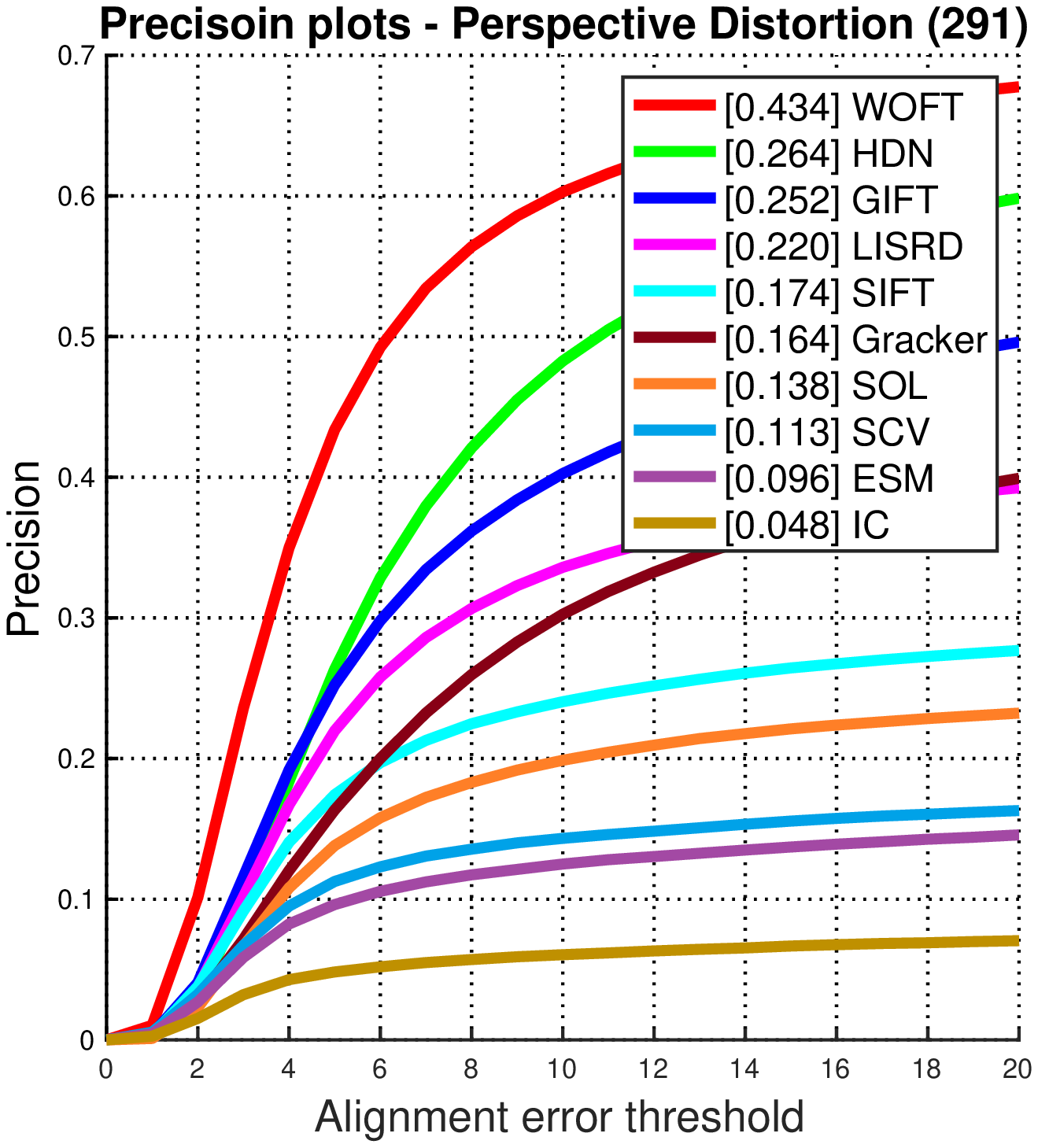} \\
\includegraphics[width=3.99cm]{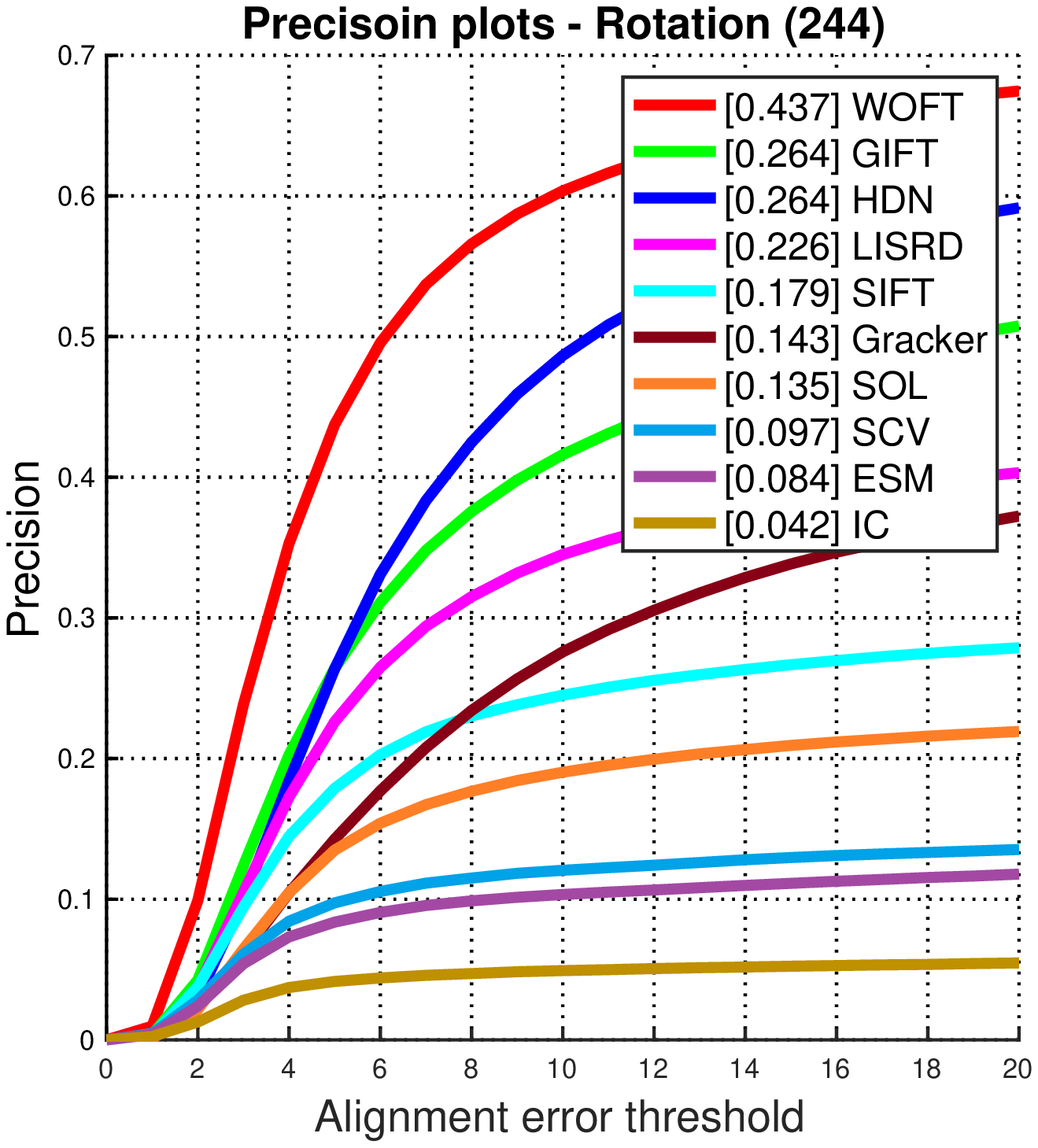} \includegraphics[width=3.99cm]{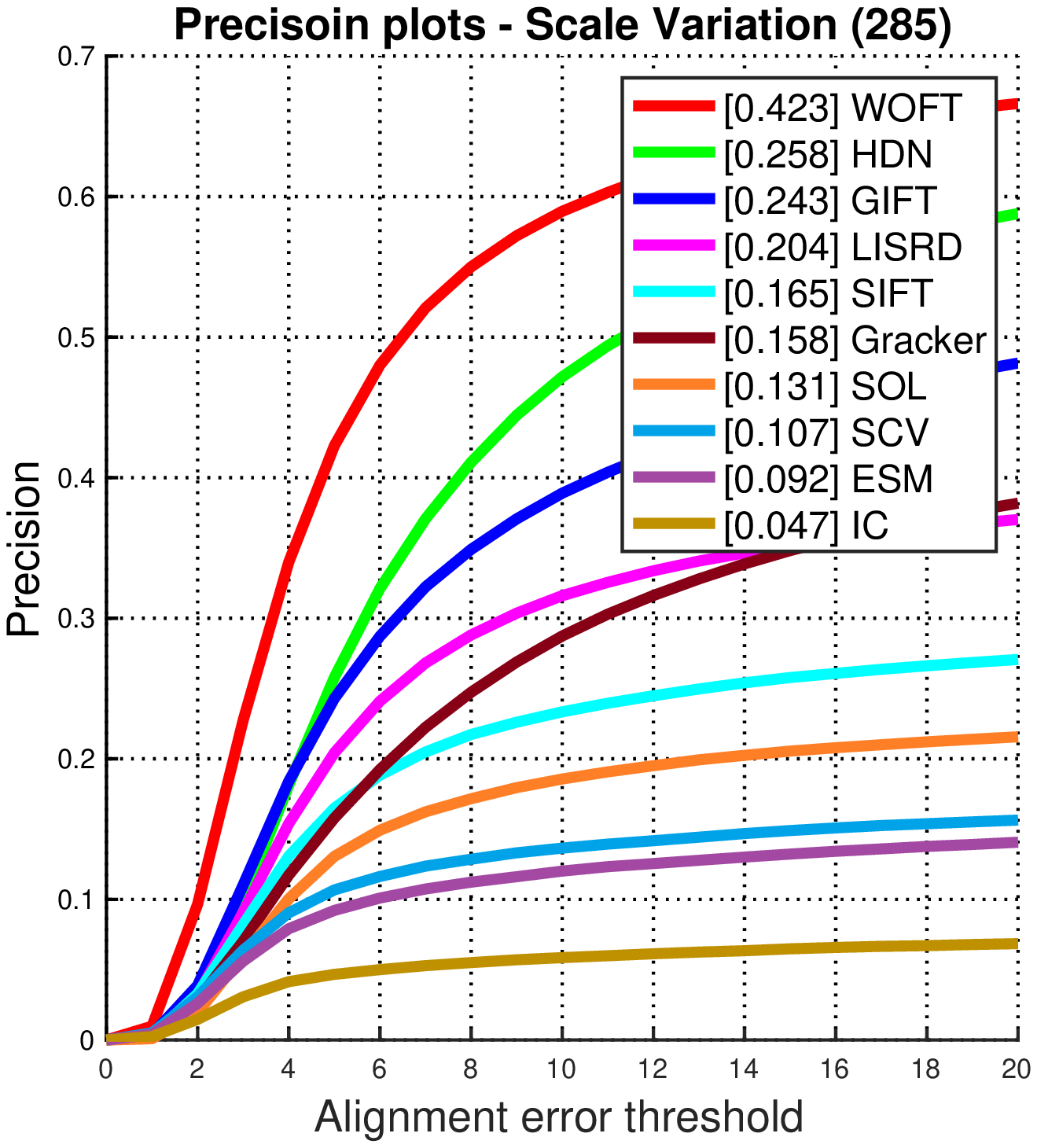} \\
\end{tabular}
\caption{Performance of trackers on each challenging factor using precision. Best viewed in color.
}
\label{att_pre}
\end{figure}

\begin{figure}[!t]
		\centering
		\begin{tabular}{c}
\includegraphics[width=3.99cm]{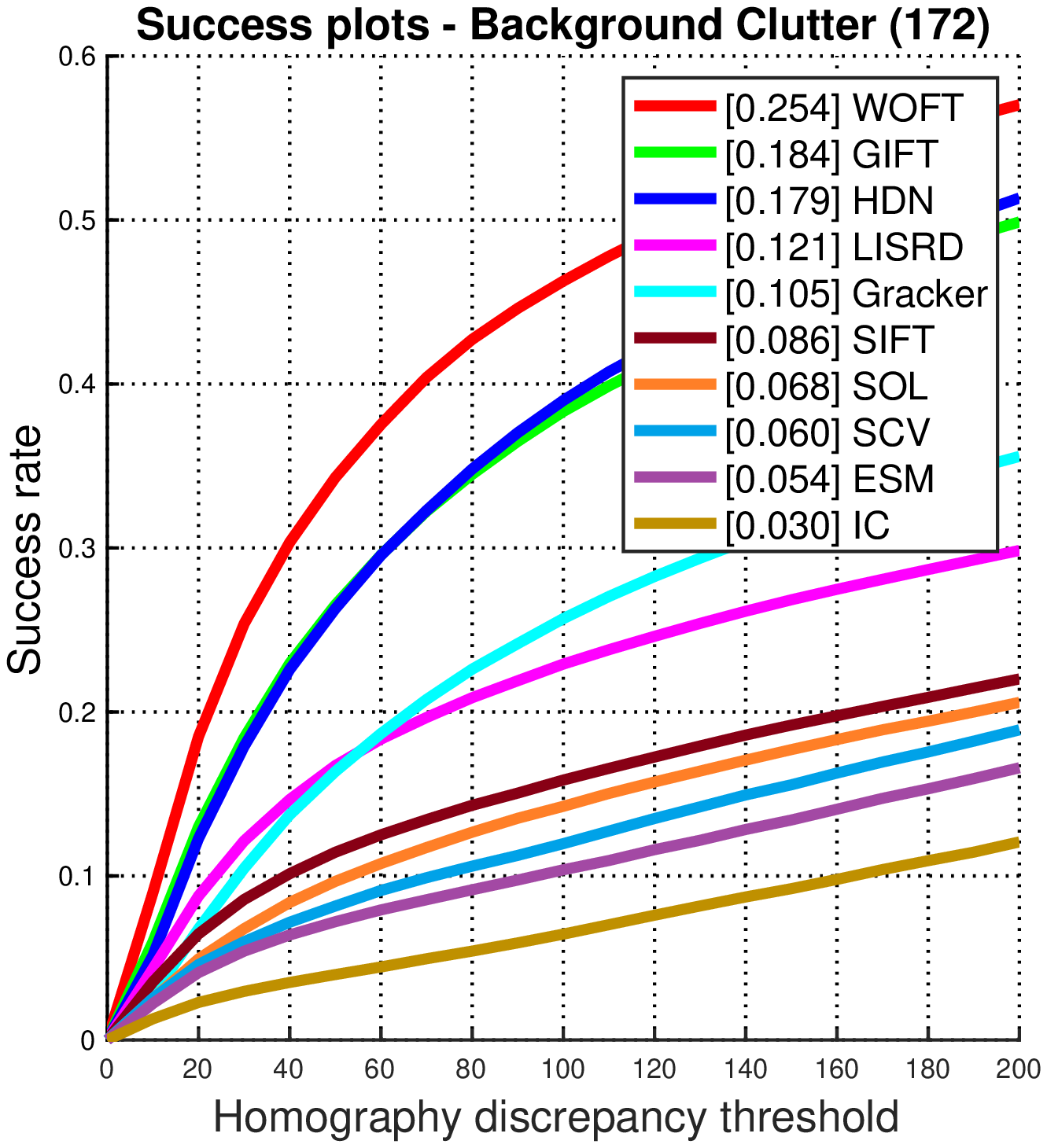} \includegraphics[width=3.99cm]{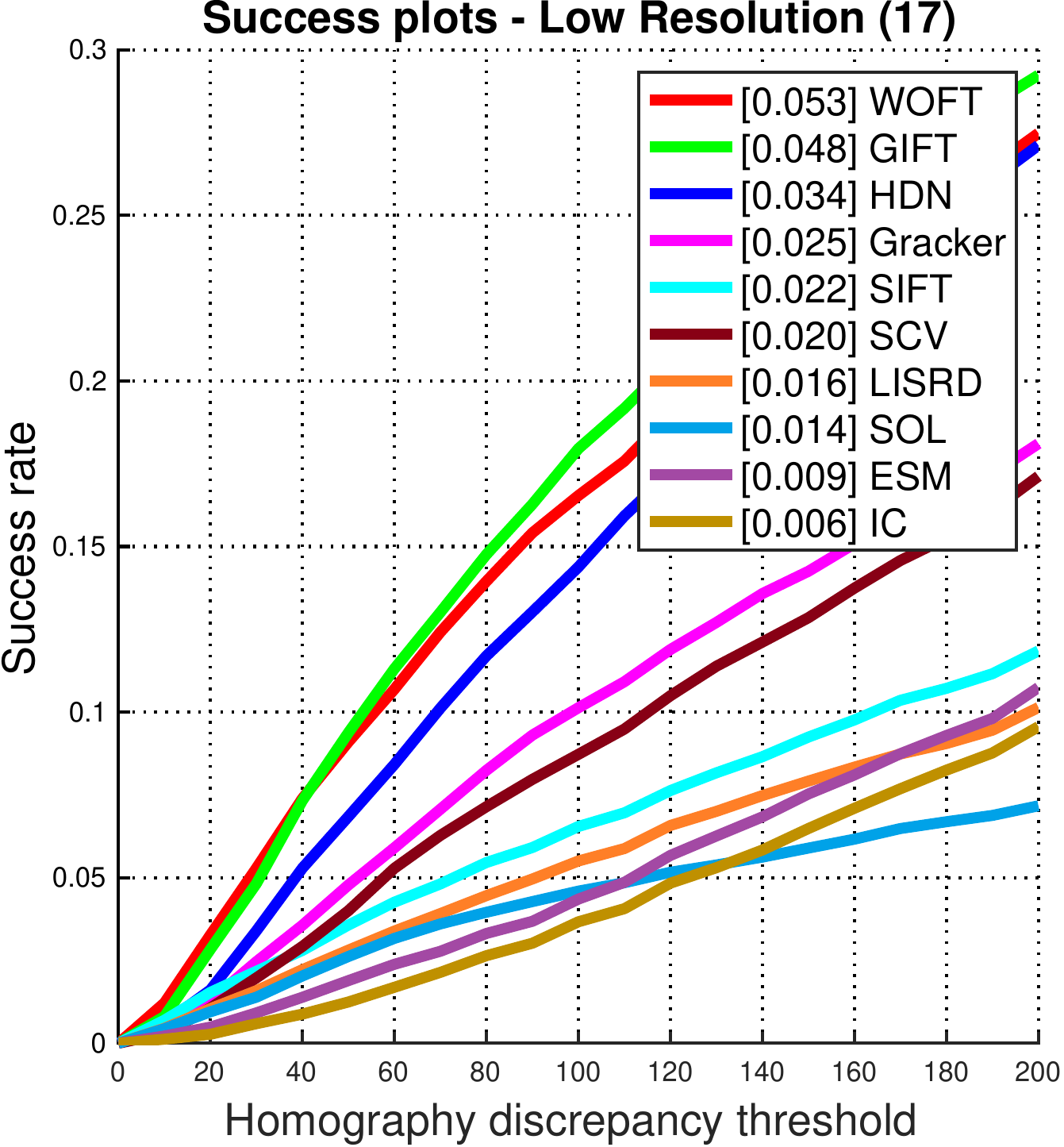} \\
\includegraphics[width=3.99cm]{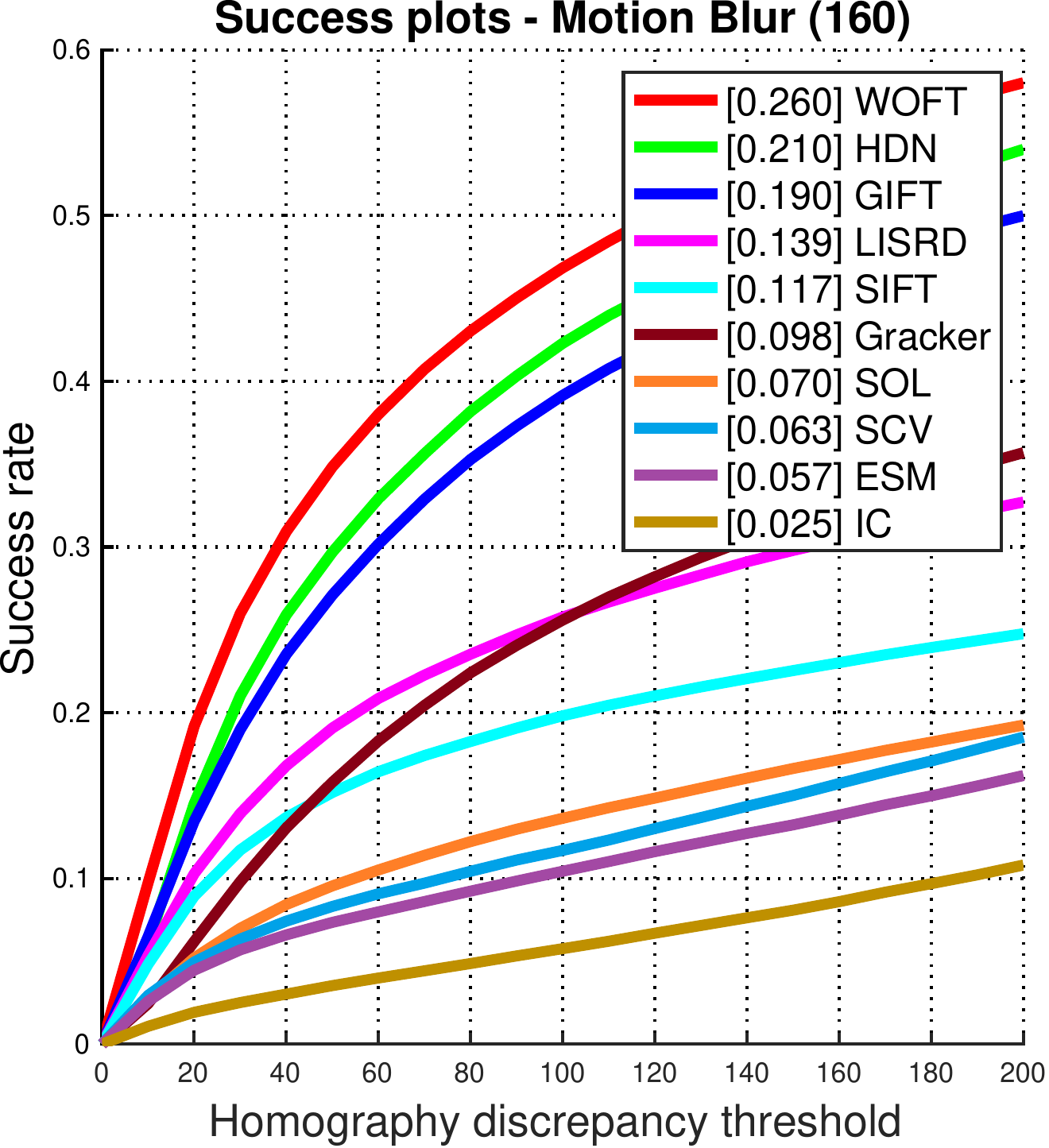} \includegraphics[width=3.99cm]{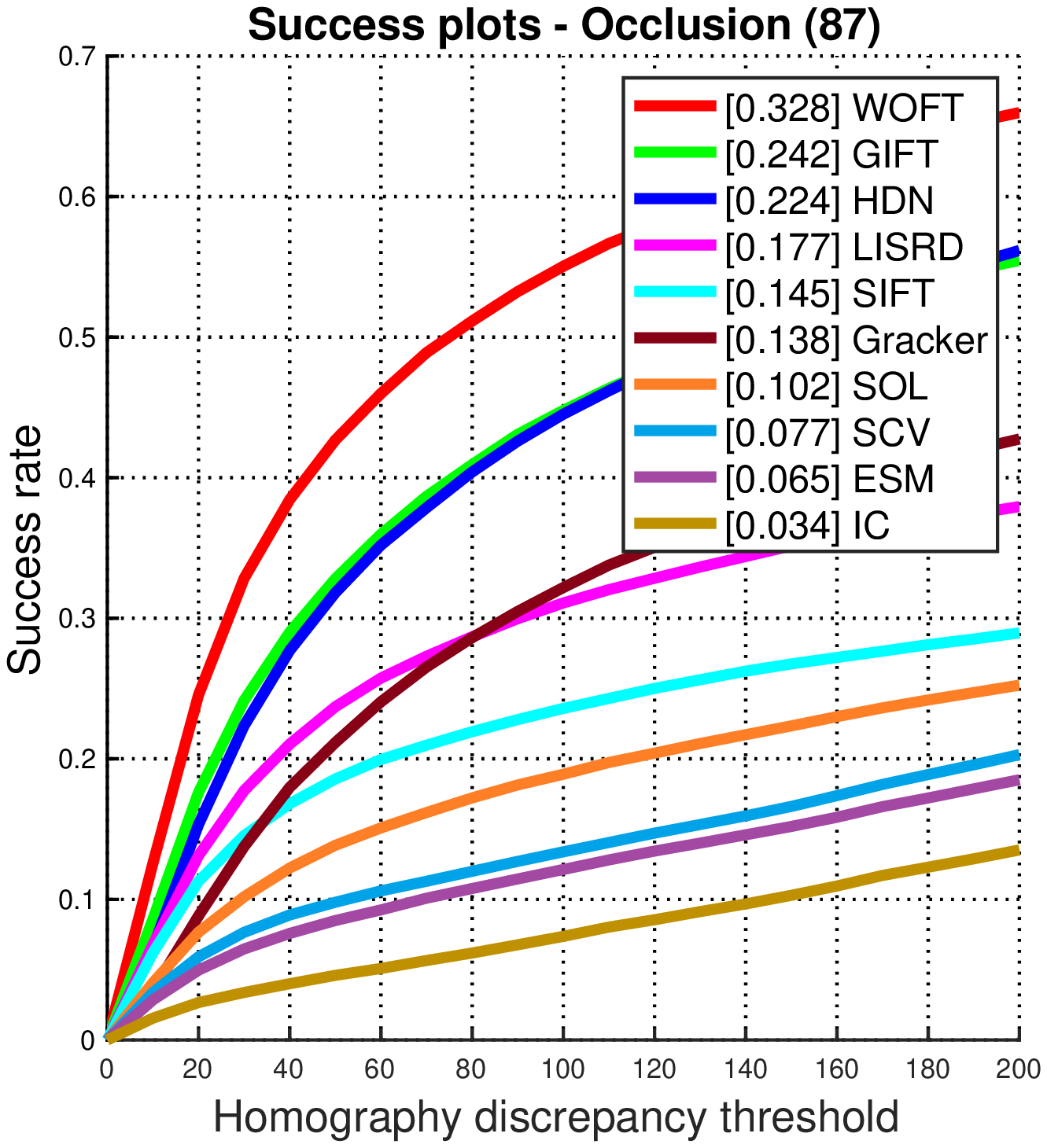} \\
\includegraphics[width=3.99cm]{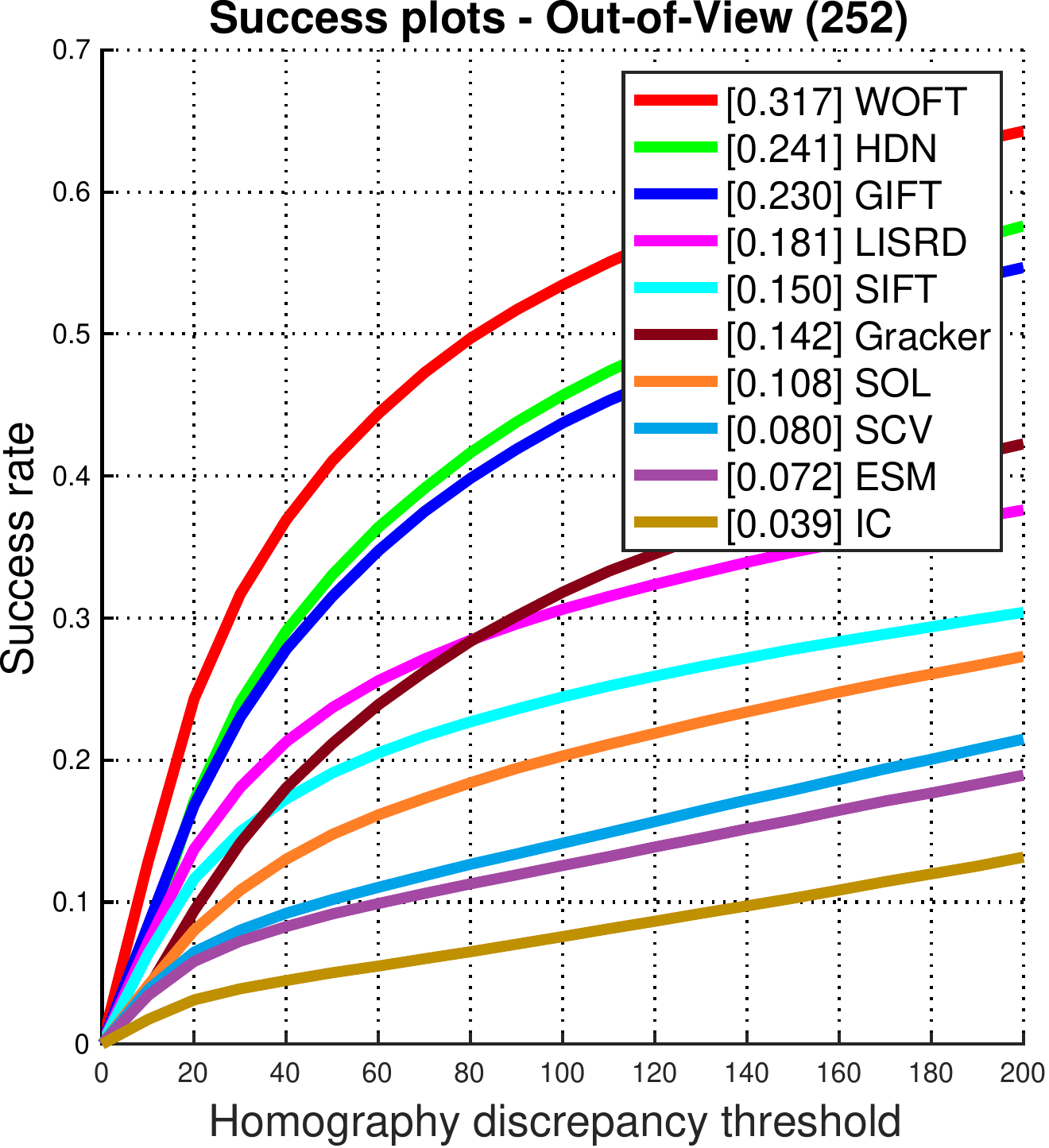} \includegraphics[width=3.99cm]{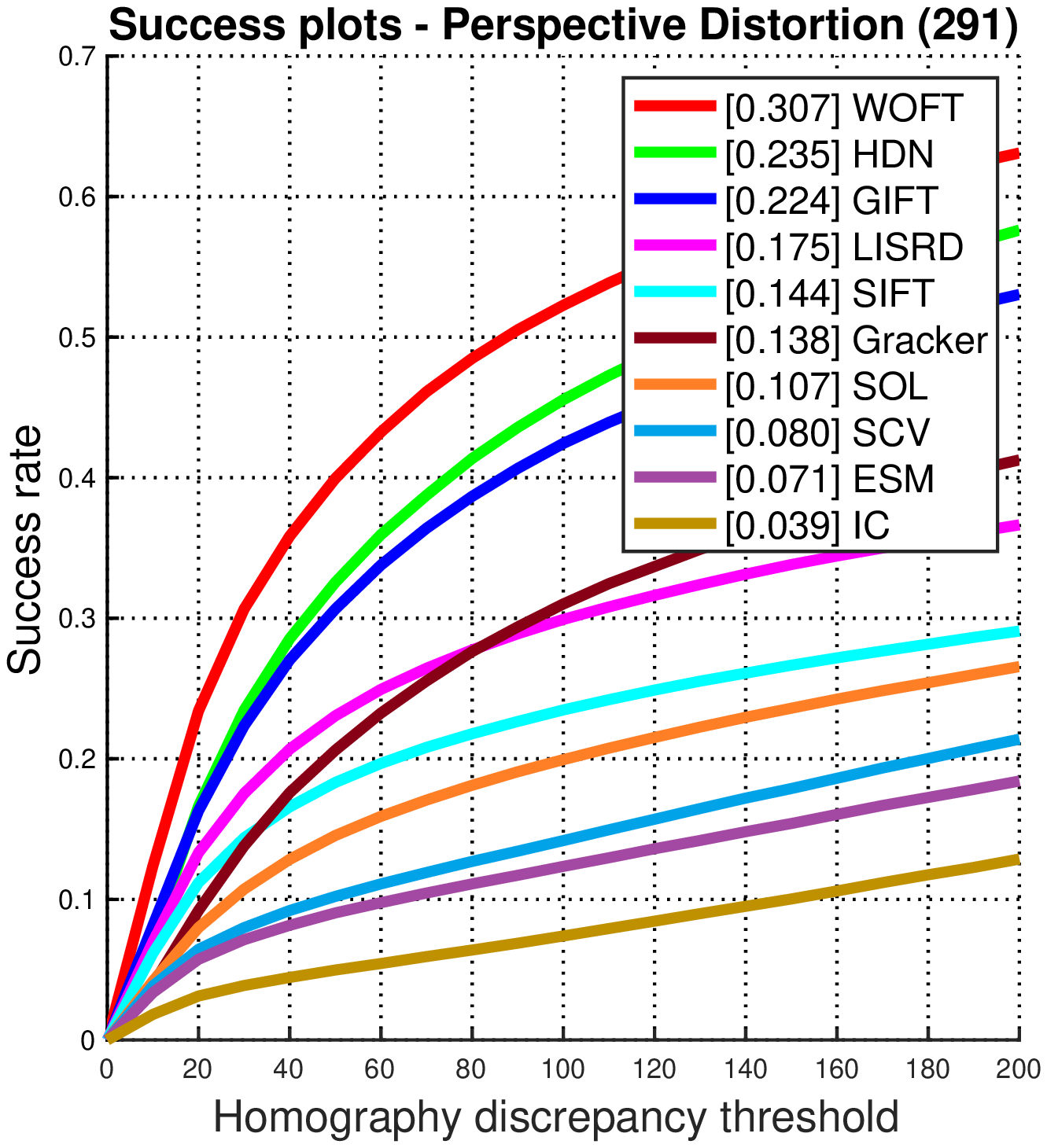} \\
\includegraphics[width=3.99cm]{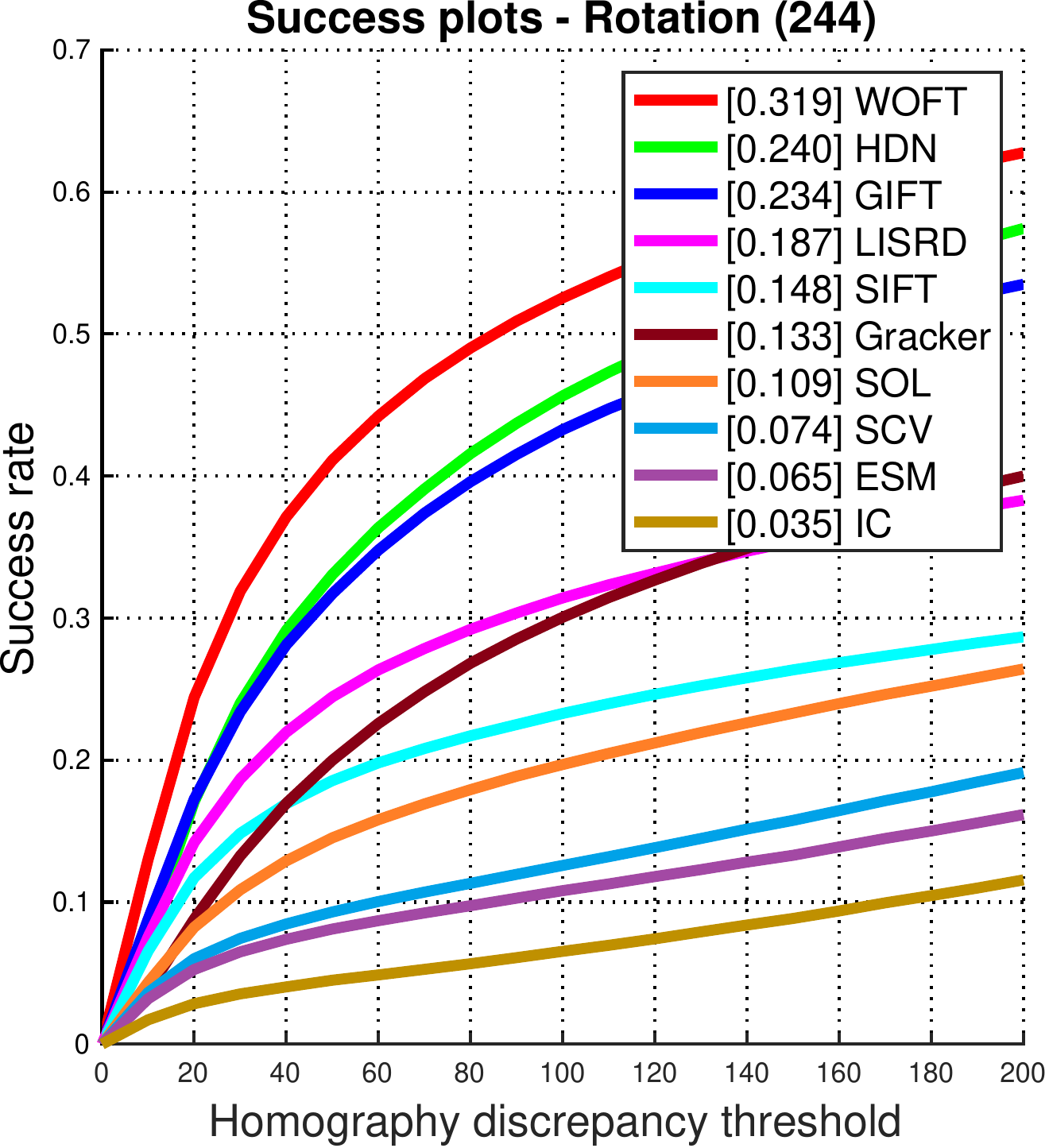} \includegraphics[width=3.99cm]{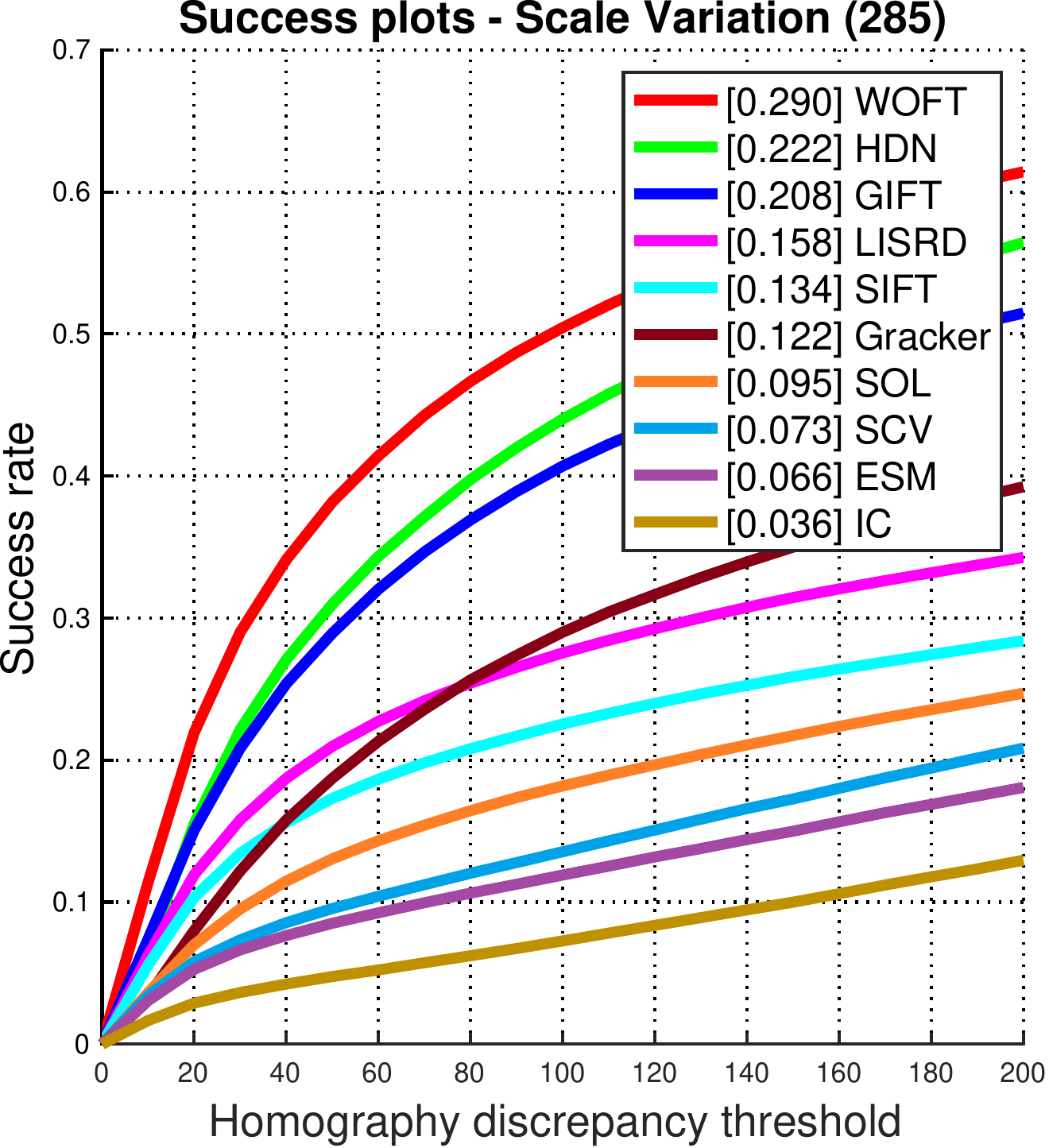} \\
\end{tabular}
\caption{Performance of trackers on each challenging factor using success. Best viewed in color.
}
\label{att_suc}
\end{figure}

We display more challenging factor-based results on PlanarTrack in this section. Fig.~\ref{att_pre} shows performance of trackers on each challenging factor using
precision, and Fig.~\ref{att_suc} the results on different challenges using success.

\subsection*{S3. Detailed Construction of PlanarTrack$_\text{BB}$}

In order to study the performance of generic object trackers in dealing with planar-like targets, we further develop a new benchmark named PlanarTrack$_\text{BB}$ based on our PlanarTrack. We achieve this by converting the four annotated corner points of the planar target into an axis-aligned bounding box. Suppose the four annotated points of the planar target are denoted as $\{(p_1^x, p_1^y), (p_2^x, p_2^y), (p_3^x, p_3^y), (p_4^x, p_4^y)\}$, then the axis-aligned box of the target will be formulated as $\{(x_{tf}, y_{tf}), (x_{br}, y_{br})\}$, where $(x_{tf}, y_{tf})$ and $(x_{br}, y_{br})$ are the coordinates of the top-left and bottom-right points of the bounding box and are obtained via

$x_{tf}=\text{max}(\text{min}(p_1^x, p_2^x, p_3^x, p_4^x), 1)$

$y_{tf}=\text{max}(\text{min}(p_1^y, p_2^y, p_3^y, p_4^y), 1)$

$x_{br}=\text{min}(\text{max}(p_1^x, p_2^x, p_3^x, p_4^x), w_\text{img})$

$y_{br}=\text{min}(\text{max}(p_1^y, p_2^y, p_3^y, p_4^y), h_\text{img})$\\
where $w_\text{img}$ and $h_\text{img}$ represent image width and height. We show several examples from PlanarTrack$_\text{BB}$ in Fig.~\ref{suppvis}. 

\newcommand{\MMM}{0.31}
\newcommand{\MM}{0.23}
\begin{figure}[!t]
		\centering
		\begin{tabular}{c}
\includegraphics[width=\MMM\linewidth]{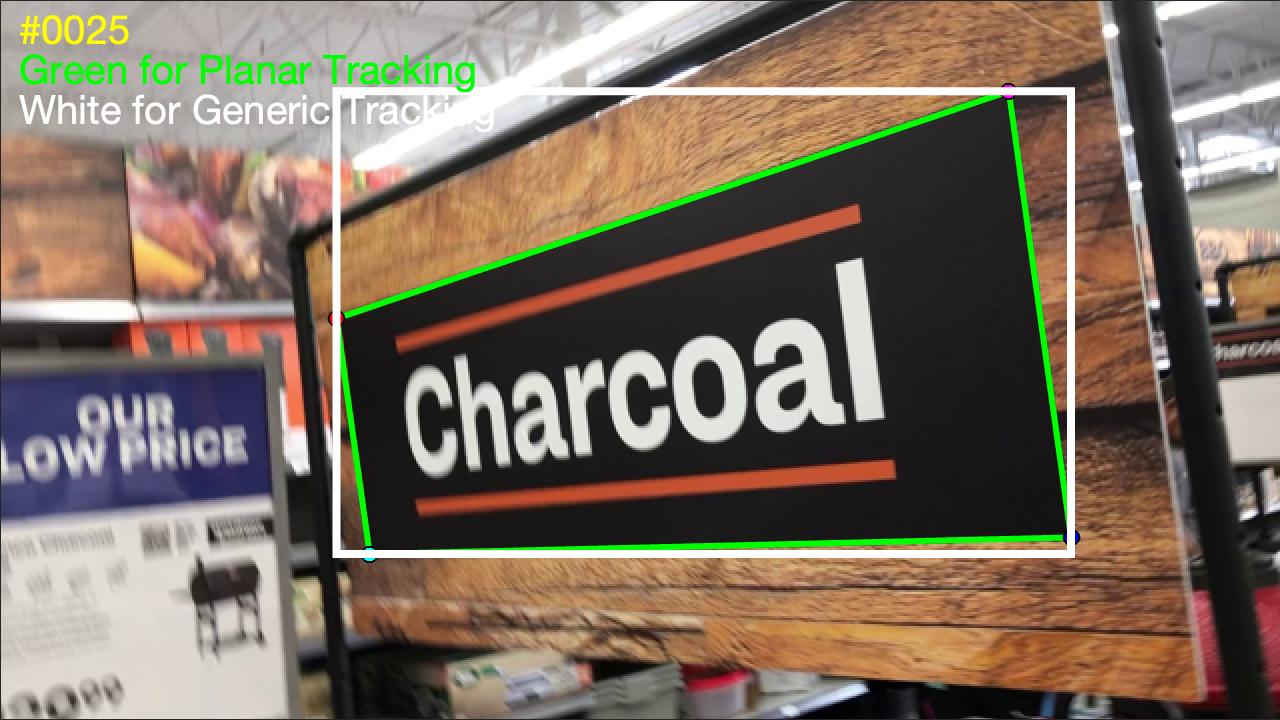} \includegraphics[width=\MMM\linewidth]{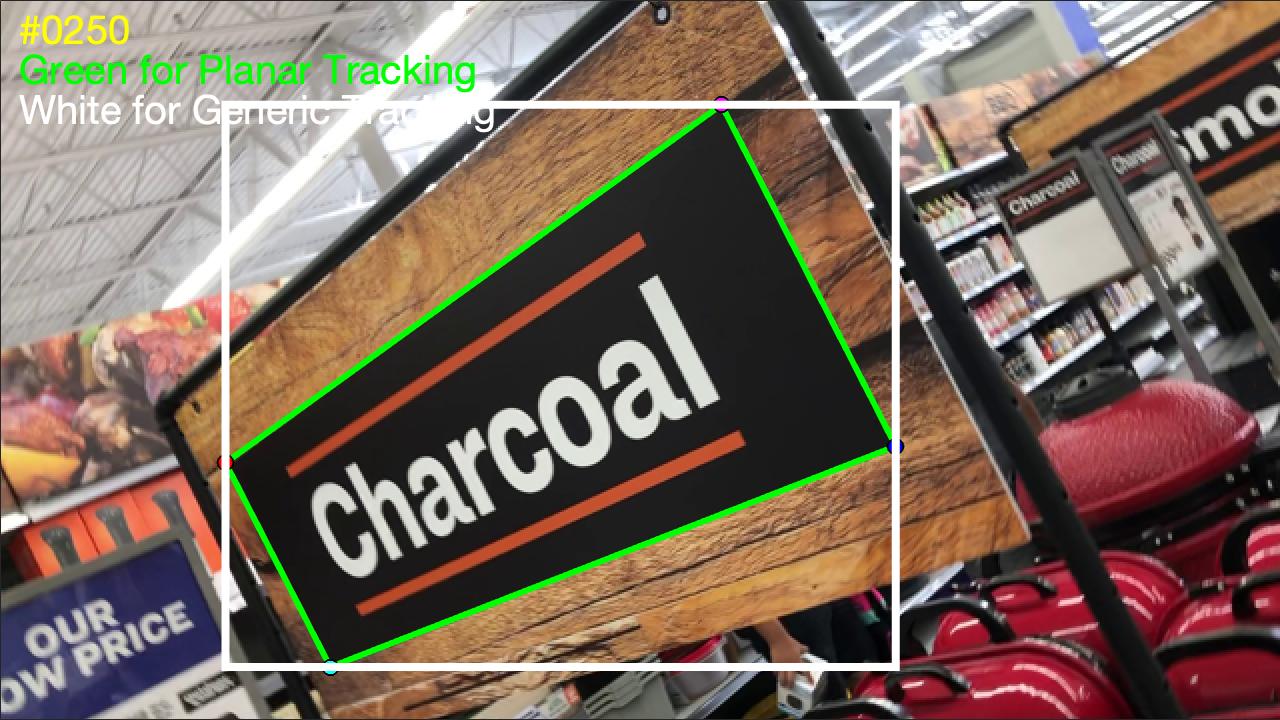}
\includegraphics[width=\MMM\linewidth]{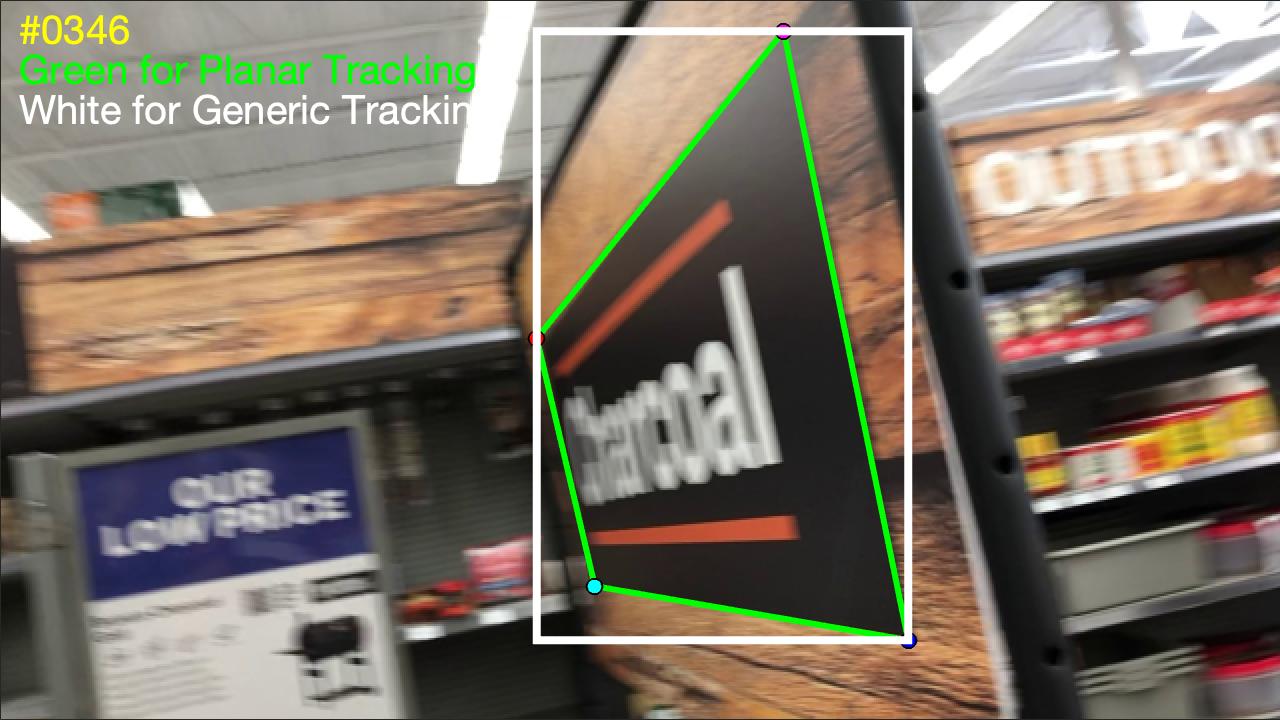}\\
\includegraphics[width=\MMM\linewidth]{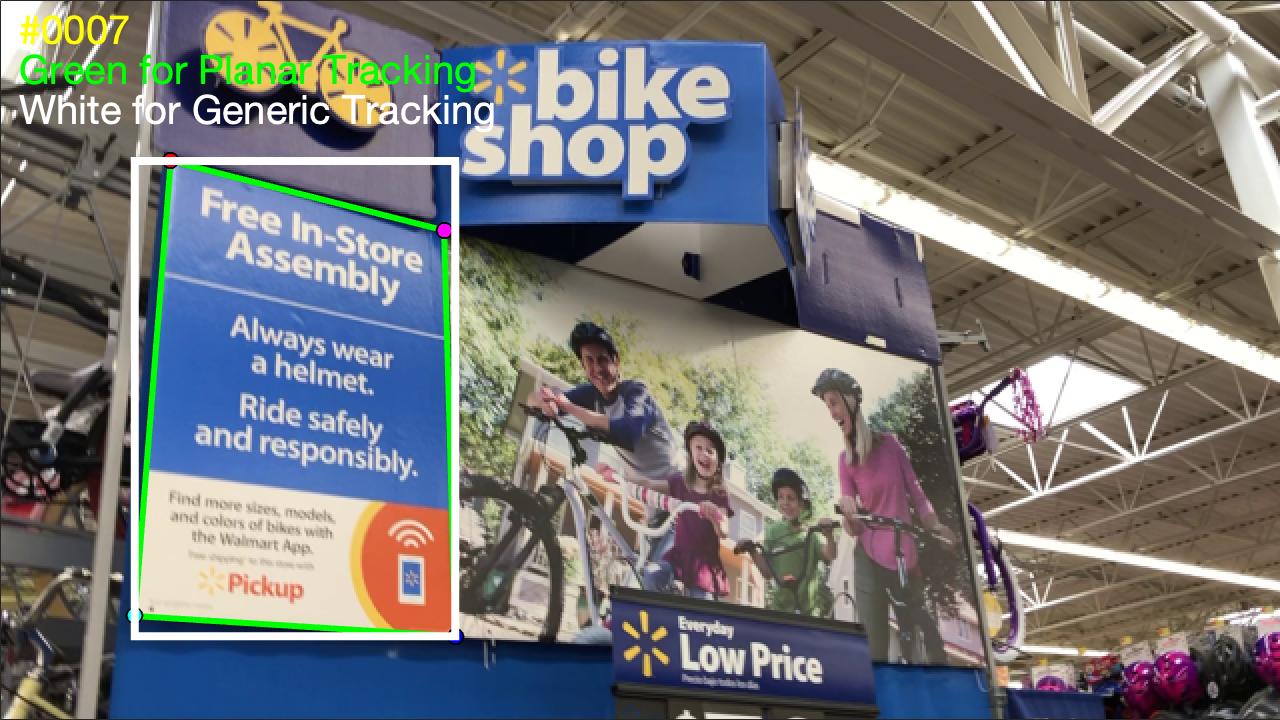} \includegraphics[width=\MMM\linewidth]{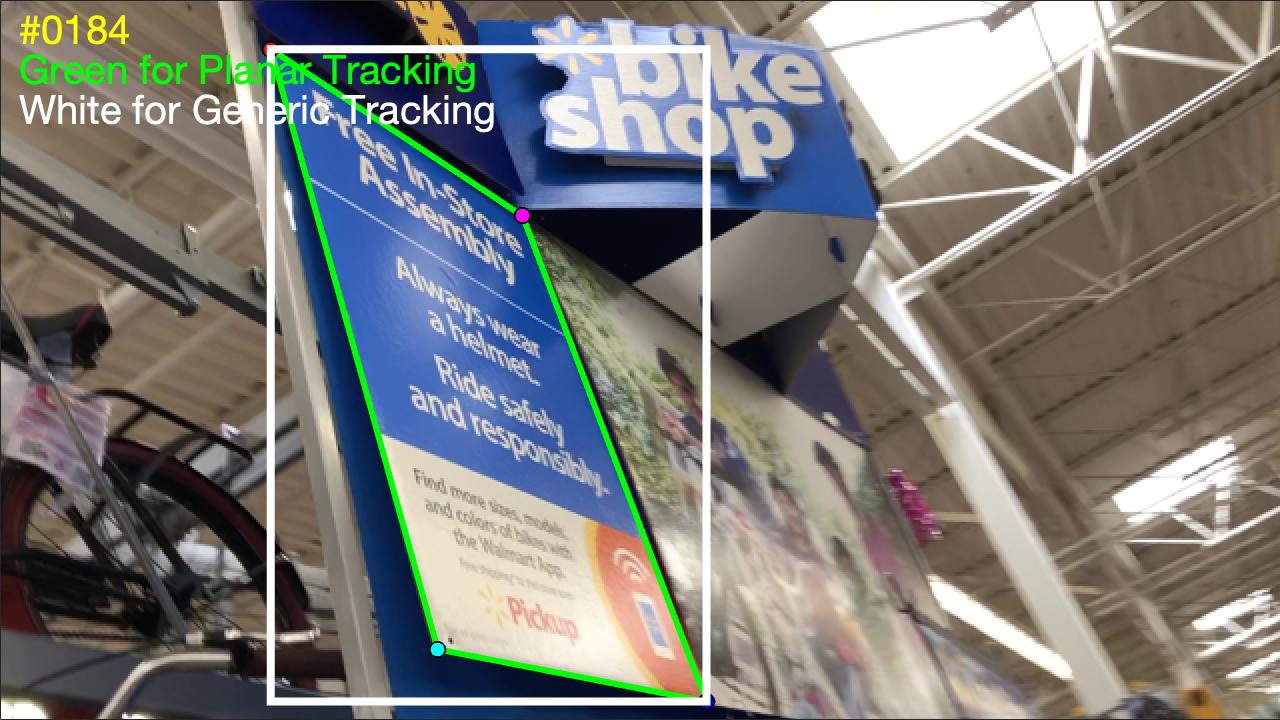}
\includegraphics[width=\MMM\linewidth]{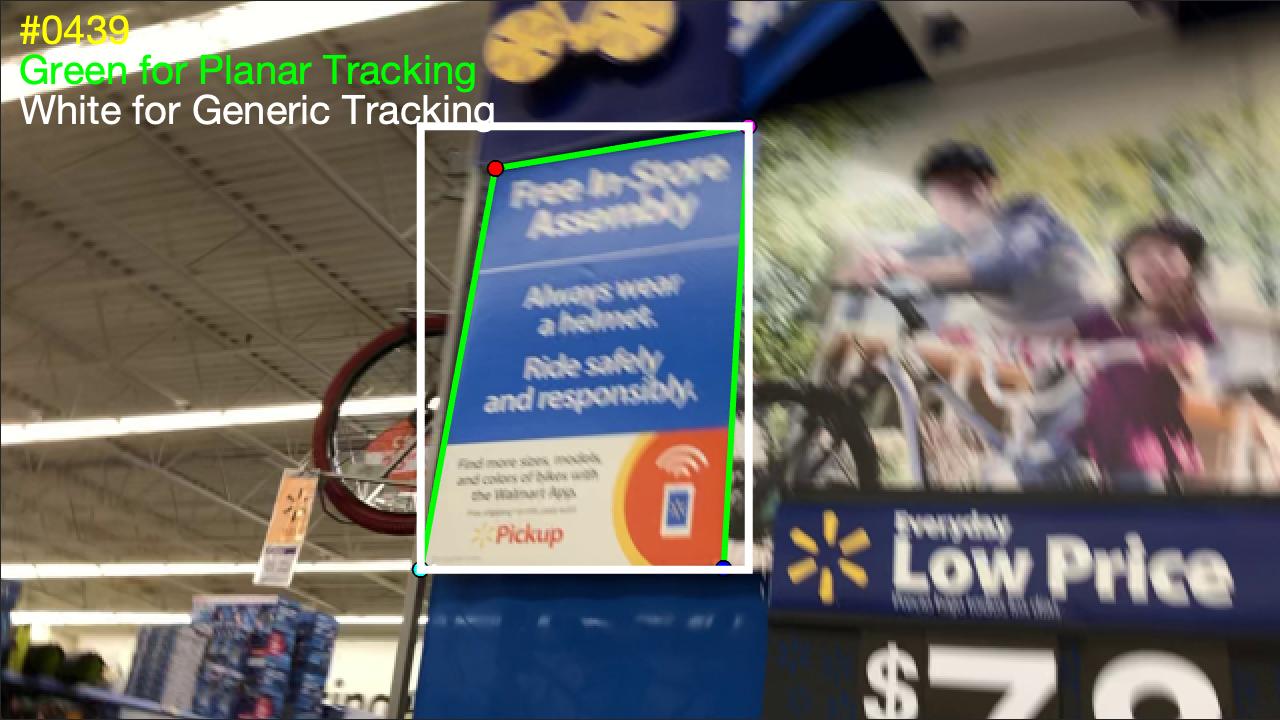}\\
\includegraphics[width=\MM\linewidth]{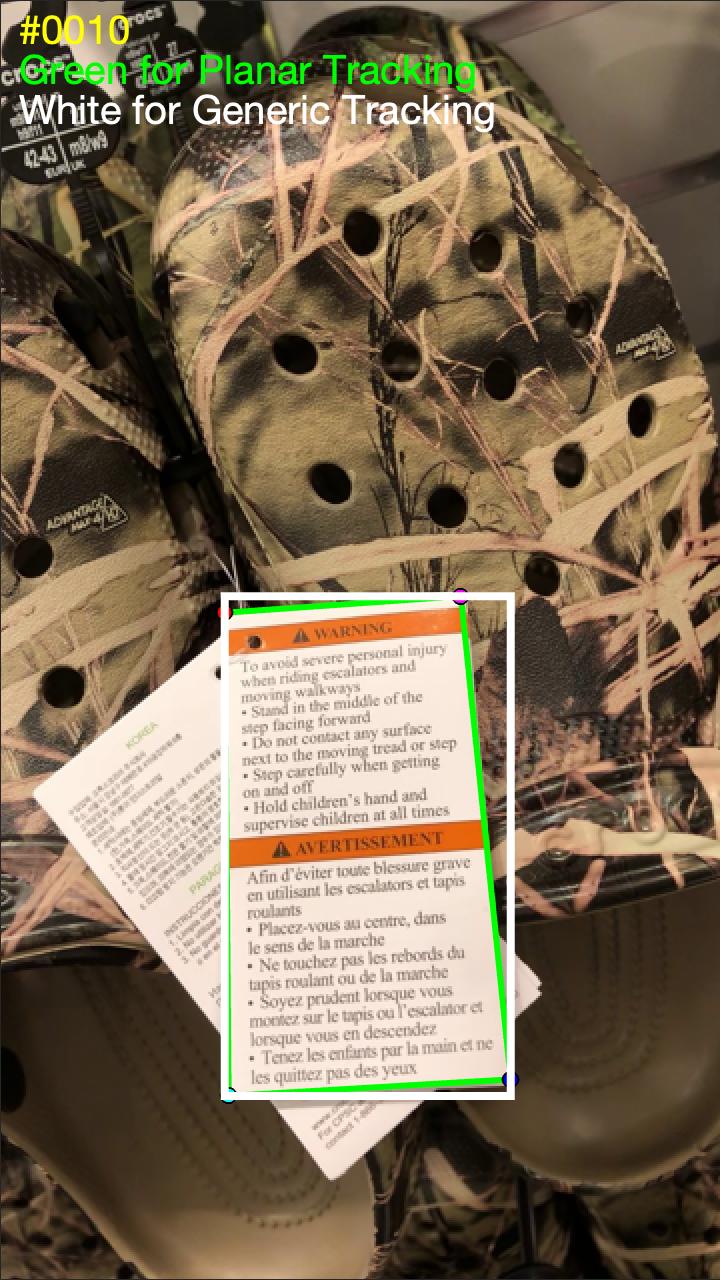} \includegraphics[width=\MM\linewidth]{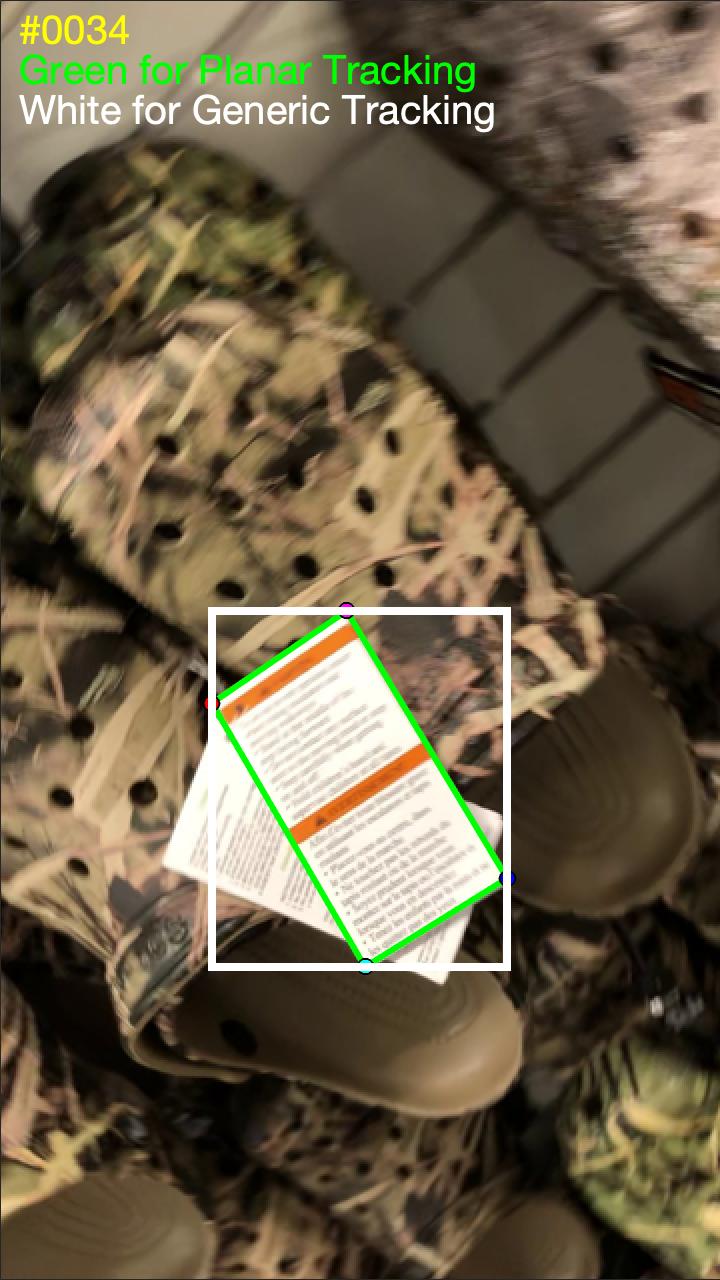}
\includegraphics[width=\MM\linewidth]{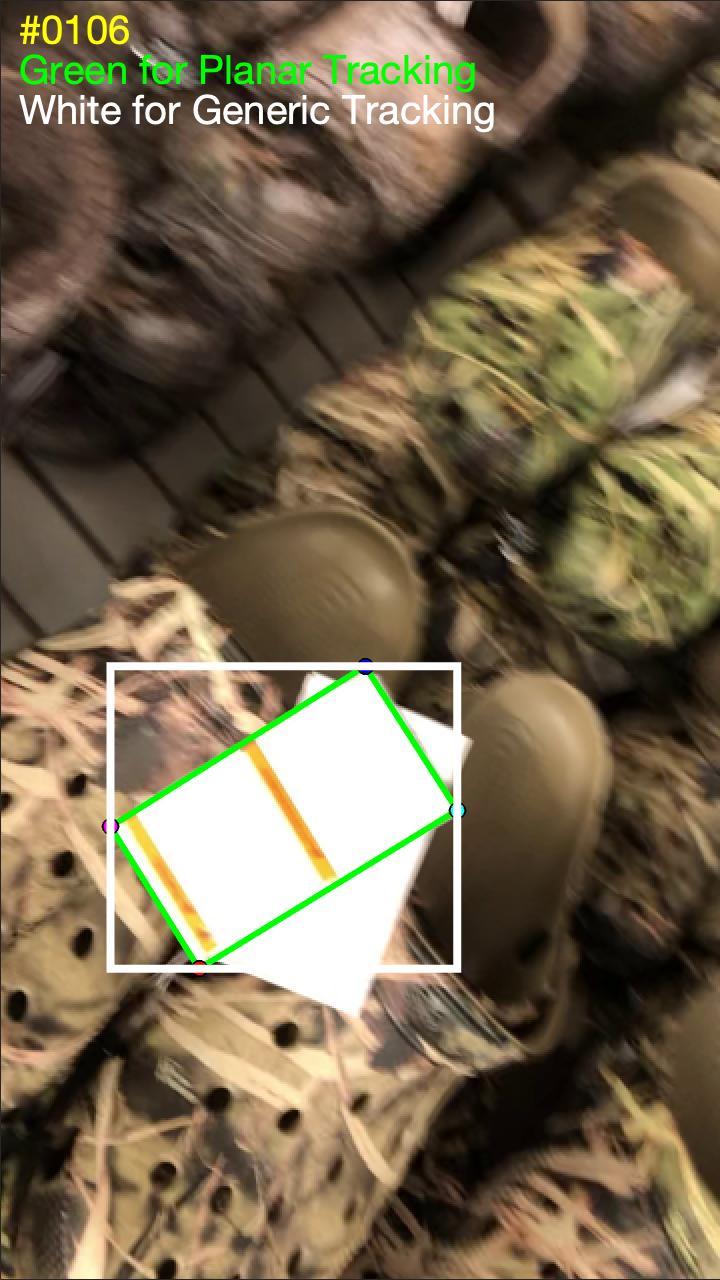}
\includegraphics[width=\MM\linewidth]{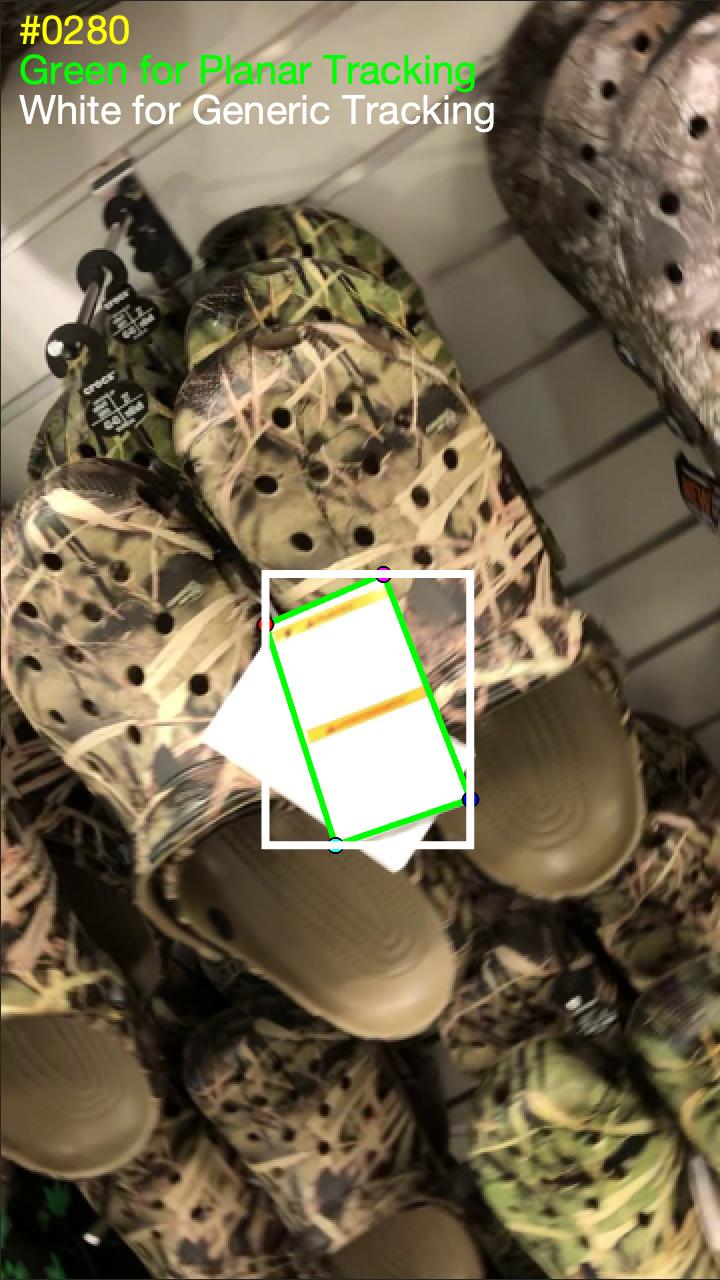}\\

\end{tabular}
\caption{Examples from PlanarTrack$_\text{BB}$. The targets are annotated by white axis-align bounding boxes for genetic visual tracking. Best viewed in color.
}
\label{suppvis}
\end{figure}

\newcommand{\M}{0.47}
\begin{figure}[!t]
		\centering
		\begin{tabular}{c}
\includegraphics[width=\M\linewidth]{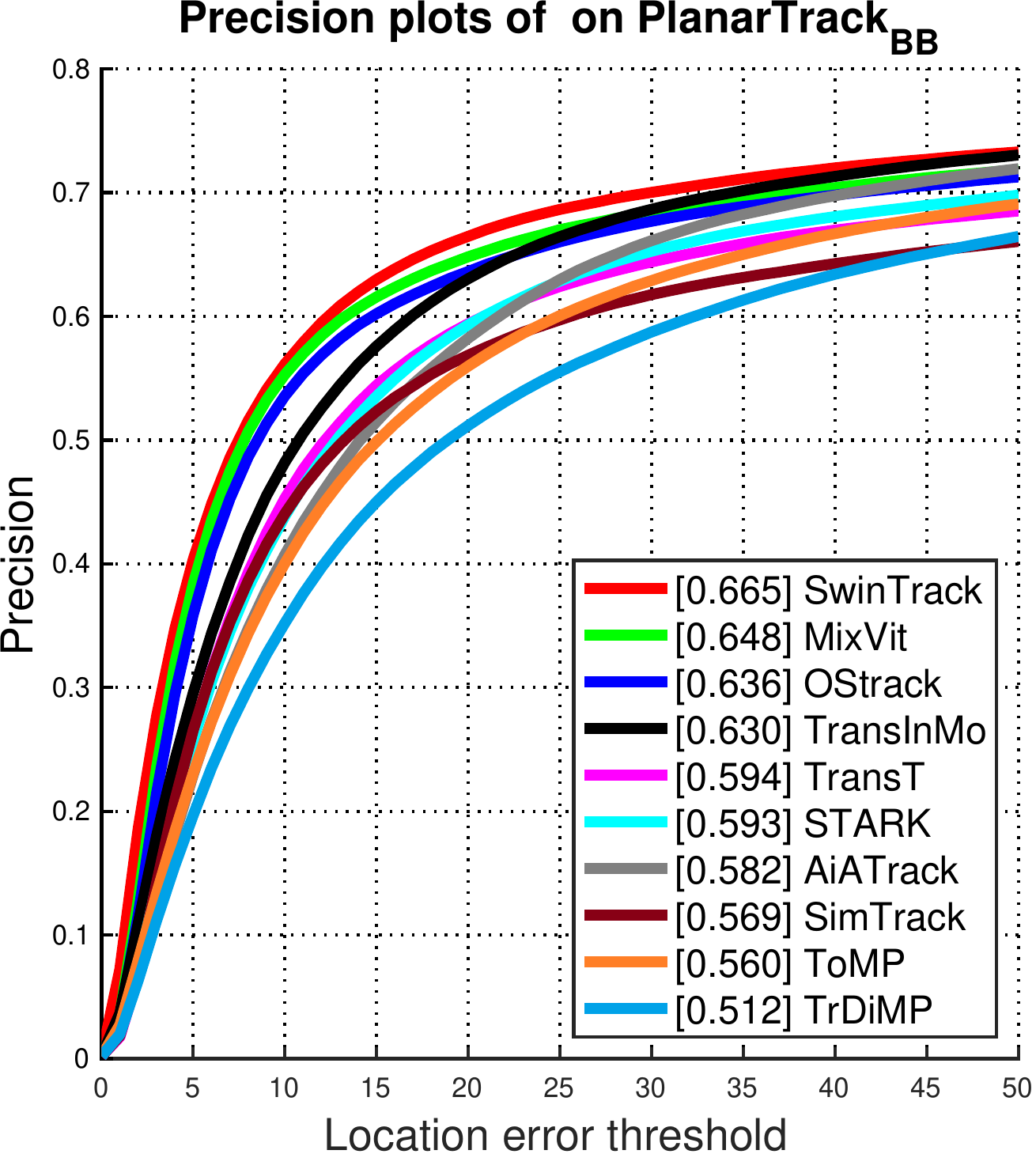} \includegraphics[width=\M\linewidth]{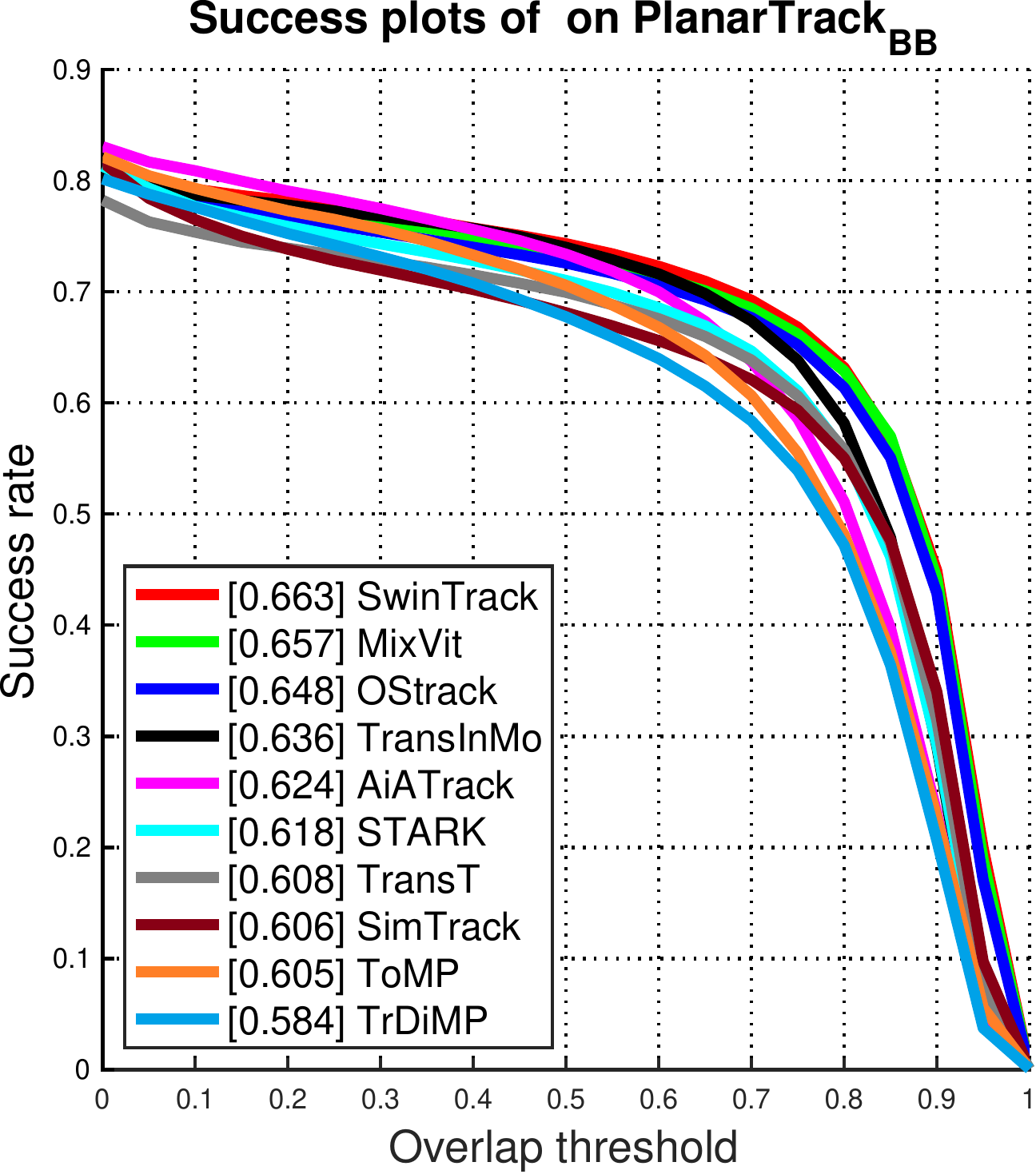}\\

\end{tabular}
\caption{Performance of ten evaluated generic visual trackers on PlanarTrack$_\text{BB}$ using bounding box-based precision and success plots. Best viewed in color.
}
\label{genvot}
\end{figure}

\subsection*{S4. More Results on PlanarTrack$_\text{BB}$}

Fig.~\ref{genvot} demonstrates the evaluation results of ten excellent generic trackers on PlanarTrack$_\text{BB}$. We utilize bounding box-based precision and success plots as in generic tracking evaluation for assessment.

{\small
\bibliographystyle{ieee_fullname}
\bibliography{egbib}

\begin{thebibliography}{10}\itemsep=-1pt

\bibitem{baker2004lucas}
Simon Baker and Iain Matthews.
\newblock Lucas-kanade 20 years on: A unifying framework.
\newblock {\em IJCV}, 56:221--255, 2004.

\bibitem{bay2006surf}
Herbert Bay, Tinne Tuytelaars, and Luc Van~Gool.
\newblock Surf: Speeded up robust features.
\newblock In {\em ECCV}, 2006.

\bibitem{benhimane2004real}
Selim Benhimane and Ezio Malis.
\newblock Real-time image-based tracking of planes using efficient second-order
  minimization.
\newblock In {\em IROS}, 2004.

\bibitem{chen2022backbone}
Boyu Chen, Peixia Li, Lei Bai, Lei Qiao, Qiuhong Shen, Bo Li, Weihao Gan, Wei
  Wu, and Wanli Ouyang.
\newblock Backbone is all your need: a simplified architecture for visual
  object tracking.
\newblock In {\em ECCV}, 2022.

\bibitem{chen2017illumination}
Lin Chen, Fan Zhou, Yu Shen, Xiang Tian, Haibin Ling, and Yaowu Chen.
\newblock Illumination insensitive efficient second-order minimization for
  planar object tracking.
\newblock In {\em ICRA}, 2017.

\bibitem{chen2021transformer}
Xin Chen, Bin Yan, Jiawen Zhu, Dong Wang, Xiaoyun Yang, and Huchuan Lu.
\newblock Transformer tracking.
\newblock In {\em CVPR}, 2021.

\bibitem{cui2022mixformer}
Yutao Cui, Cheng Jiang, Limin Wang, and Gangshan Wu.
\newblock Mixformer: End-to-end tracking with iterative mixed attention.
\newblock In {\em CVPR}, 2022.

\bibitem{dick2013realtime}
Travis Dick, Camilo~Perez Quintero, Martin J{\"a}gersand, and Azad Shademan.
\newblock Realtime registration-based tracking via approximate nearest
  neighbour search.
\newblock In {\em RSS}, 2013.

\bibitem{fan2021lasot}
Heng Fan, Hexin Bai, Liting Lin, Fan Yang, Peng Chu, Ge Deng, Sijia Yu,
  Mingzhen Huang, Juehuan Liu, Yong Xu, et~al.
\newblock Lasot: A high-quality large-scale single object tracking benchmark.
\newblock {\em IJCV}, 129:439--461, 2021.

\bibitem{fan2019lasot}
Heng Fan, Liting Lin, Fan Yang, Peng Chu, Ge Deng, Sijia Yu, Hexin Bai, Yong
  Xu, Chunyuan Liao, and Haibin Ling.
\newblock Lasot: A high-quality benchmark for large-scale single object
  tracking.
\newblock In {\em CVPR}, 2019.

\bibitem{fan2021transparent}
Heng Fan, Halady~Akhilesha Miththanthaya, Siranjiv~Ramana Rajan, Xiaoqiong Liu,
  Zhilin Zou, Yuewei Lin, Haibin Ling, et~al.
\newblock Transparent object tracking benchmark.
\newblock In {\em ICCV}, 2021.

\bibitem{gao2022aiatrack}
Shenyuan Gao, Chunluan Zhou, Chao Ma, Xinggang Wang, and Junsong Yuan.
\newblock Aiatrack: Attention in attention for transformer visual tracking.
\newblock In {\em ECCV}, 2022.

\bibitem{gauglitz2011evaluation}
Steffen Gauglitz, Tobias H{\"o}llerer, and Matthew Turk.
\newblock Evaluation of interest point detectors and feature descriptors for
  visual tracking.
\newblock {\em IJCV}, 94:335--360, 2011.

\bibitem{guo2022learning}
Mingzhe Guo, Zhipeng Zhang, Heng Fan, Liping Jing, Yilin Lyu, Bing Li, and
  Weiming Hu.
\newblock Learning target-aware representation for visual tracking via
  informative interactions.
\newblock In {\em IJCAI}, 2022.

\bibitem{hare2012efficient}
Sam Hare, Amir Saffari, and Philip~HS Torr.
\newblock Efficient online structured output learning for keypoint-based object
  tracking.
\newblock In {\em CVPR}, 2012.

\bibitem{huang2019got}
Lianghua Huang, Xin Zhao, and Kaiqi Huang.
\newblock Got-10k: A large high-diversity benchmark for generic object tracking
  in the wild.
\newblock {\em TPAMI}, 43(5):1562--1577, 2021.

\bibitem{liang2021planar}
Pengpeng Liang, Haoxuanye Ji, Yifan Wu, Yumei Chai, Liming Wang, Chunyuan Liao,
  and Haibin Ling.
\newblock Planar object tracking benchmark in the wild.
\newblock {\em Neurocomputing}, 454:254--267, 2021.

\bibitem{liang2018planar}
Pengpeng Liang, Yifan Wu, Hu Lu, Liming Wang, Chunyuan Liao, and Haibin Ling.
\newblock Planar object tracking in the wild: A benchmark.
\newblock In {\em ICRA}, 2018.

\bibitem{lieberknecht2009dataset}
Sebastian Lieberknecht, Selim Benhimane, Peter Meier, and Nassir Navab.
\newblock A dataset and evaluation methodology for template-based tracking
  algorithms.
\newblock In {\em ISMAR}, 2009.

\bibitem{lin2022swintrack}
Liting Lin, Heng Fan, Zhipeng Zhang, Yong Xu, and Haibin Ling.
\newblock Swintrack: A simple and strong baseline for transformer tracking.
\newblock In {\em NeurIPS}, 2022.

\bibitem{lin2014microsoft}
Tsung-Yi Lin, Michael Maire, Serge Belongie, James Hays, Pietro Perona, Deva
  Ramanan, Piotr Doll{\'a}r, and C~Lawrence Zitnick.
\newblock Microsoft coco: Common objects in context.
\newblock In {\em ECCV}, 2014.

\bibitem{liu2019gift}
Yuan Liu, Zehong Shen, Zhixuan Lin, Sida Peng, Hujun Bao, and Xiaowei Zhou.
\newblock Gift: Learning transformation-invariant dense visual descriptors via
  group cnns.
\newblock {\em NeurIPS}, 2019.

\bibitem{lowe2004distinctive}
David~G Lowe.
\newblock Distinctive image features from scale-invariant keypoints.
\newblock {\em IJCV}, 60:91--110, 2004.

\bibitem{mayer2022transforming}
Christoph Mayer, Martin Danelljan, Goutam Bhat, Matthieu Paul, Danda~Pani
  Paudel, Fisher Yu, and Luc Van~Gool.
\newblock Transforming model prediction for tracking.
\newblock In {\em CVPR}, 2022.

\bibitem{muller2018trackingnet}
Matthias Muller, Adel Bibi, Silvio Giancola, Salman Alsubaihi, and Bernard
  Ghanem.
\newblock Trackingnet: A large-scale dataset and benchmark for object tracking
  in the wild.
\newblock In {\em ECCV}, 2018.

\bibitem{ozuysal2009fast}
Mustafa Ozuysal, Michael Calonder, Vincent Lepetit, and Pascal Fua.
\newblock Fast keypoint recognition using random ferns.
\newblock {\em TPAMI}, 32(3):448--461, 2009.

\bibitem{pautrat2020online}
R{\'e}mi Pautrat, Viktor Larsson, Martin~R Oswald, and Marc Pollefeys.
\newblock Online invariance selection for local feature descriptors.
\newblock In {\em ECCV}, 2020.

\bibitem{richa2011visual}
Rog{\'e}rio Richa, Raphael Sznitman, Russell Taylor, and Gregory Hager.
\newblock Visual tracking using the sum of conditional variance.
\newblock In {\em IROS}, 2011.

\bibitem{roy2015tracking}
Ankush Roy, Xi Zhang, Nina Wolleb, Camilo~Perez Quintero, and Martin
  J{\"a}gersand.
\newblock Tracking benchmark and evaluation for manipulation tasks.
\newblock In {\em ICRA}, 2015.

\bibitem{vserych2023planar}
Jon{\'a}{\v{s}} {\v{S}}er{\`y}ch and Ji{\v{r}}{\'\i} Matas.
\newblock Planar object tracking via weighted optical flow.
\newblock In {\em WACV}, 2023.

\bibitem{valmadre2018long}
Jack Valmadre, Luca Bertinetto, Joao~F Henriques, Ran Tao, Andrea Vedaldi,
  Arnold~WM Smeulders, Philip~HS Torr, and Efstratios Gavves.
\newblock Long-term tracking in the wild: A benchmark.
\newblock In {\em ECCV}, 2018.

\bibitem{wang2021transformer}
Ning Wang, Wengang Zhou, Jie Wang, and Houqiang Li.
\newblock Transformer meets tracker: Exploiting temporal context for robust
  visual tracking.
\newblock In {\em CVPR}, 2021.

\bibitem{wang2017gracker}
Tao Wang and Haibin Ling.
\newblock Gracker: A graph-based planar object tracker.
\newblock {\em TPAMI}, 40(6):1494--1501, 2017.

\bibitem{wang2021towards}
Xiao Wang, Xiujun Shu, Zhipeng Zhang, Bo Jiang, Yaowei Wang, Yonghong Tian, and
  Feng Wu.
\newblock Towards more flexible and accurate object tracking with natural
  language: Algorithms and benchmark.
\newblock In {\em CVPR}, 2021.

\bibitem{wu2013online}
Yi Wu, Jongwoo Lim, and Ming-Hsuan Yang.
\newblock Online object tracking: A benchmark.
\newblock In {\em CVPR}, 2013.

\bibitem{yan2021learning}
Bin Yan, Houwen Peng, Jianlong Fu, Dong Wang, and Huchuan Lu.
\newblock Learning spatio-temporal transformer for visual tracking.
\newblock In {\em ICCV}, 2021.

\bibitem{ye2022joint}
Botao Ye, Hong Chang, Bingpeng Ma, Shiguang Shan, and Xilin Chen.
\newblock Joint feature learning and relation modeling for tracking: A
  one-stream framework.
\newblock In {\em ECCV}, 2022.

\bibitem{zhan2022homography}
Xinrui Zhan, Yueran Liu, Jianke Zhu, and Yang Li.
\newblock Homography decomposition networks for planar object tracking.
\newblock In {\em AAAI}, 2022.

\bibitem{zhang2022hvc}
Haoxian Zhang and Yonggen Ling.
\newblock Hvc-net: Unifying homography, visibility, and confidence learning for
  planar object tracking.
\newblock In {\em ECCV}, 2022.

\end{thebibliography}
}

\end{document}